\documentclass{article}

\usepackage{arxiv}

\usepackage[utf8]{inputenc}
\usepackage[T1]{fontenc}
\PassOptionsToPackage{hyphens}{url}\usepackage{hyperref}
\usepackage{url}
\usepackage{booktabs}
\usepackage{amsfonts}
\usepackage{nicefrac}
\usepackage{microtype}
\usepackage{lipsum}
\usepackage{graphicx}
\usepackage{multirow}
\usepackage{wrapfig}
\usepackage{rotate}
\usepackage{float}
\usepackage{xcolor}
\usepackage{amsmath}
\usepackage{natbib}

\DeclareMathOperator*{\argmin}{arg\,min}

\title{Learning to Generate Wasserstein Barycenters}

\author{
 Julien Lacombe  \\
  INSA Lyon, Univ. Lyon \\
  Lyon, France \\
  \texttt{jlacombe@protonmail.com} \\
 \And
 Julie Digne   \\
  CNRS, Univ. Lyon \\
  Lyon, France \\
 \And
 Nicolas Courty   \\
  CNRS, IRISA, Univ. Bretagne Sud \\
  Vannes, France \\
 \And
 Nicolas Bonneel  \\
  CNRS, Univ. Lyon \\
  Lyon, France \\
}

\begin{document}
\maketitle
\begin{abstract}
Optimal transport is a notoriously difficult problem to solve numerically, with current approaches often remaining intractable for very large scale applications such as those encountered in machine learning. Wasserstein barycenters -- the problem of finding measures in-between given input measures in the optimal transport sense -- is even more computationally demanding as it requires to solve an optimization problem involving optimal transport distances. By training a deep convolutional neural network, we improve by a factor of 60 the computational speed of Wasserstein barycenters over the fastest state-of-the-art approach on the GPU, resulting in milliseconds computational times on $512\times512$ regular grids. We show that our network, trained on Wasserstein barycenters of pairs of measures, generalizes well to the problem of finding Wasserstein barycenters of more than two measures. We demonstrate the efficiency of our approach for computing barycenters of sketches and transferring colors between multiple images.
\end{abstract}

\section{Introduction}
\label{sec:introduction}

Optimal transport is becoming widespread in machine learning, but also in computer graphics, vision and many other disciplines. Its framework allows for comparing probability distributions, shapes or images, as well as producing interpolations of these data. As a result, it has been used in the context of machine learning as a loss for training neural networks~\citep{arjovsky2017}, as a manifold for dictionary learning~\citep{SHBMCCPS18}, clustering~\citep{mi2018variational} and metric learning applications~\citep{HBCCP19}, as a way to sample an embedding~\citep{liutkus2019sliced} and transfer learning~\citep{courty2014domain}, and many other applications (see Sec.~\ref{sec:appml}).
However, despite recent progress in computational optimal transport, in many cases these applications have remained limited to small datasets due to the substantial computational cost of optimal transport, in terms of speed, but also memory.

We tackle the problem of efficiently computing Wasserstein barycenters of measures discretized on regular grids, a setting common to several of these machine learning applications. Wasserstein barycenters are interpolations of two or more probability distributions under optimal transport distances. As such, a common way to obtain them is to perform a minimization of a functional involving optimal transport distances or transport plans, which is thus a very costly process. Instead, we directly predict Wasserstein barycenters by training a Deep Convolutional Neural Network (DCNN) specific to this task.

An important challenge behind our work is to build an architecture that can handle a variable number of input measures with associated weights without needing to retrain a specific network. 
To achieve that, we specify and adapt an architecture designed for and trained with two input measures, and show that we can use this modified network without retraining to compute barycenters of more than two measures. Directly predicting Wasserstein barycenters avoids the need to compute a Wasserstein embedding~\citep{courty2017learning}, and our experiments suggest that this results in better Wasserstein barycenters approximations.
Our implementation is publicly available\footnote{\scriptsize{\url{https://github.com/jlacombe/learning-to-generate-wasserstein-barycenters}}}.

\paragraph{Contributions} This paper introduces a method to compute Wasserstein barycenters in milliseconds. It shows that this can be done by learning Wasserstein barycenters of only two measures on a dataset of random shapes using a DCNN, and by adapting this DCNN to handle multiple input measures without retraining. This proposed approach is 60x faster than the fastest state-of-the-art GPU library, and performs better than Wasserstein embeddings.

\section{Related Work}
\label{sec:related-work}

\subsection{Wasserstein distances and approximations} 
Optimal transport seeks the best way to warp a given probability measure $\mu_{0}$ to form another given probability measure $\mu_{1}$ by minimizing the total cost of moving individual ``particles of earth''. We restrict our description to discrete distributions. In this setting, finding the optimal transport between two probability measures is often achieved by solving a large linear program~\citep{kantorovich1942transfer} -- more details on this theory and numerical tools can be found in the book of \citet{peyre2019computational}. This minimization results in the so-called \emph{Wasserstein distance}, the mathematical distance defined by the total cost of reshaping  $\mu_{0}$ to $\mu_{1}$. This distance can be used to compare probability distributions, in particular in a machine learning context. It also results in a \emph{transport plan}, a matrix $P(x,y)$ representing the amount of mass of $\mu_0$ traveling from location $x$ in $\mu_0$ towards location $y$ in $\mu_1$.

However, the Wasserstein distance is notoriously difficult to compute -- the corresponding linear program is huge, and dedicated solvers typically solve this problem in $\mathcal{O}(N^3 \log N)$, with $N$ the size of the input measures discretization.
Recently, numerous approaches have attempted to approximate Wasserstein distances. One of the most efficient methods, the so-called Sinkhorn algorithm introduces an entropic regularization, allowing to compute such distances by iteratively performing fast matrix-vector multiplications~\citep{cuturi2013sinkhorn} or convolutions in the case of regular grids~\citep{solomon2015convolutional}. However, this comes at the expense of smoothing the transport plan and removing guarantees regarding this mathematical distance (in particular, the regularized cost $W_\epsilon(\mu_0, \mu_0) \neq 0$).
These issues are addressed by \emph{Sinkhorn divergences}~\citep{feydy2018interpolating,genevay2017learning}. This approach symmetrizes the entropy-regularized optimal transport distance, adding guarantees on this divergence (now, the cost $S_\epsilon(\mu_0, \mu_0) = 0$ by construction, although triangular inequality still does not hold) but also effectively reducing blur, while maintaining a relatively fast numerical algorithm. They show that this divergence interpolates between  optimal transport distances and Maximum Mean Discrepancies.
Sinkhorn divergences are implemented in the \emph{GeomLoss} library~\citep{geomloss}, relying on a specific computational scheme on the GPU~\citep{feydy2019fast,feydy2018interpolating,schmitzer2019stabilized} and constitutes the state-of-the-art in term of speed and approximation of optimal transport-like distances.

\subsection{Wasserstein barycenters} 
The Wasserstein barycenter of a set of probability measures corresponds to the Fréchet mean of these measures under the Wasserstein distance (i.e., a weighted mean under the Wasserstein metric). Wasserstein barycenters allow to interpolate between two or more probability measures by warping these measures (contrarily to Euclidean barycenters that blends them).
Similarly to Wasserstein distances, Wasserstein barycenters are very expensive to compute. An entropy-regularized approach based on Sinkhorn-like iterations also allows to efficiently compute blurred Wasserstein barycenters. Reducing blur via Sinkhorn divergences is also doable, but does not benefit from a very fast Sinkhorn-like algorithm: a weighted sum of Sinkhorn divergences needs to be iteratively minimized, which adds significant computational cost.
In our approach, we rely on Sinkhorn divergence-based barycenters to feed training data to a Deep Convolutional Neural Network, and thus aim at speeding up the generation of approximate wasserstein barycenters. 
Other fast transport-based barycenters include that of sliced and Radon Wasserstein barycenters, obtained via Wasserstein barycenters on 1-d projections (\cite{rabin2011wasserstein}, \cite{bonneel2015sliced}), which we compare to.

A recent trend seeks linearizations or Euclidean embeddings of optimal transport problems. Notably, \citet{Nader18} approximate Wasserstein barycenters by first solving an optimal transport map between a uniform measure towards $n$ input measures, and then linearly combining Monge maps. This allows for efficient computations -- typically of the order of half a second for 512x512 images. A similar approach is taken within the documentation of the GeomLoss library~\citep{geomloss}\footnote{\scriptsize{See  \url{https://www.kernel-operations.io/geomloss/\_auto_examples/optimal\_transport/plot\_wasserstein\_barycenters\_2D.html}}}, where a single step of a gradient descent initialized with a uniform distribution is used, which effectively corresponds to such linearization. We use this technique in our work to train our network.
\cite{wang2013linear}, \cite{Moosmuller20} and \cite{Merigot20} use a similar linearization, possibly using a non-uniform reference measure, with theoretical guarantees on the distorsion introduced by the embedding. 
Instead of explicitly building an embedding via Monge maps, such an embedding can be learned. \citet{courty2017learning} propose a siamese neural network architecture to learn an embedding in which the Euclidean distance approximates the Wasserstein distance. Wasserstein barycenters can then be approximated by interpolating within the Euclidean embedding, without requiring explicit computations of transport plans. 
They show accurate barycenters on a number of datasets of low resolution ($28\times28$). However, in general, it is unclear whether Wasserstein metrics embed into Euclidean spaces. Negative results were shown for 3d optimal transport onto a Euclidean space~\citep{andoni2016impossibility}.
Interestingly, in the reversed direction, Wasserstein spaces have been used to embed other metrics~\citep{frogner2019learning}.

Wasserstein barycenters can also be seen as a particular instance of inverse problem. There is an important literature on the resolution of inverse problems with deep learning models on instances such as (non-exhaustive list) image denoising~\citep{ulyanov2018deep}~\citep{burger2012image} ~\citep{lefkimmiatis2017non}, super-resolution ~\citep{ledig2017photo}, ~\citep{tai2017image}, ~\citep{lai2017deep}, inpainting ~\citep{yeh2017semantic}~\citep{xie2012image},~\citep{liu2018image}.

Parallel to our work,~\citet{fan2020scalable} propose a model based on input convex neural networks (ICNN) developed by~\citet{amos2017input}. Their method allows a fast approximation of Wasserstein barycenters of continuous input measures. This last work is also closely related to the semi-discrete approach of~\citet{claici2018stochastic}. 

\subsection{Applications to machine learning}
\label{sec:appml}
For its ability to compare probability measures, optimal transport has met an important success in machine learning. This is particularly the case of Wasserstein GANs~\citep{arjovsky2017} that compute a very efficient approximation of Wasserstein distances as a loss for generative adversarial models. The optimal transport loss has also been used in the context of dictionary learning~\citep{rolet2016fast}.
Other fast approximations have allowed to perform domain adaptation for transfer learning of a classifier, by advecting samples via a computed transport plan~\citep{courty2014domain}.
Among these approximations, Sliced optimal transport has been used to sample an embedding learned by an auto-encoder, by computing a flow between uniformly random samples and the image of encoded inputs~\citep{liutkus2019sliced}.

Regarding the Wasserstein barycenters we are interested in, they have been used for the task of learning a dictionary out of a set of probability measures~\citep{SHBMCCPS18}, for computing Wasserstein barycentric coordinates of probability measures~\citep{BPC16} or for metric learning~\citep{HBCCP19}.
These have been performed by automatic-differentiation of Wasserstein barycenters obtained through Sinkhorn iterations and non-linear optimization, and have thus been limited to small datasets, both due to speed and memory limitations.           
An adaptation of k-means clustering for optimal transport was proposed by~\citet{mi2018variational} and \citep{domazakis2020clustering}.  Backhoff-Veraguas et al. replaces maximum a posteriori (MAP) estimation or Bayesian model average, by computing Wasserstein barycenters of posterior distributions~\citep{backhoff2018bayesian} using a stochastic gradient descent scheme. In the context of reinforcement learning, Wasserstein barycenters are used by \cite{metelli2019propagating} as a way to regularize the update rule and offer robustness to uncertainty.
PCA in the Wasserstein space require the ability to compute Wasserstein barycenters ; they have been studied by~\citet{bigot2017geodesic} but could only be computed in 1-d where theory is simpler.
In the work of~\cite{dognin2019wasserstein}, Wasserstein barycenters are used for model ensembling, i.e., averaging the predictions of several models to build a more robust model.

In this work, we do not focus on a single application but instead provide the tools to efficiently approximate Wasserstein barycenters on 2-d regular grids.

\section{Learning Wasserstein barycenters}
This section describes our neural network and our proposed solution to train it in a scalable way.

\subsection{Proposed Model}
\begin{figure}[ht]
\centering
\centerline{\includegraphics[width=\columnwidth]{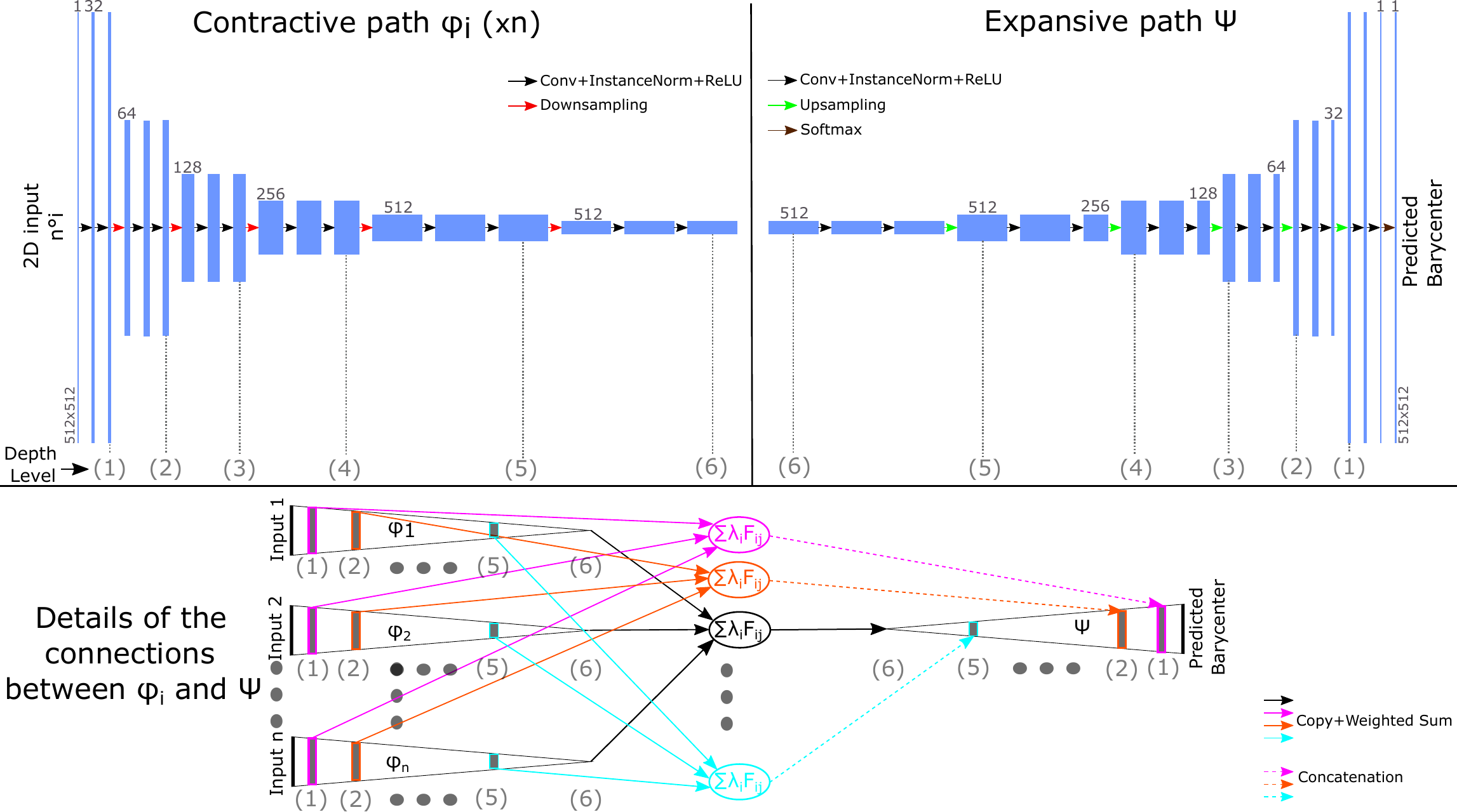}}
\caption{Our model is divided into $n$ contractive paths $\varphi_{i}$, sharing the same architecture and weights, and one expansive path $\psi$. Blue rectangles represent feature maps and arrows denote the different operations we use (see legend). At training time $n=2$, but by duplicating the contractive paths, we can adapt to the $n$ measures barycenter problem at test time, without needing to retrain the network}
\label{figure:model_schema}
\end{figure}

Our model aims at obtaining approximations of Wasserstein barycenters from $n\geq2$ probability measures $\{\mu_i\}_{i=1..n}$ discretized on $512\times512$ regular grids, and their corresponding barycentric weights $\{\lambda_{i}\}_{i=1..n}$. Based on the observation that the Sinkhorn algorithm is mainly made of successive convolutions, we propose to directly predict a Wasserstein barycenter through an end-to-end neural network approach, using a Deep Convolutional Neural Network (DCNN) architecture. This DCNN should be deep enough to allow accurate approximations but shallow enough to reduce its computational requirements. 

We propose a network consisting of $n$ contractive paths $\{\varphi_{i}\}_{i=1..n}$ and one expansive path $\psi$ (see Fig.~\ref{figure:model_schema}). Importantly enough, $n$ is not fixed and can vary at test time. In fact, the contractive paths are $n$ duplicates of the same path with the same architecture and sharing the same weights.
The $n$ contractive paths are made of successive blocks, each block consisting of two convolutional layers followed by a ReLU activation. We further add average pooling layers between each block in order to decrease the dimensionality. The expansive path is symetrically constructed, each block also being made of 2 convolutional layers with ReLU activations. To better invert average poolings, we use upsampling layers with nearest-neighbor interpolation. Finally, to recover an output probability distribution, we use a softmax activation at the end of the expansive path. All the 2D convolutions of our model use $3\times3$ kernels with a stride and a padding equal to $1$.
The architecture might look similar to the U-Net architecture introduced by~\citet{ronneberger2015u}, because of the nature of the contractive and expansive paths. However the similarities end here, since our architecture uses a variable number of contractive paths to handle multiple inputs. The connections we use from the contractive paths to the expansive path also highly differ: first, we take all the feature maps from each contractive path and not only a part of it as it is done in U-Net, and, second, we compute a weighted sum of all these activations using barycentric weights which results in a weighted feature map which is then symmetrically concatenate to the corresponding activations in the expansive path. Our network is deeper than U-Net and we do not use the same succession of layers nor the same downsampling and upsampling methods which are respectively max-pooling and up-convolutions in the case of U-Net. We also use Instance Normalization~\citep{ulyanov2016instance} which has empirically shown better results than Batch Normalization for our model. These normalization layers are placed before each ReLU activation. 

The connections going from the contractive paths to the expansive path are defined as follows: after each block in a contractive path $\varphi_{i}$ at depth level $j$, we take the resulting activations $\{F_{ij}\}_{i=1..n}$, compute their linear combination $F'_{j}=\sum_{i\in n}\lambda_{i} F_{ij}$, and concatenate it symmetrically to the corresponding activations in the expansive path (see figure~\ref{figure:model_schema}). 

\subsection{Training}

Our solution allows to generalize a network trained for computing the barycenter of two measures to an arbitrary number of input measures while remaining fast to train.

\paragraph{Variable number of inputs.} We expect our network to produce accurate results without constructing an explicit embedding whose existence remains uncertain~\citep{andoni2008earth}. However, a Euclidean embedding trivially generalizes to an arbitrary number of input measures. A key insight to our work is that, since contractive paths weights are shared, our network can be trained using only two contractive paths for the task of predicting Wasserstein barycenters of two probability measures. Once trained, contractive paths can be duplicated to the desired number of input measures. In practice, we found this procedure to yield accurate barycenters (see Sec.~\ref{sec:results}).

\paragraph{Loss function.}
Training the network requires comparing the predicted Wasserstein barycenter to a groundtruth Wasserstein barycenter. Ideally, such comparison should be performed via an optimal transport cost -- those are ideal to compare probability distributions. However, computing optimal transport costs on large training datasets would be intractable. Instead, we resort to a Kullback-Leibler divergence between the output distribution and the desired barycenter.

\paragraph{Optimizer.}
To optimize the model parameters, we use a stochastic gradient descent with warm restarts (SGDR) ~\citep{loshchilov2016sgdr}. The exact learning rate schedule we used for our models is shown in appendix \ref{appendix:training_details}, Fig. \ref{figure:evo_lr}.

\paragraph{Training data.}
We strive to train our network with datasets that would cover a wide range of input sketches. To achieve this, we built a dataset made of $100k$ pairs of $512\times512$ random shape contours with random barycentric weights and their corresponding 2D Wasserstein barycenter. Thereafter we call this dataset \emph{ContoursDS}.
The 2D shapes are generated in a Constructive Solid Geometry fashion: we randomly assemble primitives shapes using logical operators and detect contours in post-processing. A primitive corresponds to a filled ellipse, triangle, rectangle or a line. We assemble  these primitives together by using the classical boolean operators OR, AND, XOR, NOT.
To generate a shape, we initialize it with a random primitive. Then we combine it with another random primitive using a randomly chosen operator, and repeat this operation $d$ times ($0\leq d\leq 50$) where $d$ follows the probability distribution $d\sim \frac13 (\mathcal U({0,50}) + \mathcal N(0,2.5) +\mathcal N(50,2.5))$ which promotes simple ($d$ close to $0$) and complex ($d$ close to $50$) shapes.
Finally, we apply a Sobel filter to create contours. We thus create $10k$ random 2D shapes from which we build $100k$ Wasserstein barycenters.

We then use the GeomLoss library~\citep{geomloss} to build good approximations of Wasserstein barycenters in a reasonable time, with random pairs of inputs sampled from the set of generated shape contours. Given two 2D input distributions $\mu_{1}$ and $\mu_{2}$ with their corresponding barycentric weights $\lambda_{1}$ and $\lambda_{2} = 1-\lambda_1$, their barycenter $b^{*}$ can be found by minimizing: $b^{*}=\argmin_{b}\lambda_{1} S_{\epsilon}(b,\mu_{1})+\lambda_2 S_{\epsilon}(b,\mu_{2})$ where $S_{\epsilon}$ corresponds to the Sinkhorn divergence with quadratic ground metric, and $\epsilon$ the regularization parameter (we use $\epsilon=1\mathrm{e}{-4}$). We use a Lagrangian gradient descent scheme that first samples the distributions as $b=\sum_{j=1}^{N}b_{j}\delta_{x_{j}}$ and then performs a gradient descent using $x^{(k+1)}_{j}=x^{(k)}_{j}+\lambda_{1} v_{j}^{\mu_{1}}+\lambda_{2} v_{j}^{\mu_{2}}$ where $v_{j}^{\mu_{i}}$ is the displacement vector. This vector is computed as the gradient of the Sinkhorn divergence: $v_{j}^{\mu_{i}}=-\frac{1}{b_{j}}\nabla_{x_{j}}S_{\epsilon,p}(b,\mu_{i})$. These successive updates can be computationally expensive when inputs are large. 
To speed up computations, we use a linearized approach that performs a single descent step, starting from a uniform distribution. In practice, this allows to precompute one optimal transport map between a uniform distribution and each of the input measures in the database, and obtain approximate Wasserstein barycenters by using a weighted average of these transport maps. 

While it is quite obvious that our model trained with an application-specific dataset will produce the best results for this application, our model trained exclusively on \emph{ContoursDS} achieves results that are close enough and which can be in practice sufficient for the applications we consider. Figure \ref{fig:color_transfer_pentagon} demonstrates this in the context of color transfer. Interpolated color histograms are clearly best predicted by our model trained with the application-specific dataset; however the final color transfer results are very similar to the ones obtained using the histograms predicted by our model trained on ContourDS.

\section{Experimental Results}
\label{sec:results}

While our model is exclusively trained on our synthetic \emph{ContoursDS} dataset, at test time we also consider three additional datasets : the \emph{Quick, Draw!} dataset from \cite{quickdraw}, the \emph{Coil20} dataset \citep{nane1996columbia} and \emph{HistoDS}, a dataset of chrominance histograms.
The \emph{Quick, Draw!} dataset contains $50$ million grayscale drawings divided in multiple classes and has been created by asking users to draw with a mouse a given object in a limited time. The \emph{Coil20} is made of images of $20$ objects rotating on a black background and contains $72$ images per object for a total of $1440$ images. We rasterized these two datasets to $512\times512$ images. Finally, \emph{HistoDS} contains $100k$ $512\times 512$ chrominance histograms extracted from $10350$ Flickr\footnote{\scriptsize{\url{https://www.flickr.com/}}} images of various content and sizes obtained using the Flickr API.

\subsection{Two-way interpolation results}
\begin{figure}[ht]
\centering
\begin{tabular}{cccccc}
\raisebox{2\normalbaselineskip}[0pt][0pt]{\rotatebox[origin=c]{90}{\tiny Input 1}} & \includegraphics[width=0.12\columnwidth]{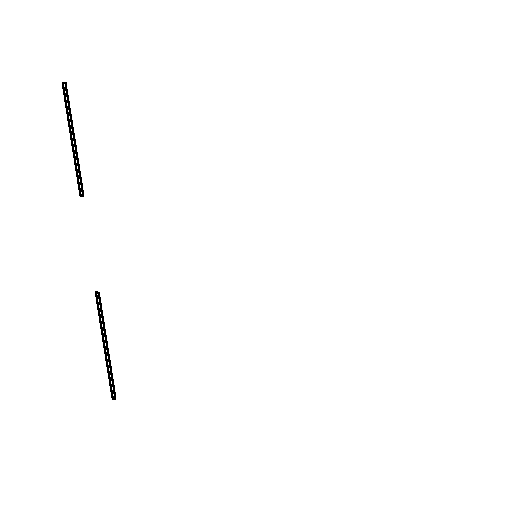} & \includegraphics[width=0.12\columnwidth]{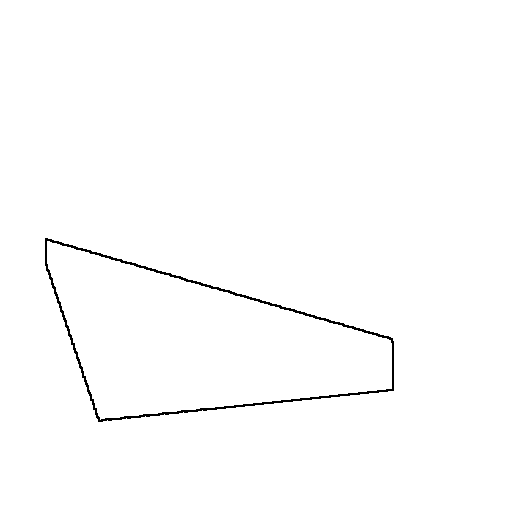} & \includegraphics[width=0.12\columnwidth]{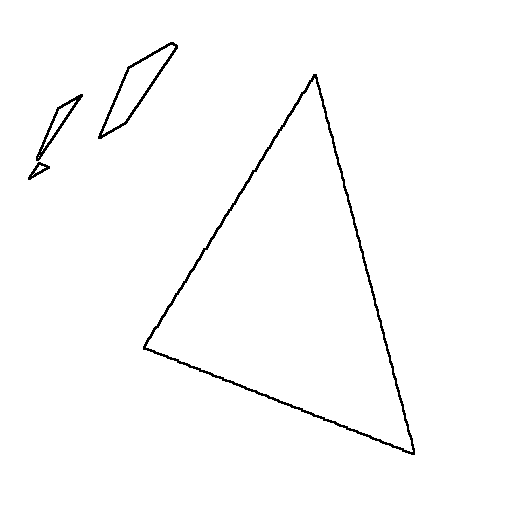} & \includegraphics[width=0.12\columnwidth]{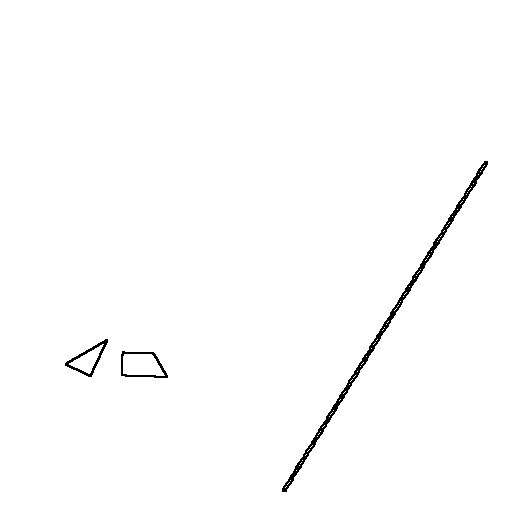} & \includegraphics[width=0.12\columnwidth]{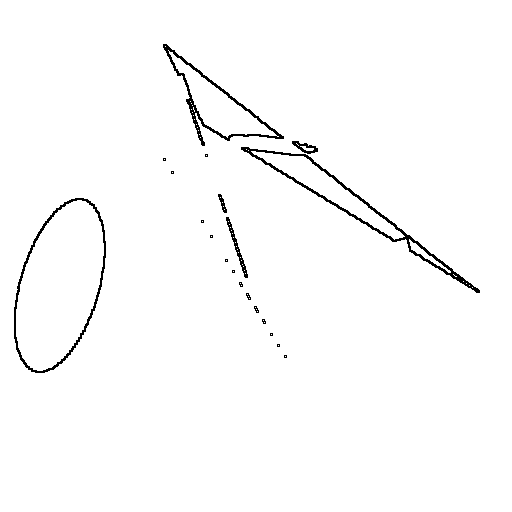} \\ 
 & {\tiny $\lambda_{1}=0.4382$} & {\tiny $\lambda_{1}=0.5863$} & {\tiny $\lambda_{1}=0.4586$} & {\tiny $\lambda_{1}=0.6573$} & {\tiny $\lambda_{1}=0.2567$} 
 \\
\raisebox{2\normalbaselineskip}[0pt][0pt]{\rotatebox[origin=c]{90}{\tiny Input 2}} & \includegraphics[width=0.12\columnwidth]{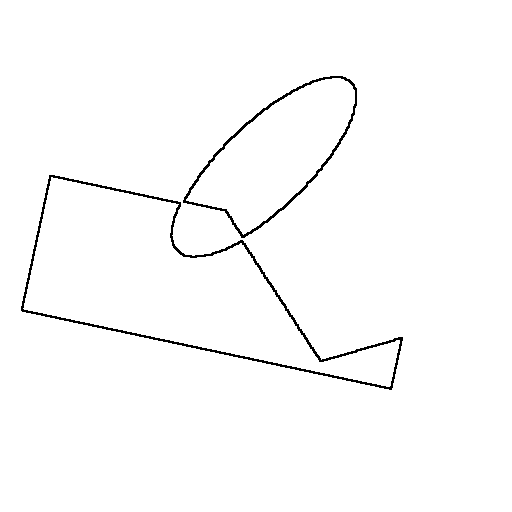} & \includegraphics[width=0.12\columnwidth]{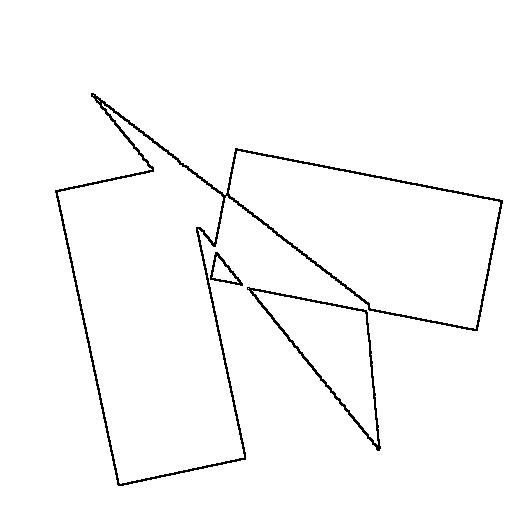} & \includegraphics[width=0.12\columnwidth]{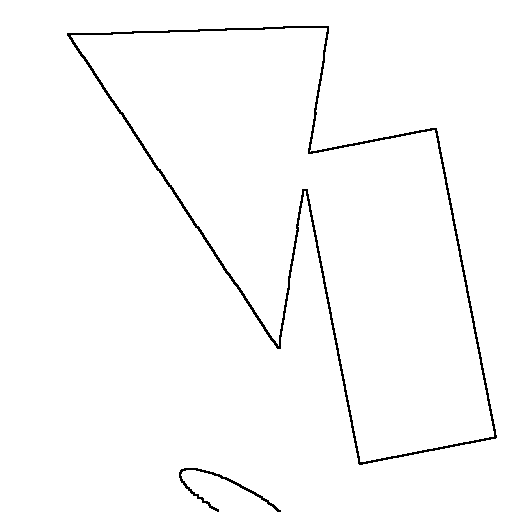} & \includegraphics[width=0.12\columnwidth]{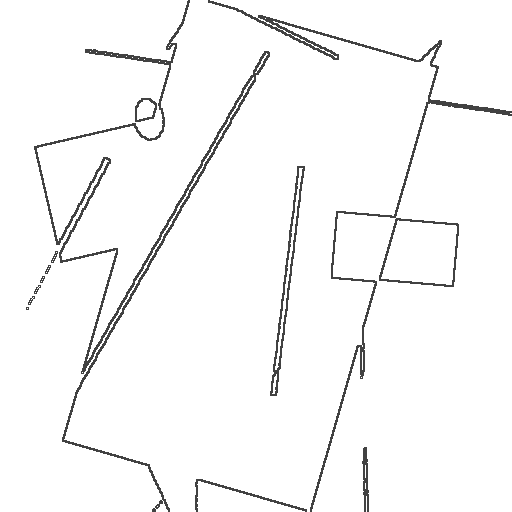} &  \includegraphics[width=0.12\columnwidth]{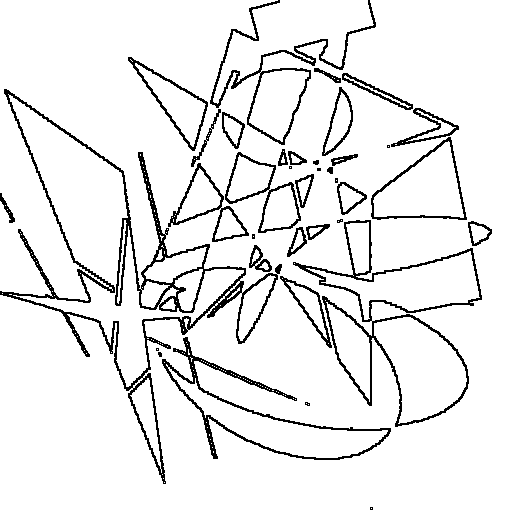} \\
 & {\tiny $\lambda_{2}=0.5618$} & {\tiny $\lambda_{2}=0.4137$} & {\tiny $\lambda_{2}=0.5414$} & {\tiny $\lambda_{2}=0.3127$} & {\tiny $\lambda_{2}=0.7433$}
 \\
\raisebox{2\normalbaselineskip}[0pt][0pt]{\rotatebox[origin=c]{90}{\tiny GeomLoss\hspace{-0.5cm}}} & \includegraphics[width=0.12\columnwidth]{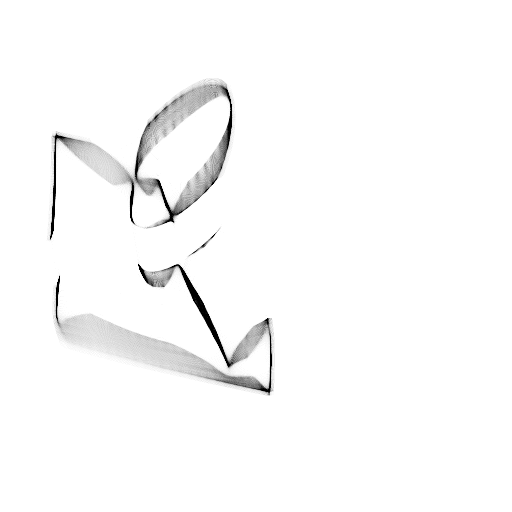} & \includegraphics[width=0.12\columnwidth]{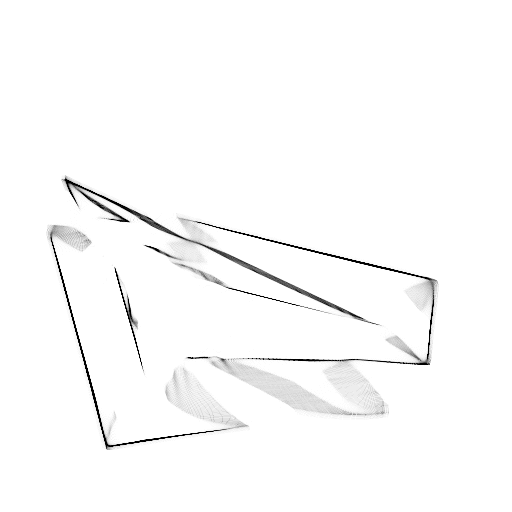} & \includegraphics[width=0.12\columnwidth]{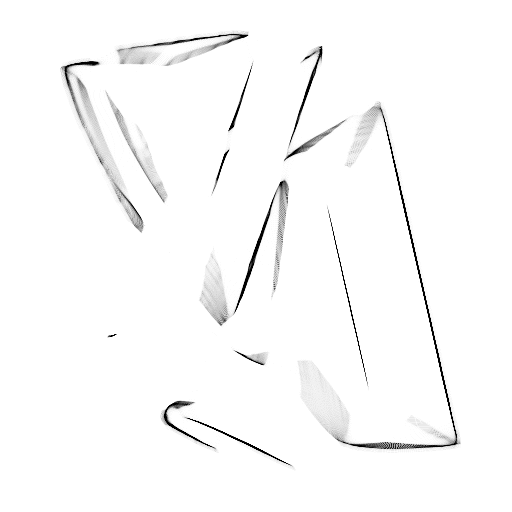} &  \includegraphics[width=0.12\columnwidth]{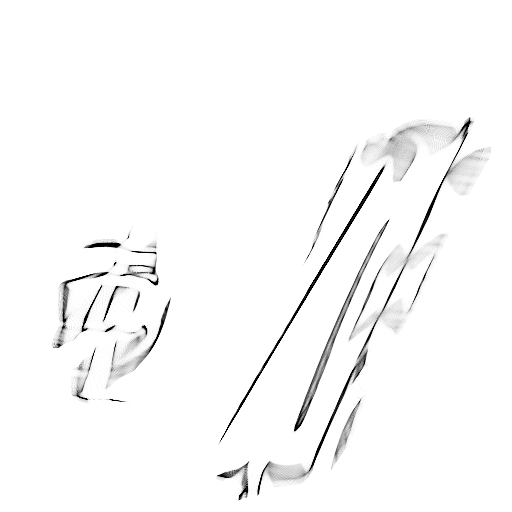} & \includegraphics[width=0.12\columnwidth]{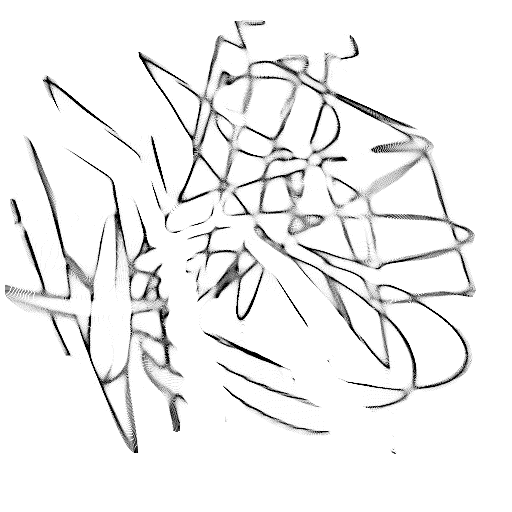} \\

\raisebox{2\normalbaselineskip}[0pt][0pt]{\rotatebox[origin=c]{90}{\tiny Our Model}} & \includegraphics[width=0.12\columnwidth]{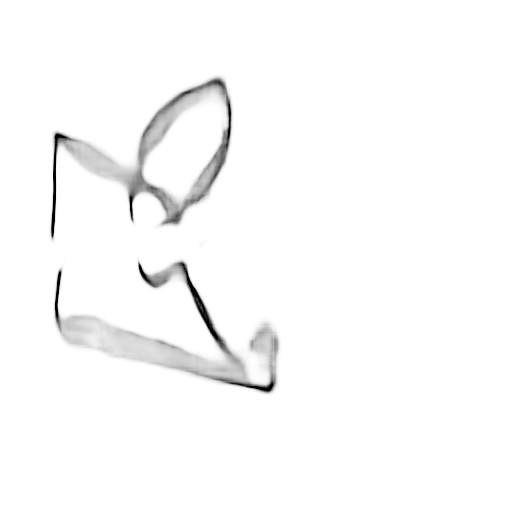} & \includegraphics[width=0.12\columnwidth]{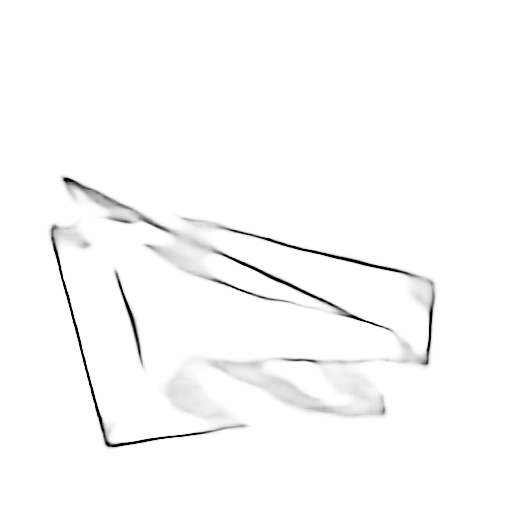} & \includegraphics[width=0.12\columnwidth]{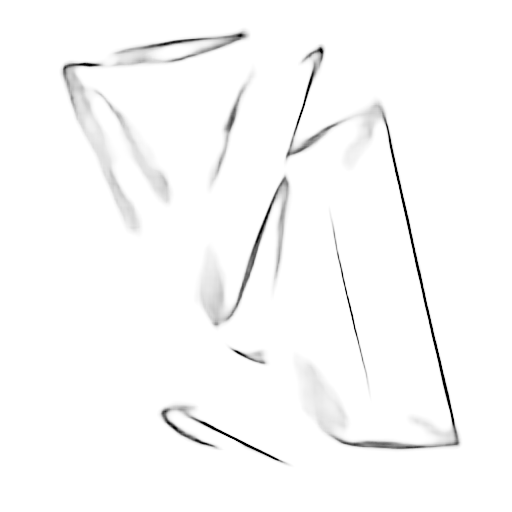} & 
\includegraphics[width=0.12\columnwidth]{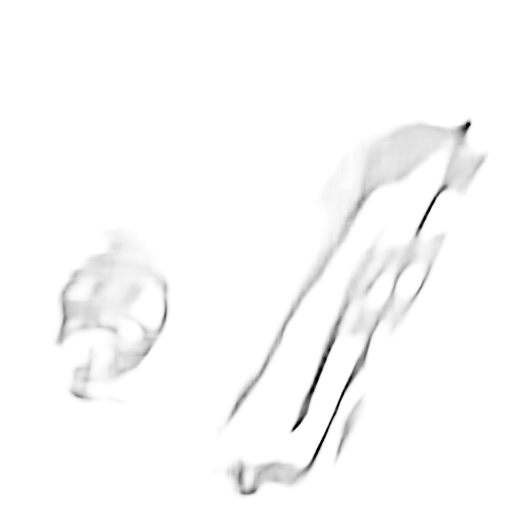} & \includegraphics[width=0.12\columnwidth]{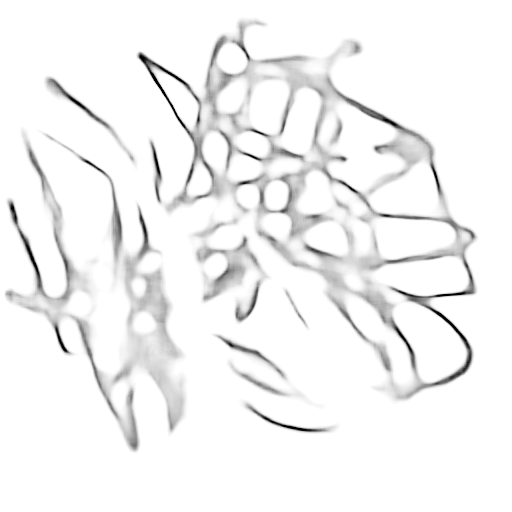} \\
 
 \raisebox{2\normalbaselineskip}[0pt][0pt]{\rotatebox[origin=c]{90}{\tiny Linear Program}} & \includegraphics[width=0.12\columnwidth]{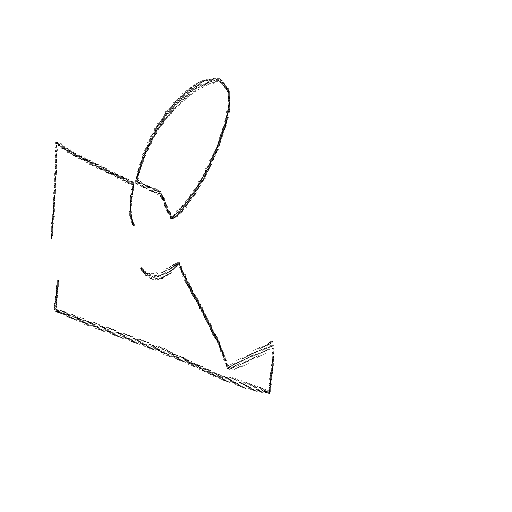} & \includegraphics[width=0.12\columnwidth]{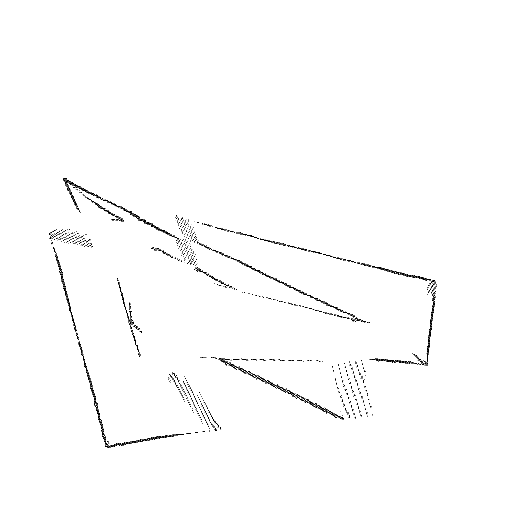} & \includegraphics[width=0.12\columnwidth]{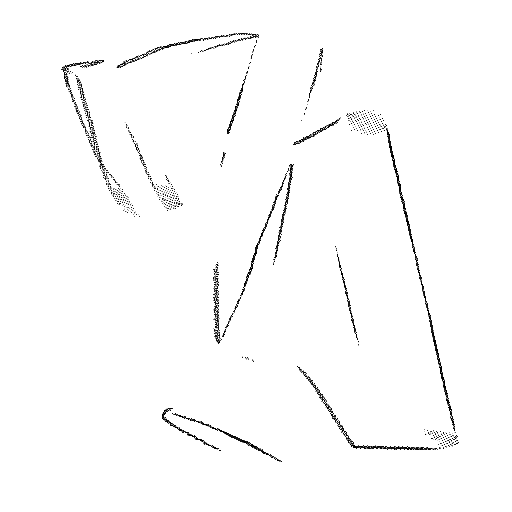} & 
 \includegraphics[width=0.12\columnwidth]{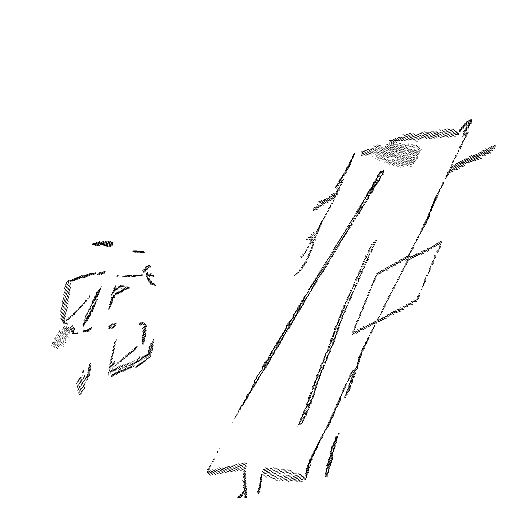} &  \includegraphics[width=0.12\columnwidth]{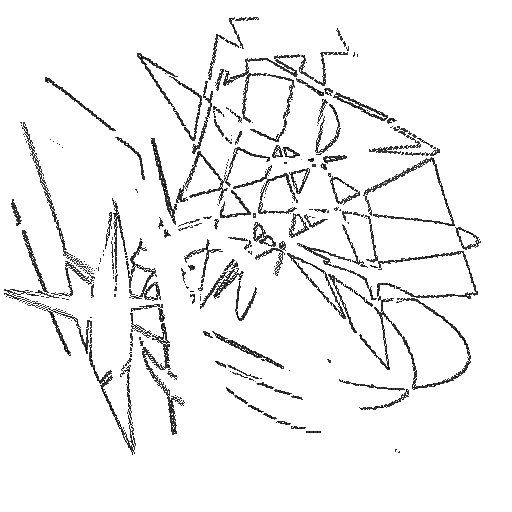} \\

 \raisebox{2\normalbaselineskip}[0pt][0pt]{\rotatebox[origin=c]{90}{\tiny Regularized}} & \includegraphics[width=0.12\columnwidth]{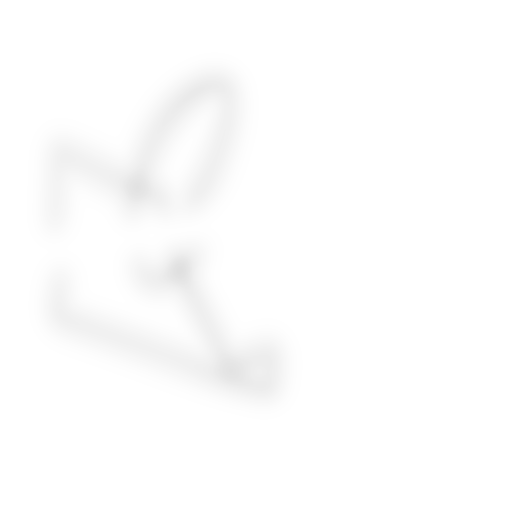} & \includegraphics[width=0.12\columnwidth]{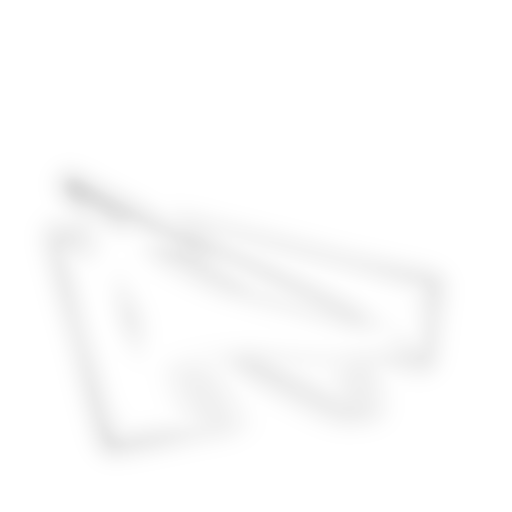} & \includegraphics[width=0.12\columnwidth]{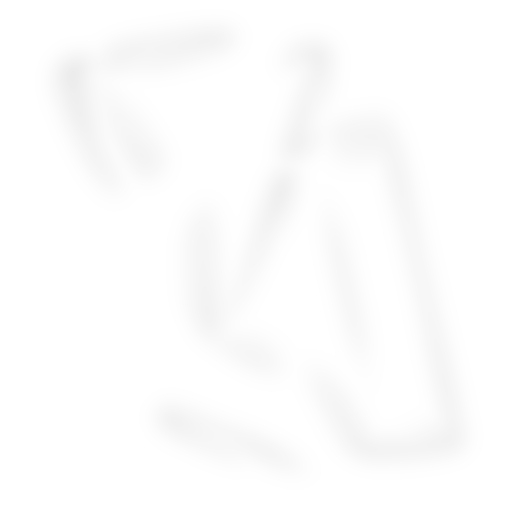} & \includegraphics[width=0.12\columnwidth]{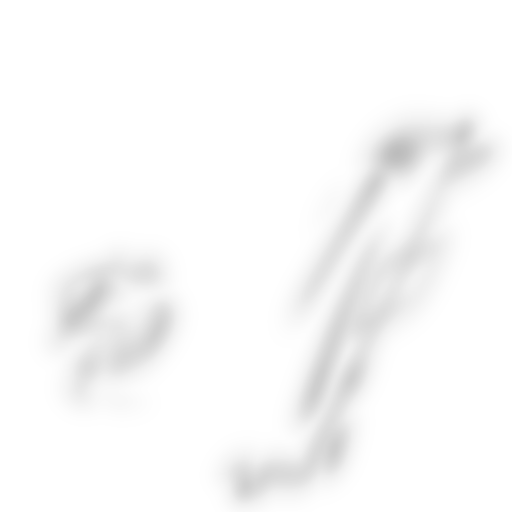} & \includegraphics[width=0.12\columnwidth]{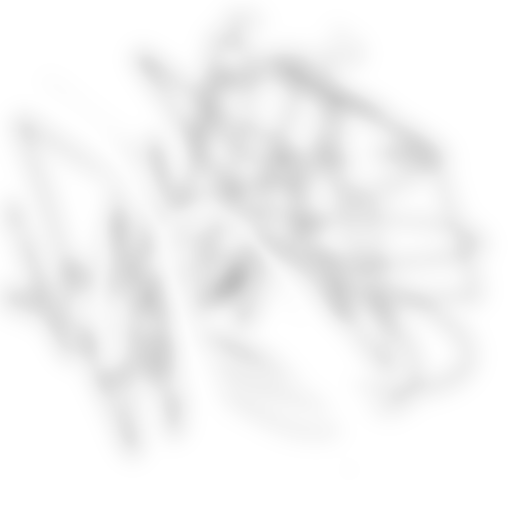} \\

 \raisebox{2\normalbaselineskip}[0pt][0pt]{\rotatebox[origin=c]{90}{\tiny Radon}} & \includegraphics[width=0.12\columnwidth]{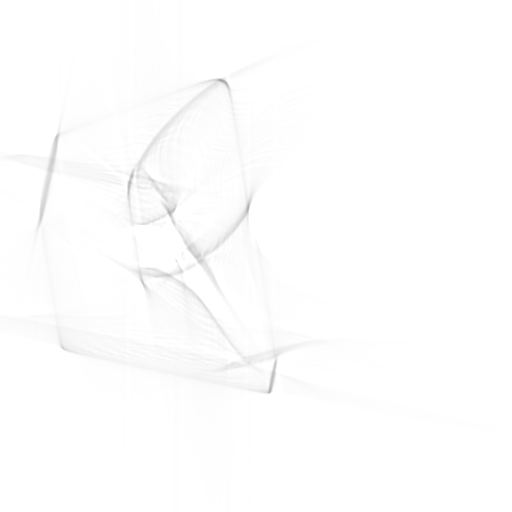} & \includegraphics[width=0.12\columnwidth]{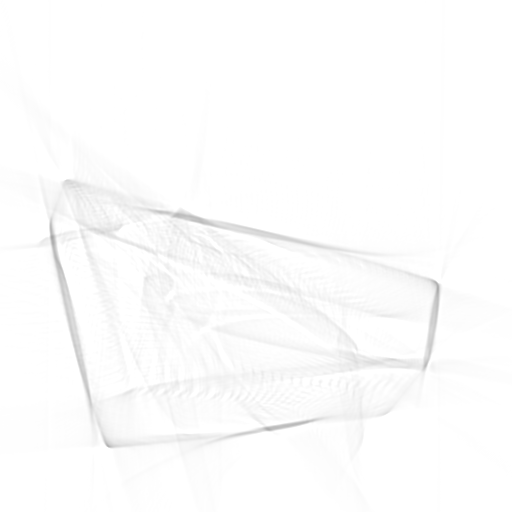} & \includegraphics[width=0.12\columnwidth]{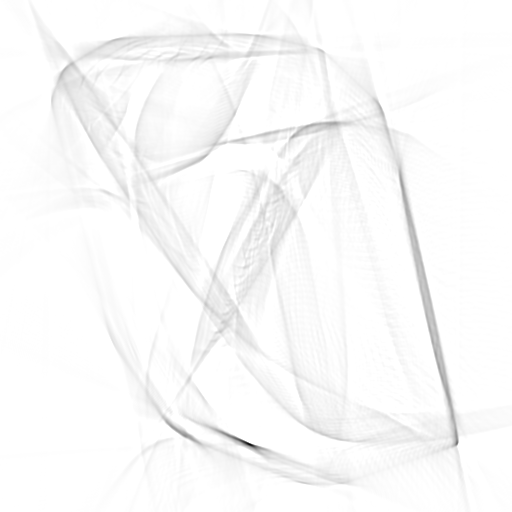} & 
 \includegraphics[width=0.12\columnwidth]{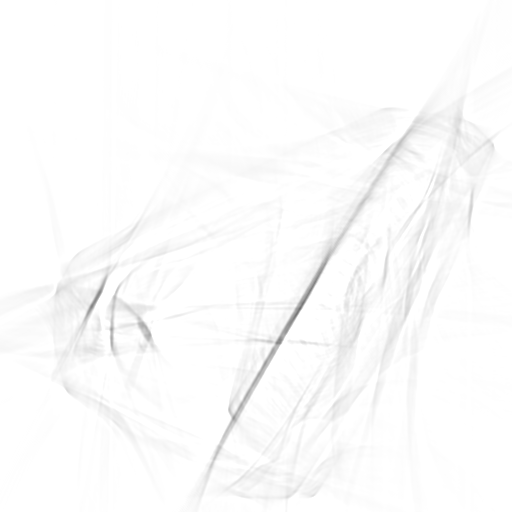} &  \includegraphics[width=0.12\columnwidth]{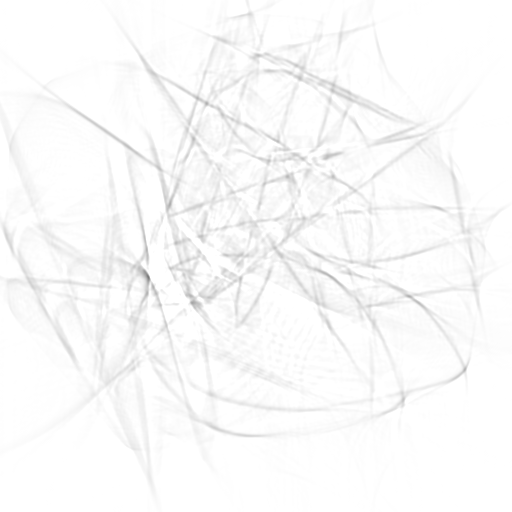} 
\end{tabular}
\caption{We illustrate typical results and comparisons to GeomLoss~\protect\citep{geomloss}, a linear program via a network simplex~\protect\citep{BPPH11}, regularized barycenters computed in log-domain (see for instance ~\citet{peyre2019computational}) with a regularization parameter of $1\mathrm{e}{-3}$ and Radon barycenters~\protect\citep{bonneel2015sliced}}
\label{figure:datasets_samples}
\end{figure}

In Fig.~\ref{figure:datasets_samples}, we show a visual comparison between barycenters obtained with Geomloss and our method. Wassertein barycenters are taken from the test dataset and the corresponding predictions are shown. We also compare these results to classical approaches (linear program, regularized barycenters) and to another approximation method known as Radon barycenters ~\protect\citep{bonneel2015sliced}. 

\begin{figure}[ht]
\centering
\begin{tabular}{ccc|ccc}

{\tiny Inputs} & {\tiny GeomLoss} & {\tiny Our Model} & {\tiny Inputs} & {\tiny GeomLoss} & {\tiny Our Model} \\
 
\includegraphics[width=0.08\columnwidth]{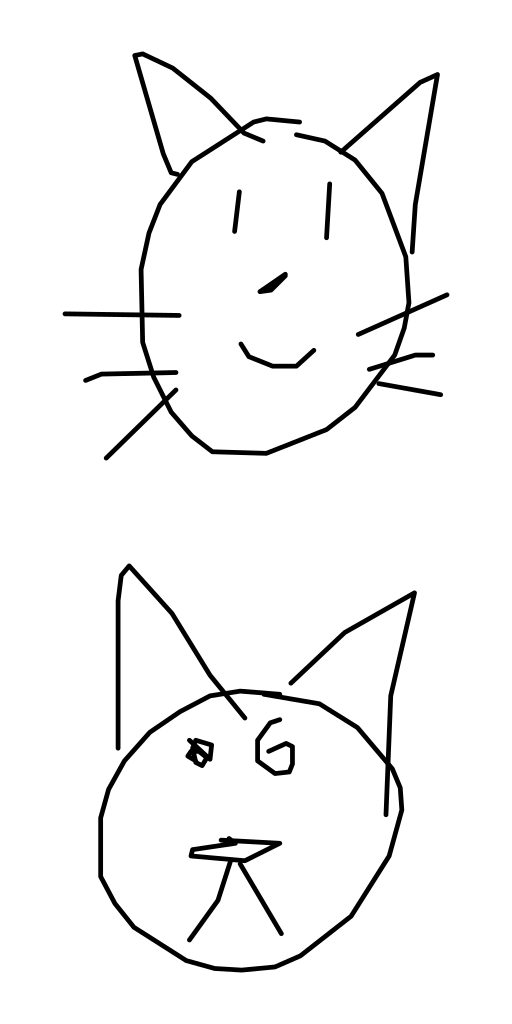} &
\includegraphics[width=0.16\columnwidth]{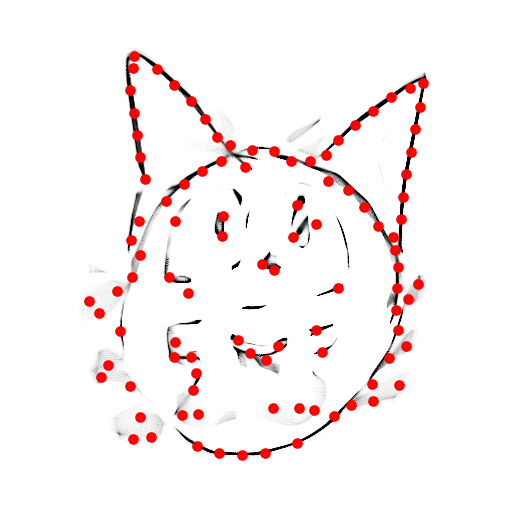} & \includegraphics[width=0.16\columnwidth]{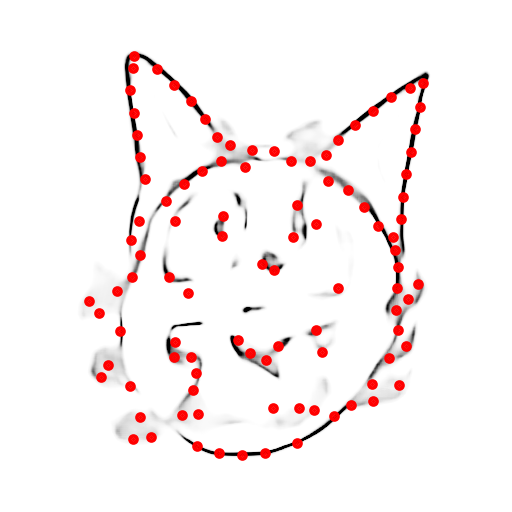} &
\includegraphics[width=0.08\columnwidth]{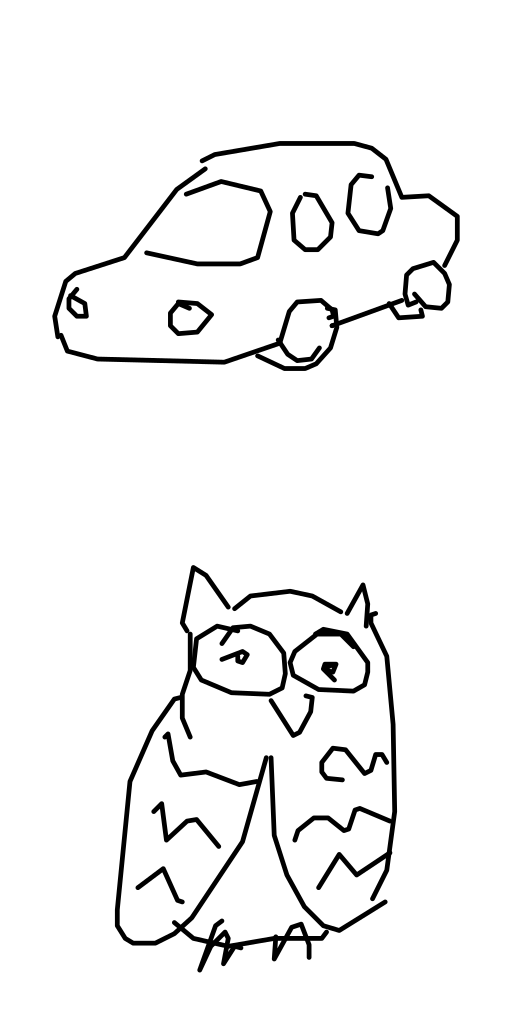} & \includegraphics[width=0.16\columnwidth]{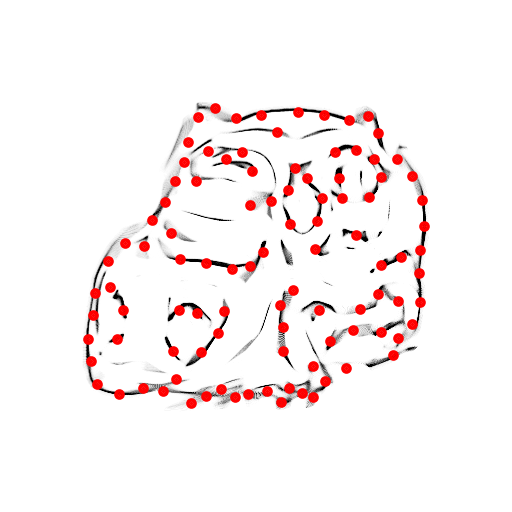} & \includegraphics[width=0.16\columnwidth]{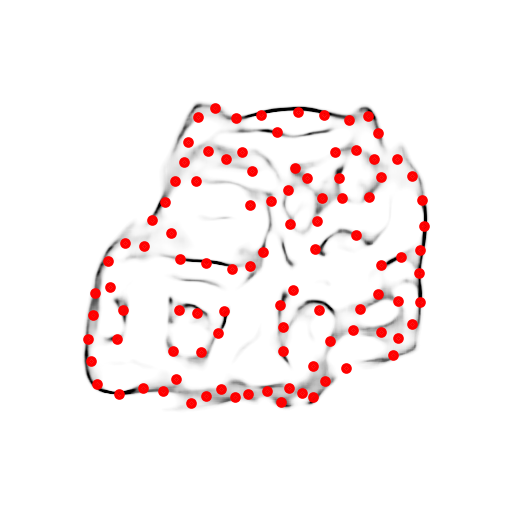} \\

\end{tabular}
\caption{We superimpose the centroids ($\lambda_{1}=\lambda_{2}=0.5$) found by the method of ~\citet{claici2018stochastic} (in red) using 100 Dirac masses over the ones computed by GeomLoss and by our model, on images from the \emph{Quick, Draw!} dataset. The solution of ~\citet{claici2018stochastic} was found within 37 hours of computation}
\label{figure:comparison_with_claici}
\end{figure}

To further visually assess that the barycenters we are approximating are close to the exact ones, we also present a comparison with the method of ~\citet{claici2018stochastic} in Fig.~\ref{figure:comparison_with_claici}. Input distributions are taken from the \emph{Quick, Draw!} dataset. 

\begin{figure}[ht]
\centering

\begin{tabular}{@{}c@{}c@{}c@{}c@{}c@{}c@{}c|@{}c@{}c@{}c@{}c@{}c@{}c@{}}
\raisebox{\normalbaselineskip}[0pt][0pt]{\rotatebox[origin=c]{90}{\tiny GeomLoss}} & \includegraphics[width=0.075\columnwidth]{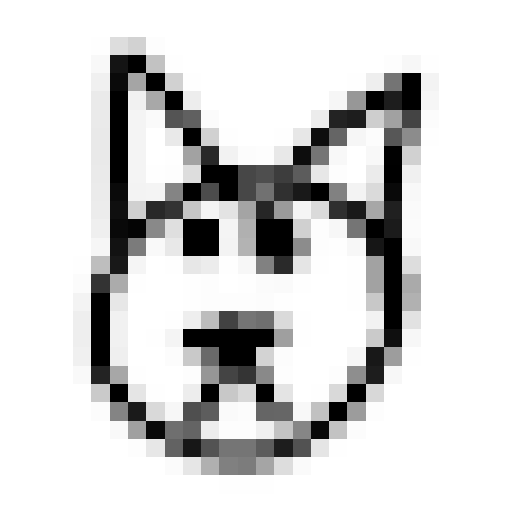} 
& \includegraphics[width=0.075\columnwidth]{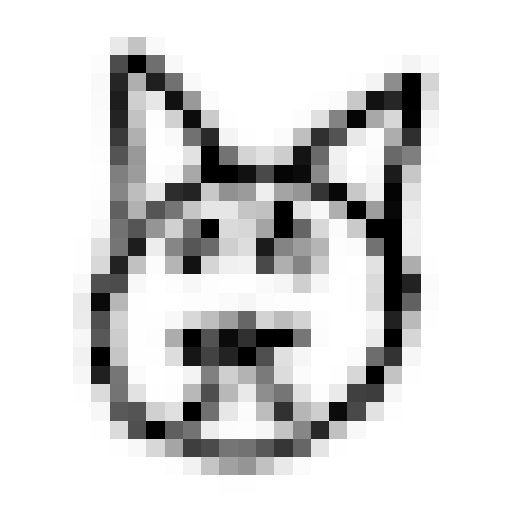} 
& \includegraphics[width=0.075\columnwidth]{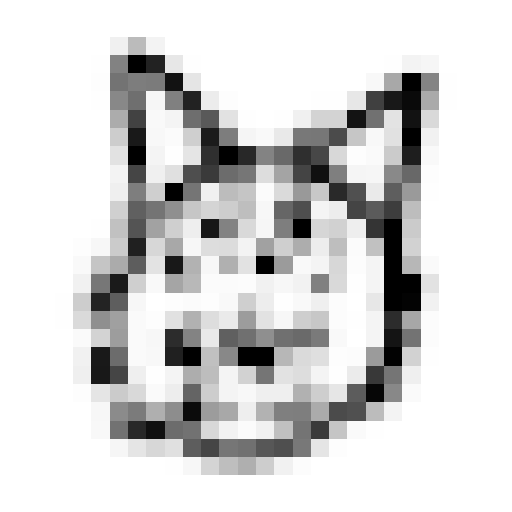} 
& \includegraphics[width=0.075\columnwidth]{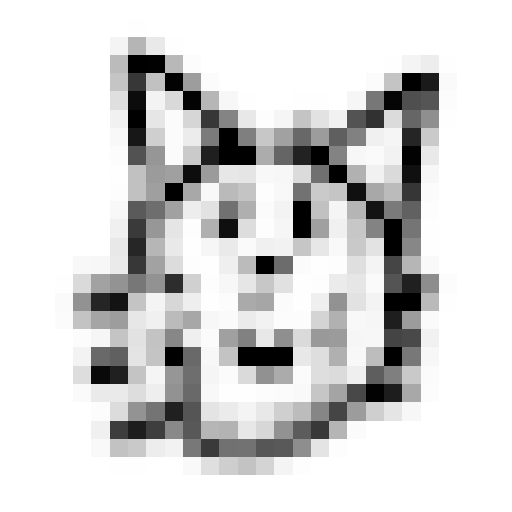} 
& \includegraphics[width=0.075\columnwidth]{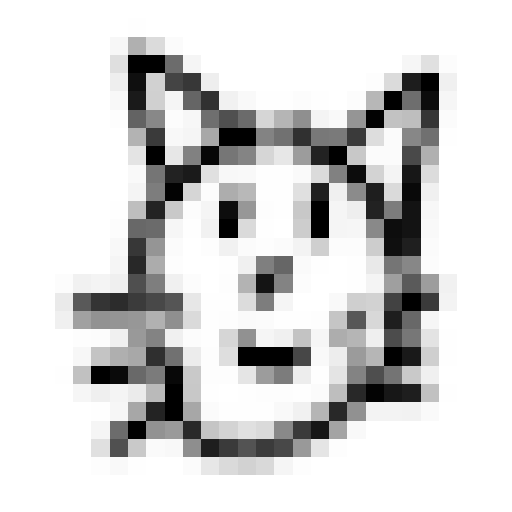}
& \includegraphics[width=0.075\columnwidth]{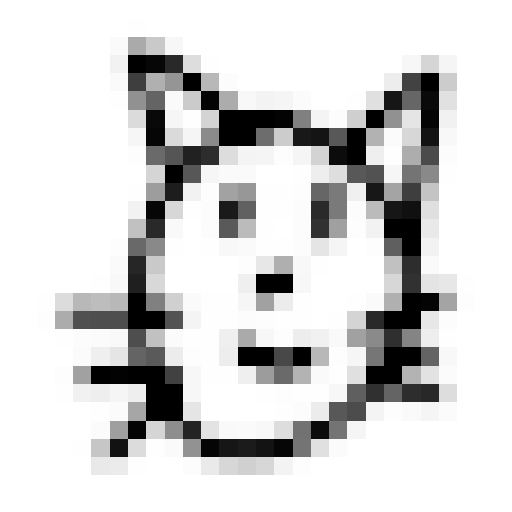}  
& \includegraphics[width=0.075\columnwidth]{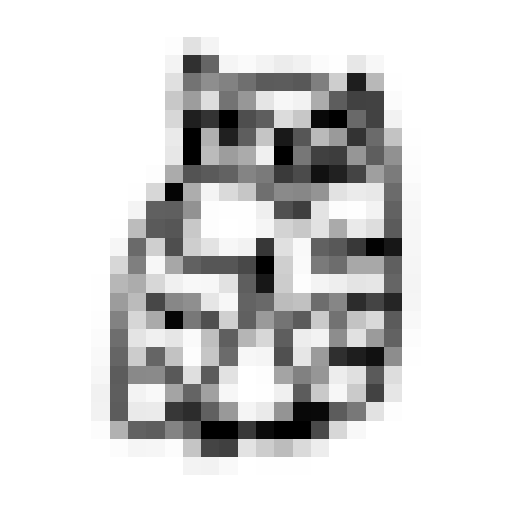} 
& \includegraphics[width=0.075\columnwidth]{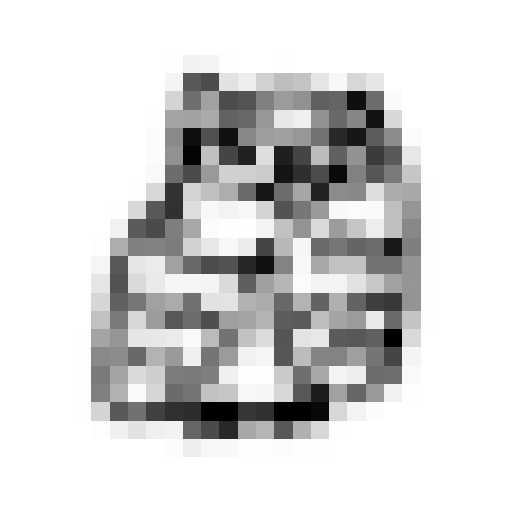} 
& \includegraphics[width=0.075\columnwidth]{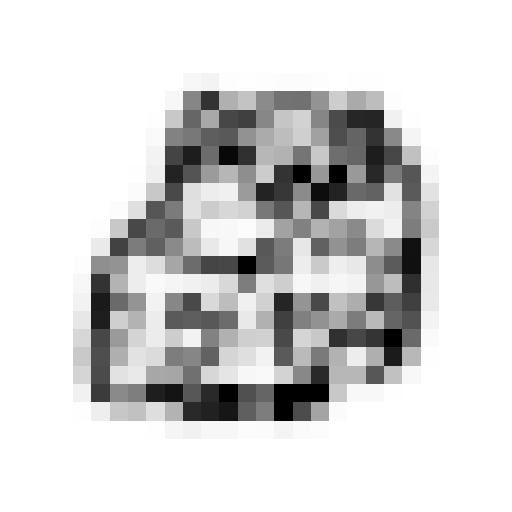} 
& \includegraphics[width=0.075\columnwidth]{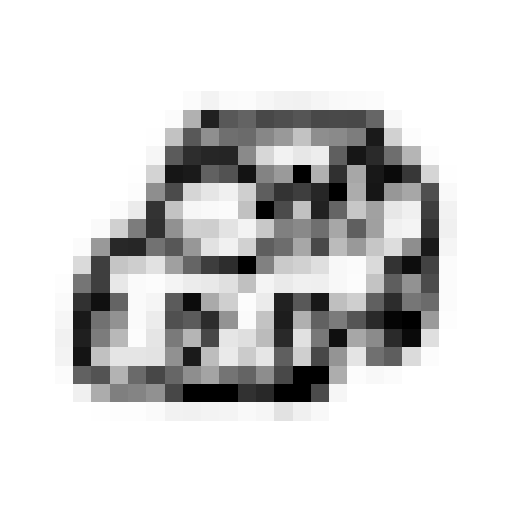} 
& \includegraphics[width=0.075\columnwidth]{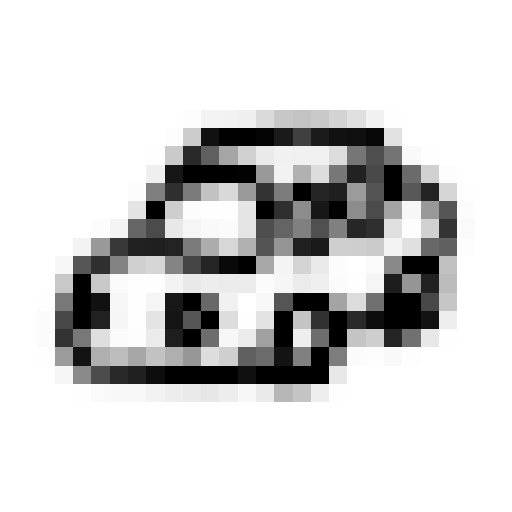} 
& \includegraphics[width=0.075\columnwidth]{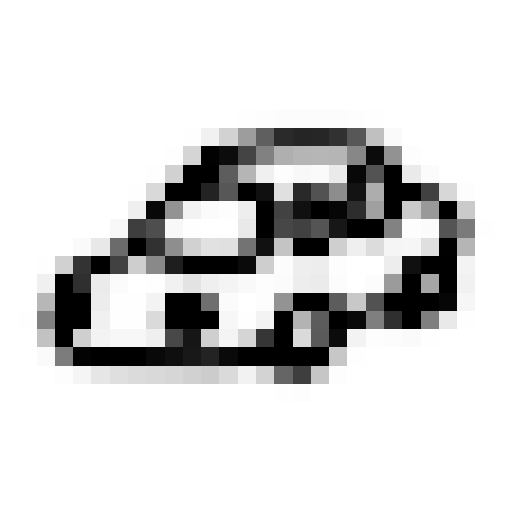} \\

\raisebox{\normalbaselineskip}[0pt][0pt]{\rotatebox[origin=c]{90}{\tiny Ours}} 
& \includegraphics[width=0.075\columnwidth]{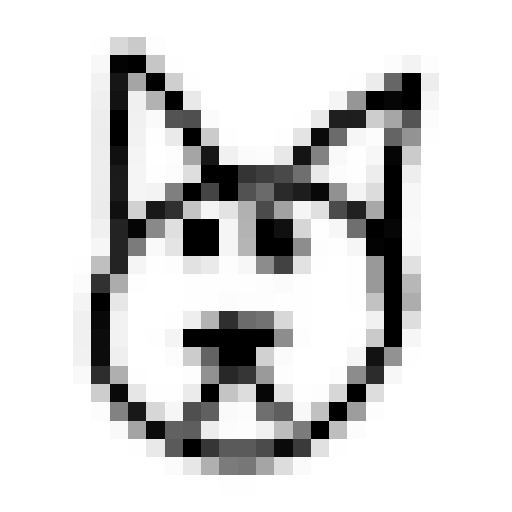} 
& \includegraphics[width=0.075\columnwidth]{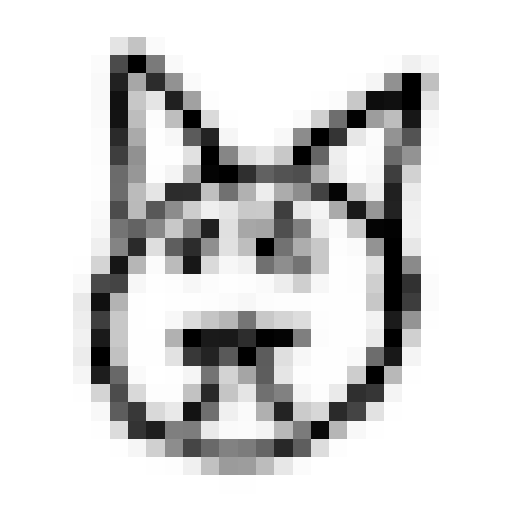} 
& \includegraphics[width=0.075\columnwidth]{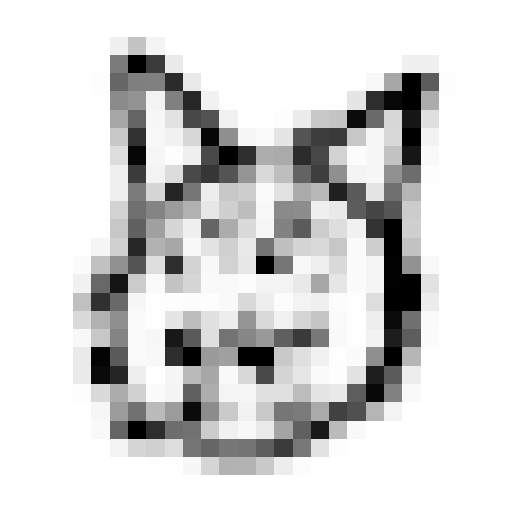} 
& \includegraphics[width=0.075\columnwidth]{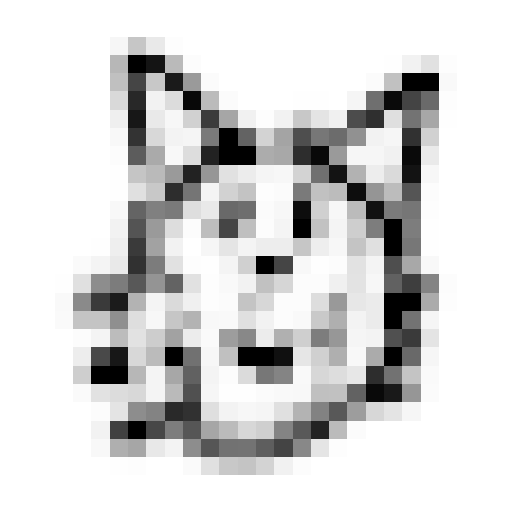} 
& \includegraphics[width=0.075\columnwidth]{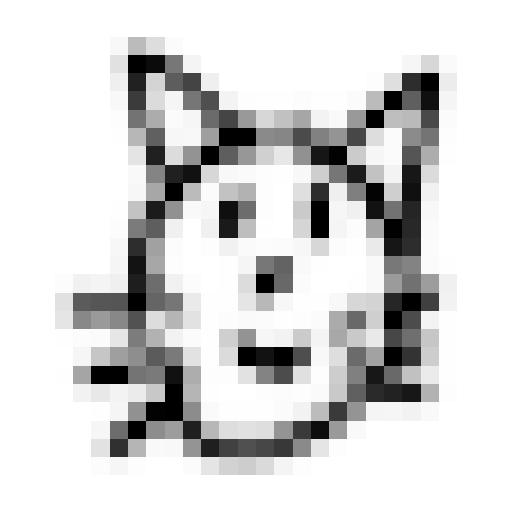} 
& \includegraphics[width=0.075\columnwidth]{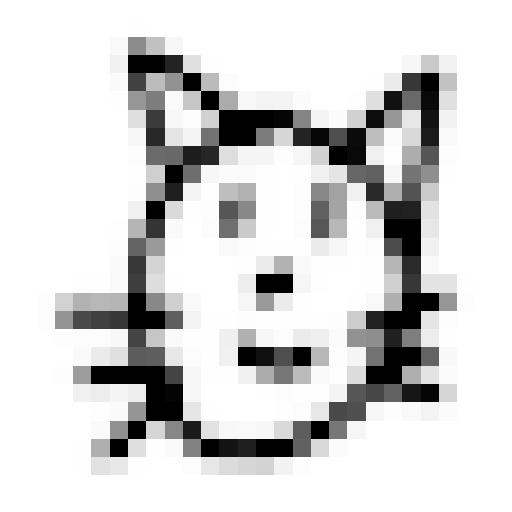} 
& \includegraphics[width=0.075\columnwidth]{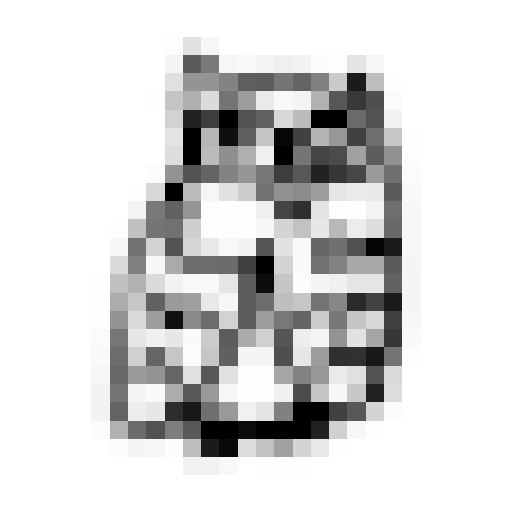} 
& \includegraphics[width=0.075\columnwidth]{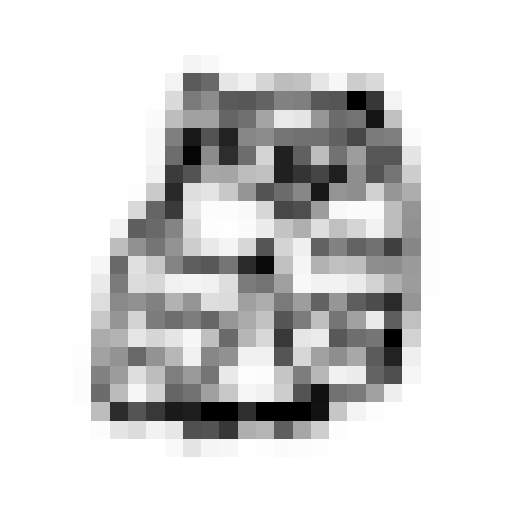} 
& \includegraphics[width=0.075\columnwidth]{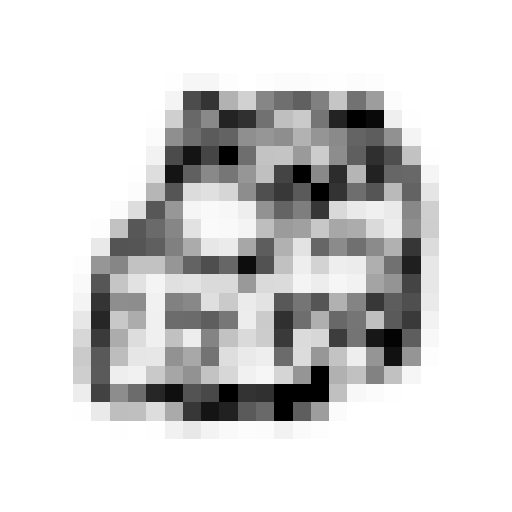} 
& \includegraphics[width=0.075\columnwidth]{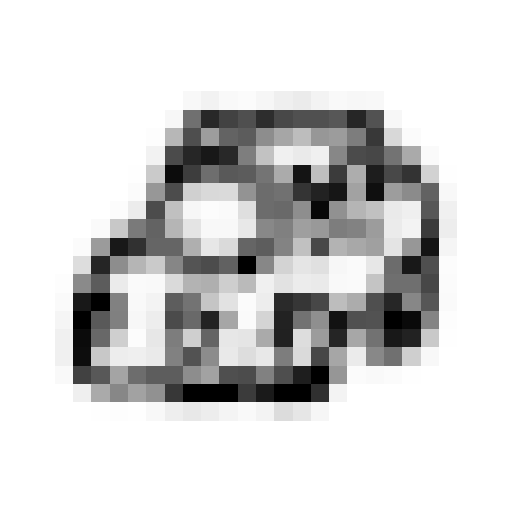} 
& \includegraphics[width=0.075\columnwidth]{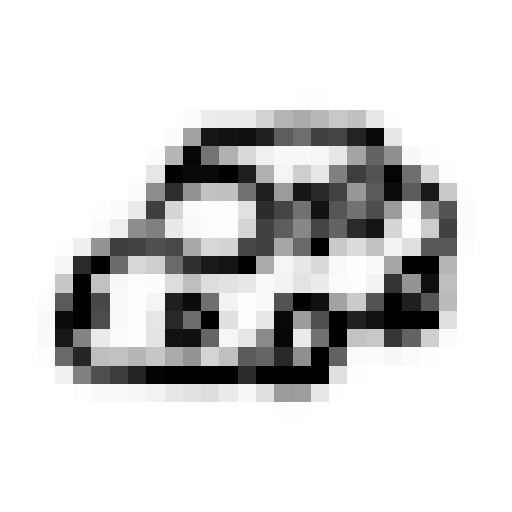} 
& \includegraphics[width=0.075\columnwidth]{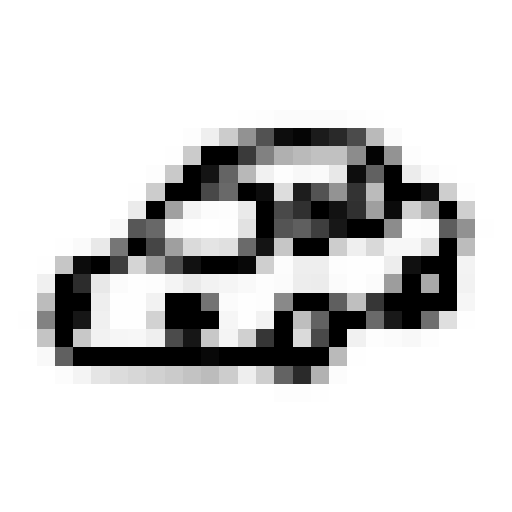} \\

\raisebox{\normalbaselineskip}[0pt][0pt]{\rotatebox[origin=c]{90}{\tiny DWE}} 
& \includegraphics[width=0.075\columnwidth]{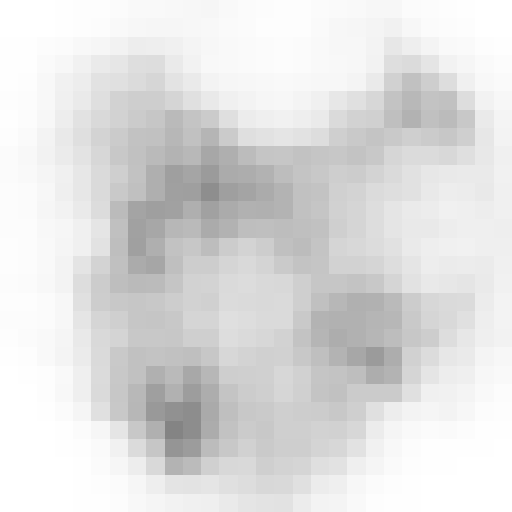} 
& \includegraphics[width=0.075\columnwidth]{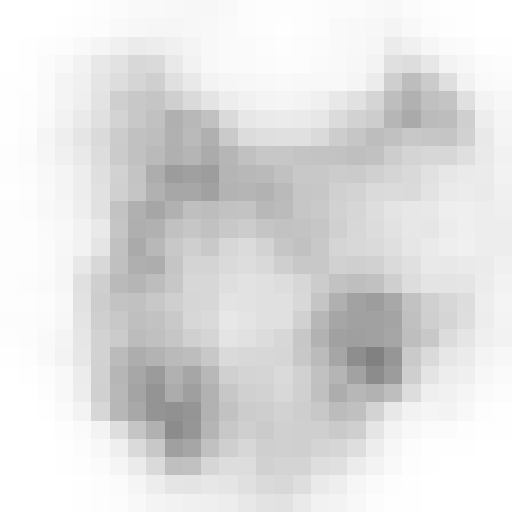} 
& \includegraphics[width=0.075\columnwidth]{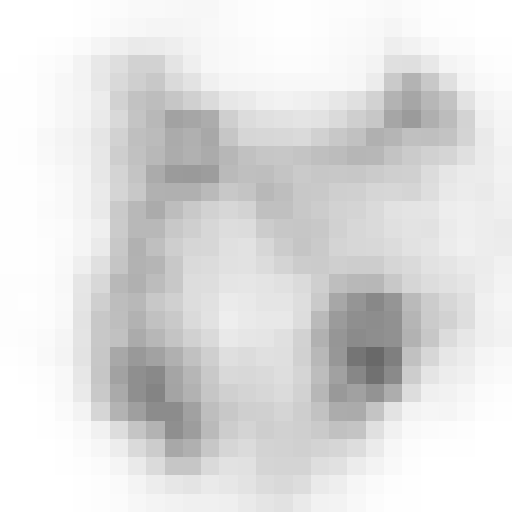} 
& \includegraphics[width=0.075\columnwidth]{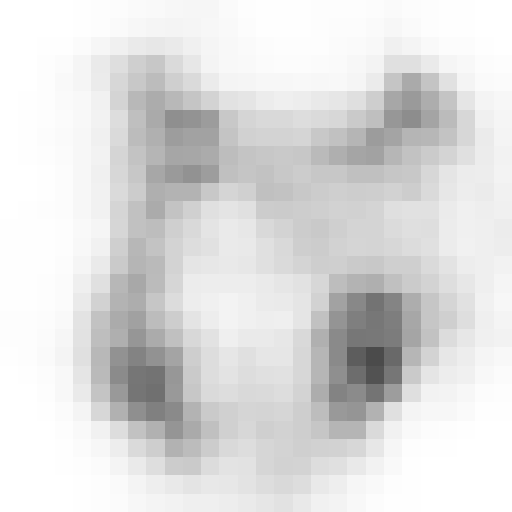} 
& \includegraphics[width=0.075\columnwidth]{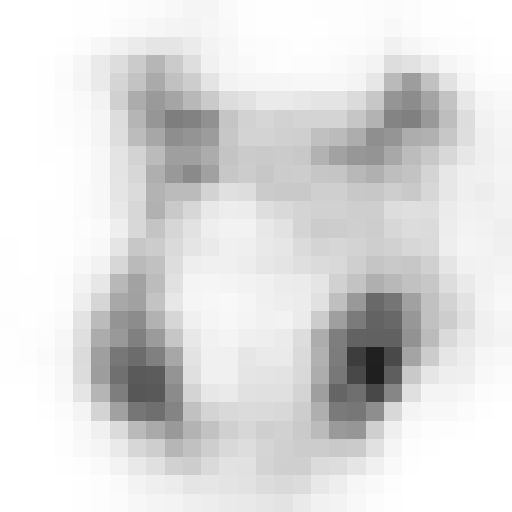} 
& \includegraphics[width=0.075\columnwidth]{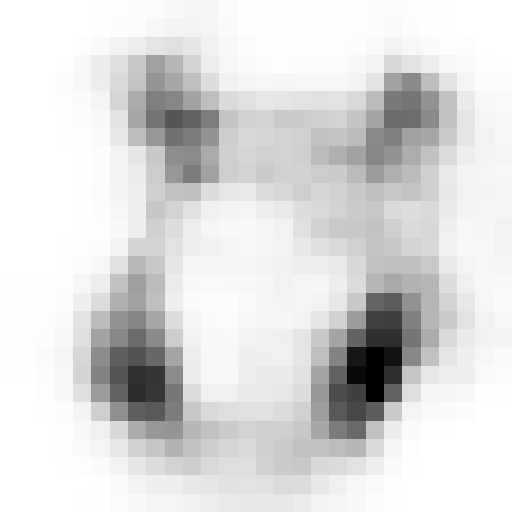} 
& \includegraphics[width=0.075\columnwidth]{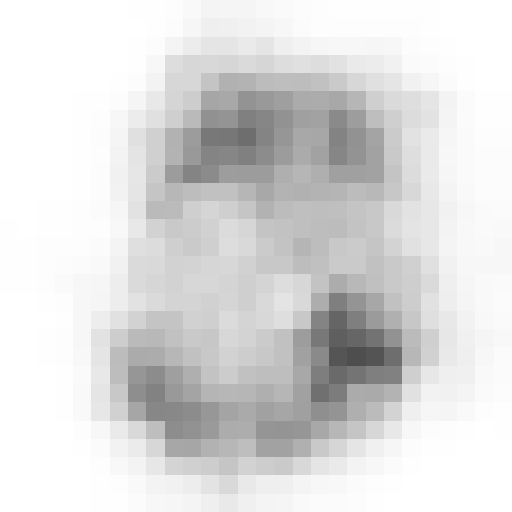} 
& \includegraphics[width=0.075\columnwidth]{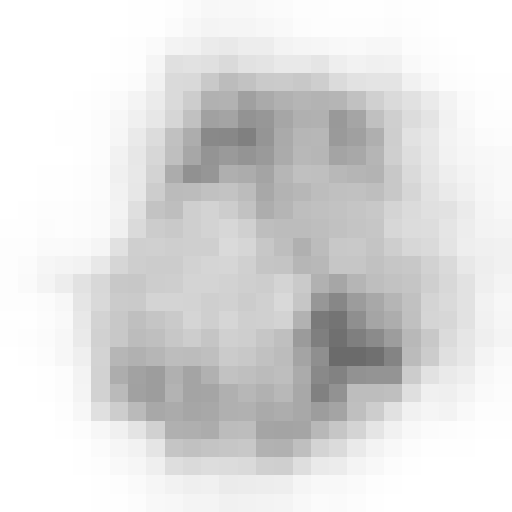} 
& \includegraphics[width=0.075\columnwidth]{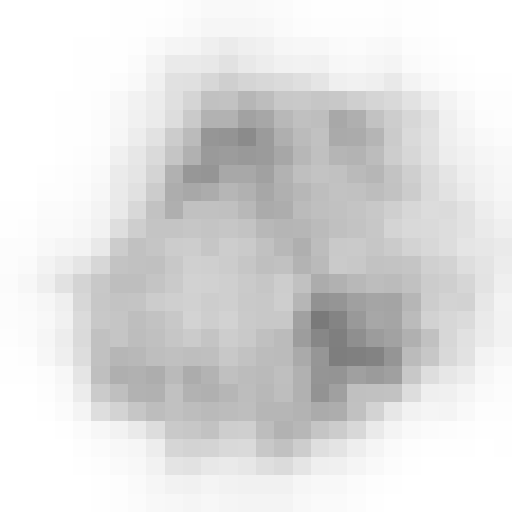} 
& \includegraphics[width=0.075\columnwidth]{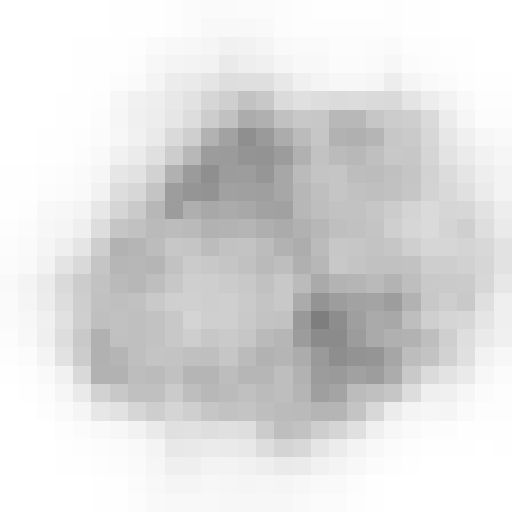} 
& \includegraphics[width=0.075\columnwidth]{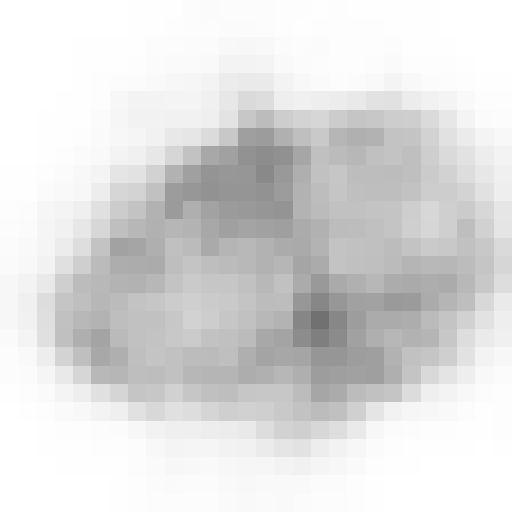} 
& \includegraphics[width=0.075\columnwidth]{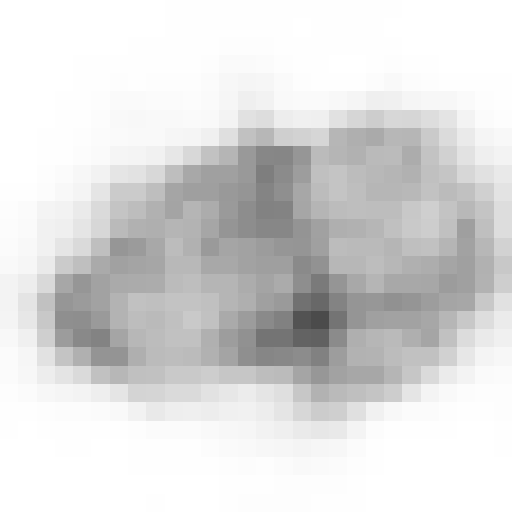} \\
\end{tabular}

\caption{Interpolations between two $28\times28$ images from the \emph{Quick, Draw!} dataset using Geomloss, our model and the original Deep Wasserstein Embedding (DWE) method from ~\citep{courty2017learning}. Our model directly considers $512\times512$ inputs and its results are downsampled from $512\times512$ to $28\times28$}
\label{figure:28x28_dwe_comparison}
\end{figure}

\begin{figure}[ht]
\centering

\begin{tabular}{@{}c@{}c@{}c@{}c@{}c@{}c@{}c|@{}c@{}c@{}c@{}c@{}c@{}c@{}}
\raisebox{\normalbaselineskip}[0pt][0pt]{\rotatebox[origin=c]{90}{\tiny GeomLoss}}
& \includegraphics[width=0.075\columnwidth]{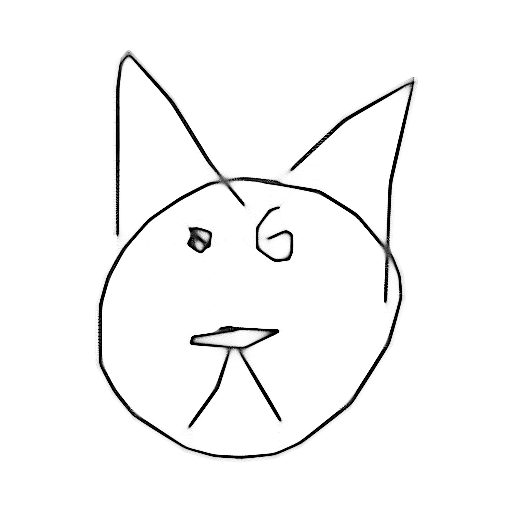} 
& \includegraphics[width=0.075\columnwidth]{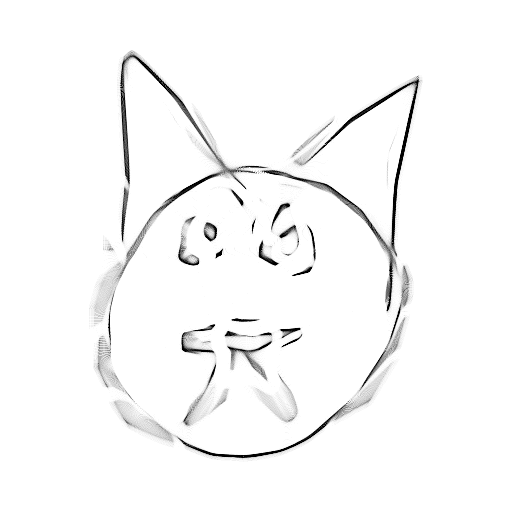} 
& \includegraphics[width=0.075\columnwidth]{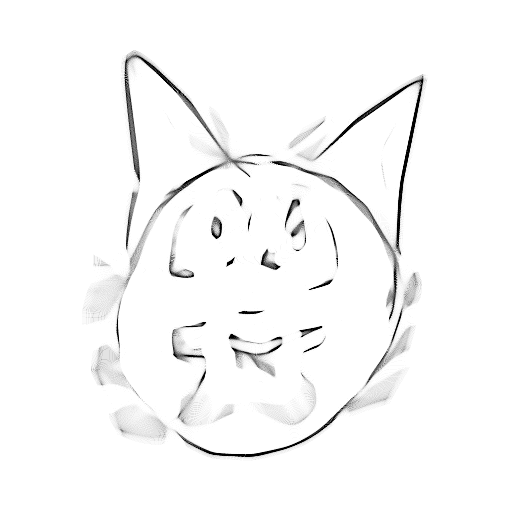} 
& \includegraphics[width=0.075\columnwidth]{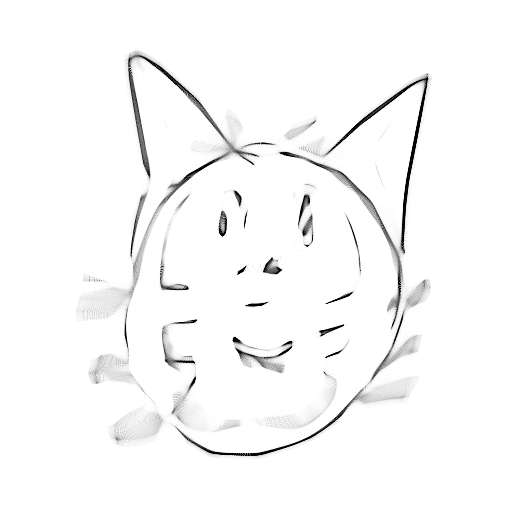} 
& \includegraphics[width=0.075\columnwidth]{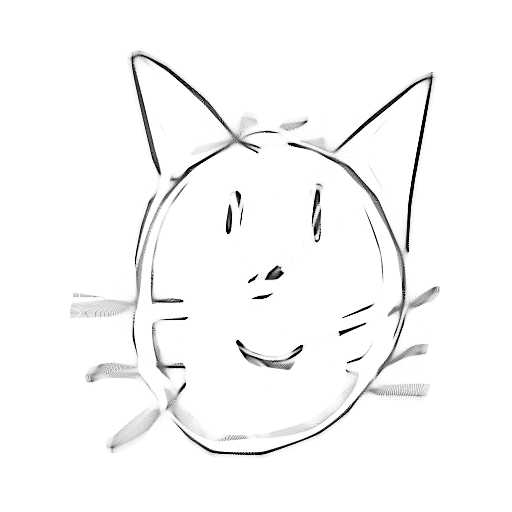} 
& \includegraphics[width=0.075\columnwidth]{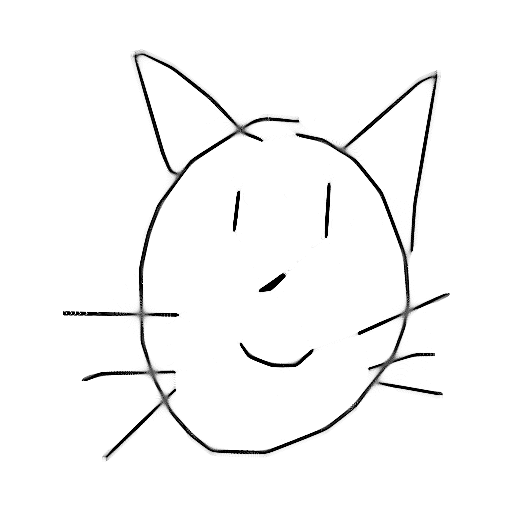} 
& \includegraphics[width=0.075\columnwidth]{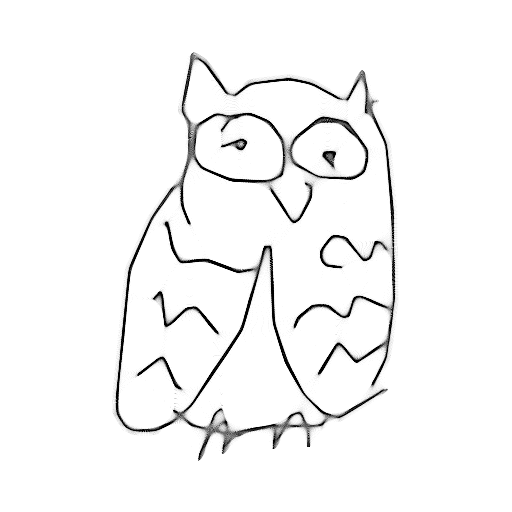} 
& \includegraphics[width=0.075\columnwidth]{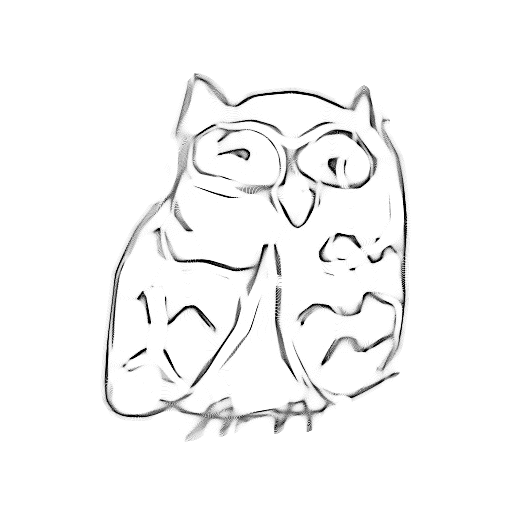} 
& \includegraphics[width=0.075\columnwidth]{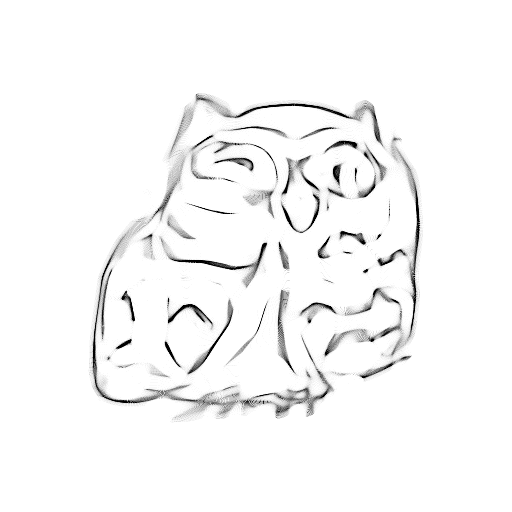} 
& \includegraphics[width=0.075\columnwidth]{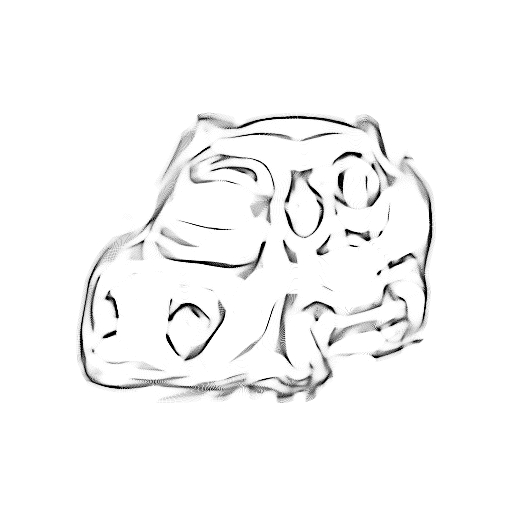} 
& \includegraphics[width=0.075\columnwidth]{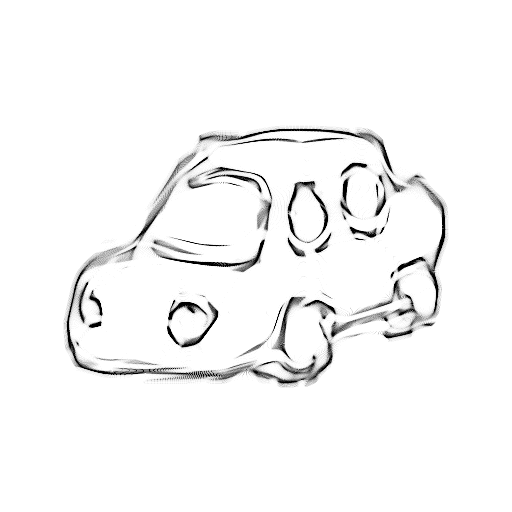} 
& \includegraphics[width=0.075\columnwidth]{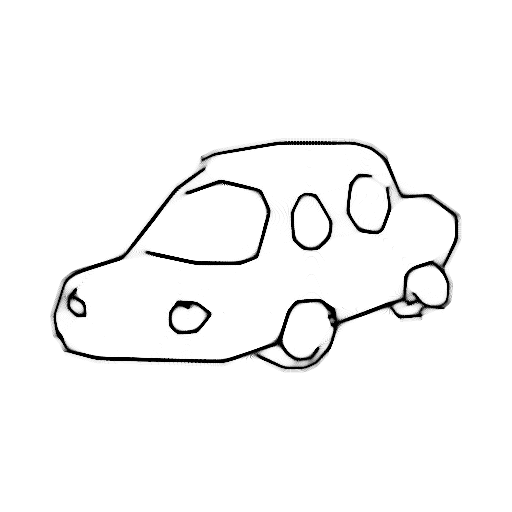} \\

\raisebox{\normalbaselineskip}[0pt][0pt]{\rotatebox[origin=c]{90}{\tiny Ours}} 
& \includegraphics[width=0.075\columnwidth]{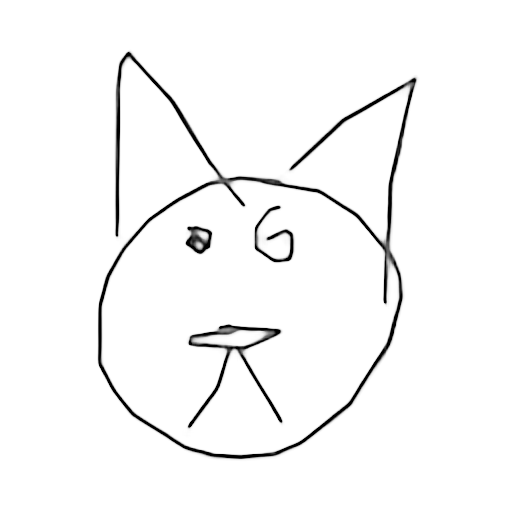} 
& \includegraphics[width=0.075\columnwidth]{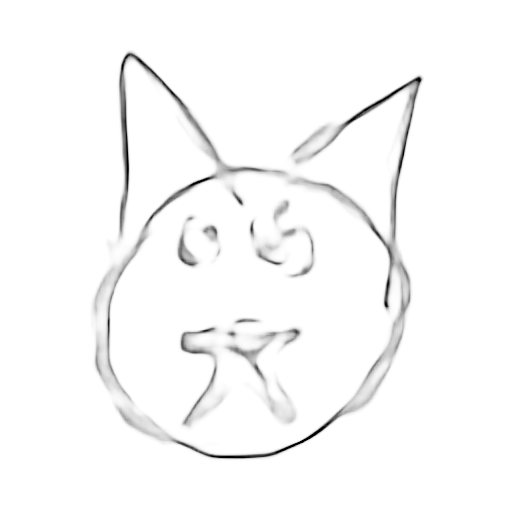} 
& \includegraphics[width=0.075\columnwidth]{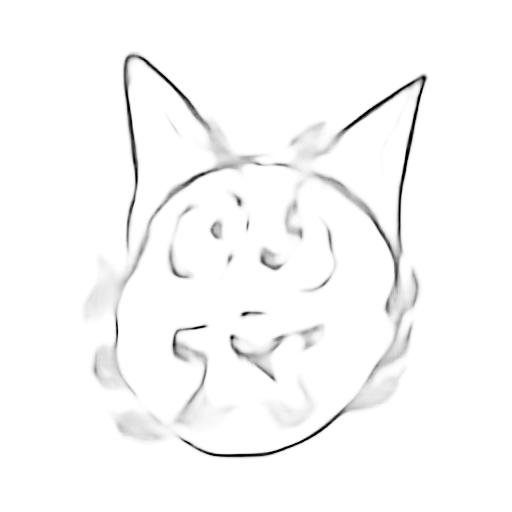} 
& \includegraphics[width=0.075\columnwidth]{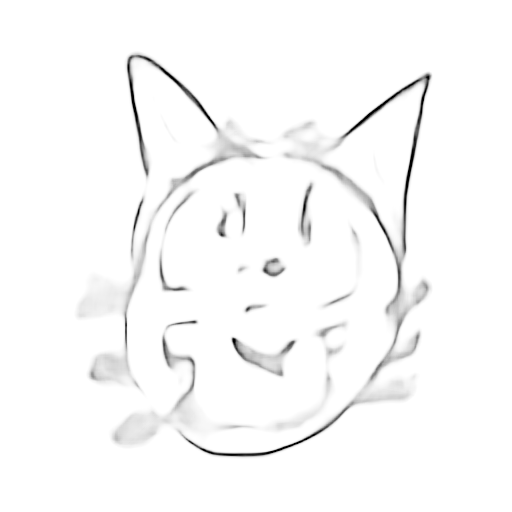} 
& \includegraphics[width=0.075\columnwidth]{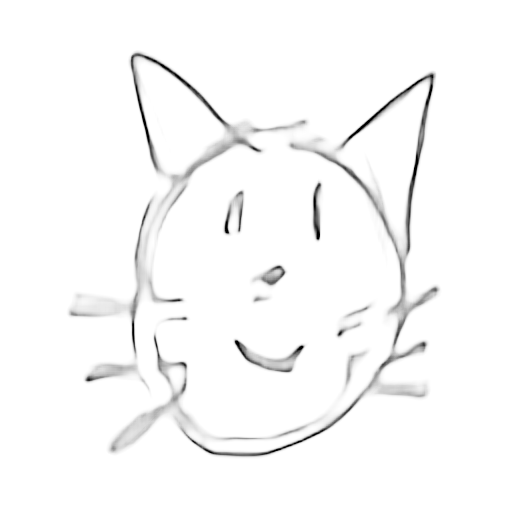} 
& \includegraphics[width=0.075\columnwidth]{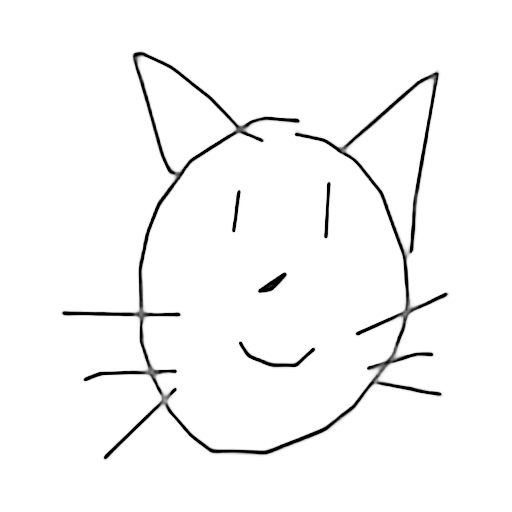} 
& \includegraphics[width=0.075\columnwidth]{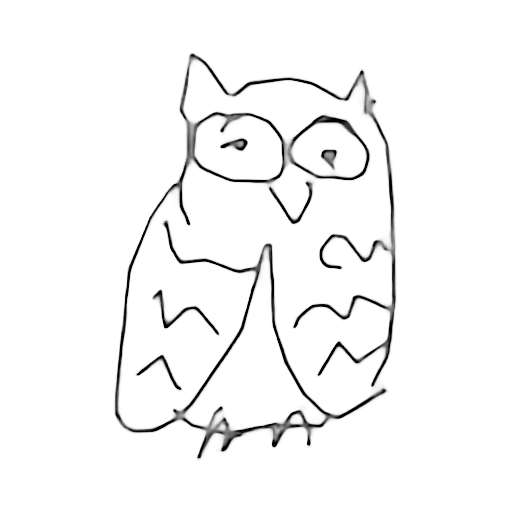} 
& \includegraphics[width=0.075\columnwidth]{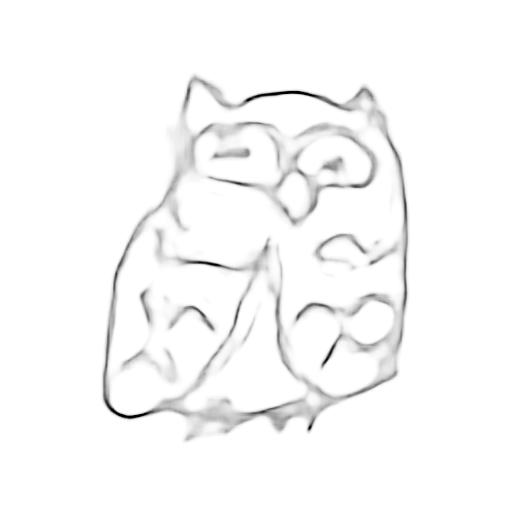} 
& \includegraphics[width=0.075\columnwidth]{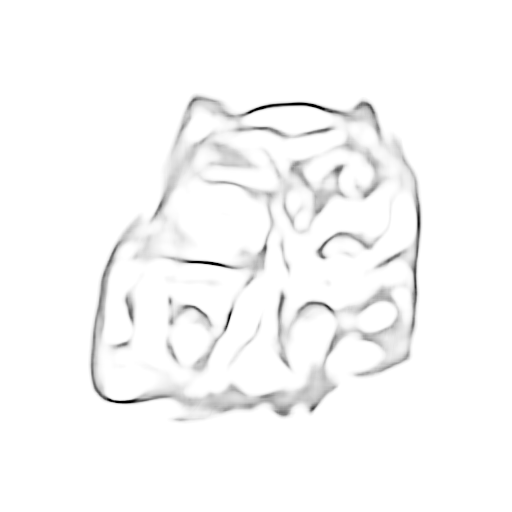} 
& \includegraphics[width=0.075\columnwidth]{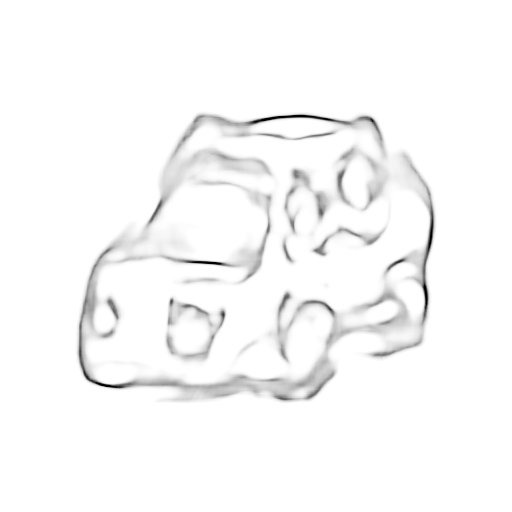} 
& \includegraphics[width=0.075\columnwidth]{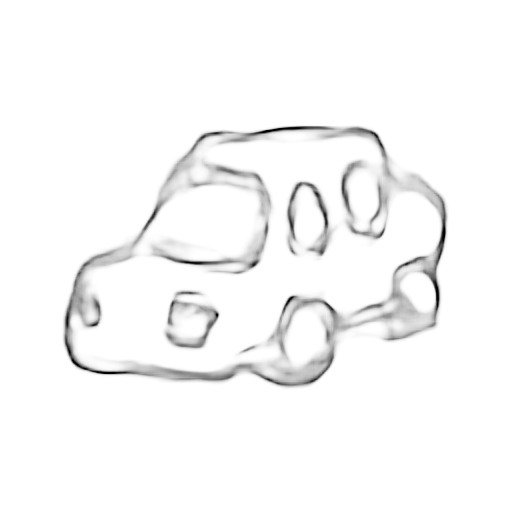} 
& \includegraphics[width=0.075\columnwidth]{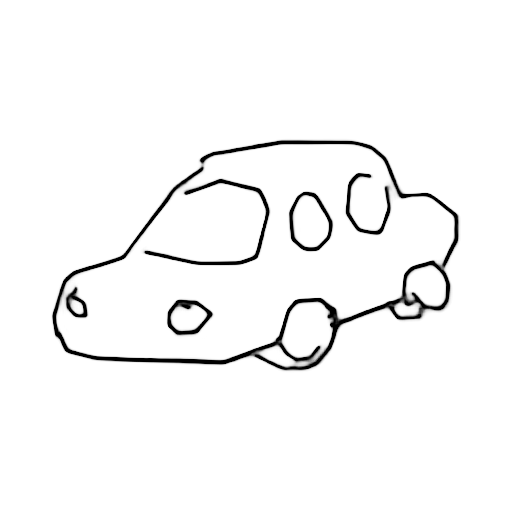} \\

\raisebox{\normalbaselineskip}[0pt][0pt]{\rotatebox[origin=c]{90}{\tiny DWE}} 
& \includegraphics[width=0.075\columnwidth]{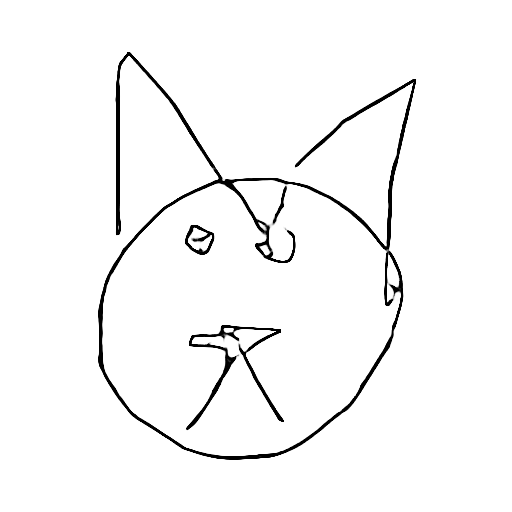} 
& \includegraphics[width=0.075\columnwidth]{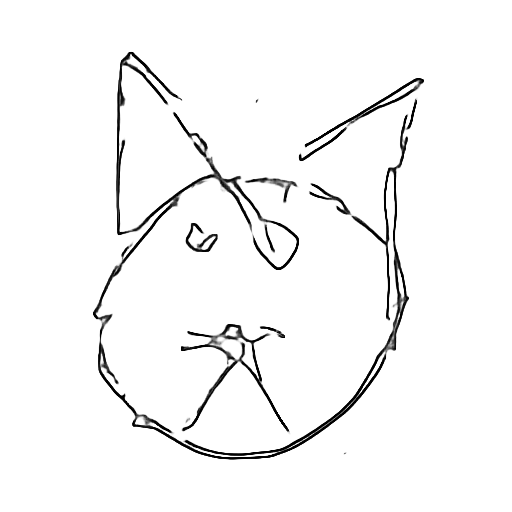} 
& \includegraphics[width=0.075\columnwidth]{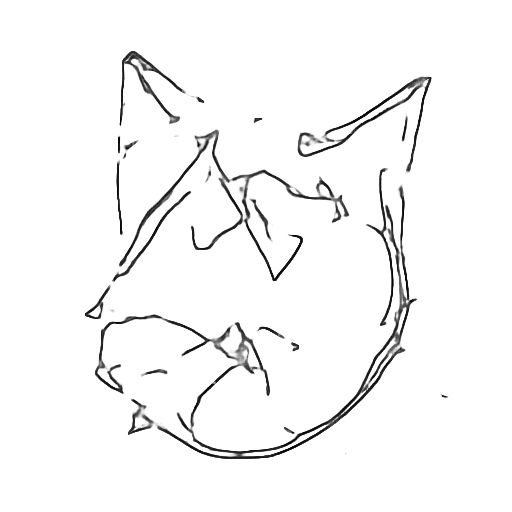} 
& \includegraphics[width=0.075\columnwidth]{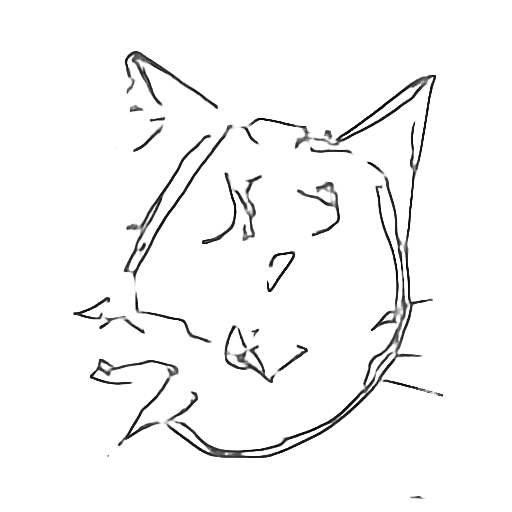} 
& \includegraphics[width=0.075\columnwidth]{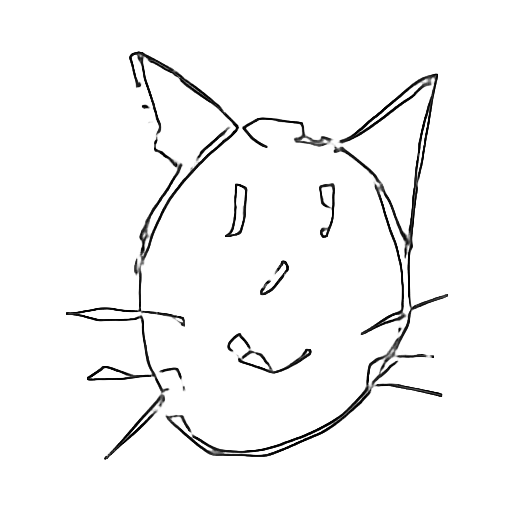} 
& \includegraphics[width=0.075\columnwidth]{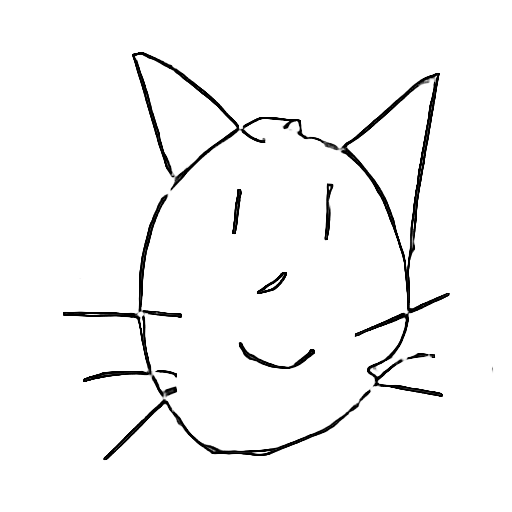} 
& \includegraphics[width=0.075\columnwidth]{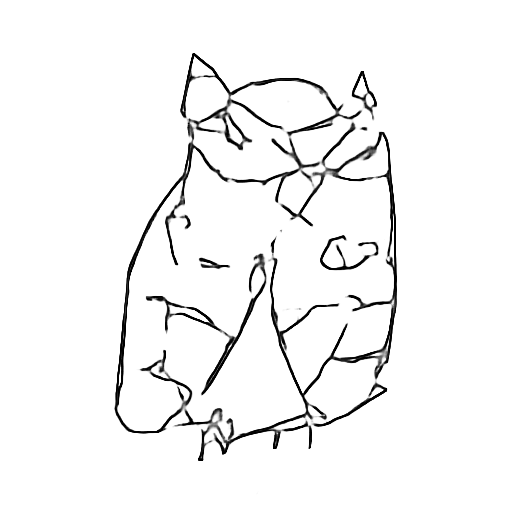} 
& \includegraphics[width=0.075\columnwidth]{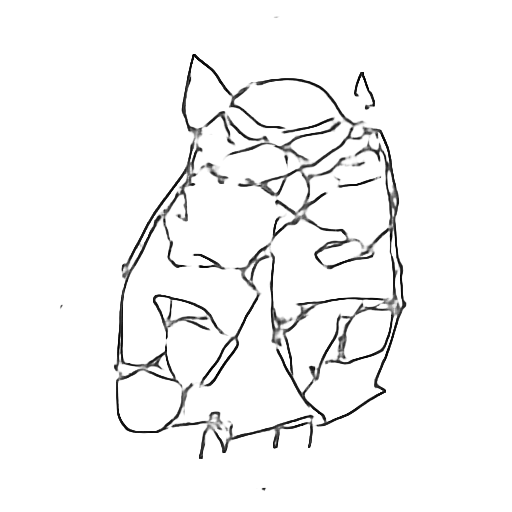} 
& \includegraphics[width=0.075\columnwidth]{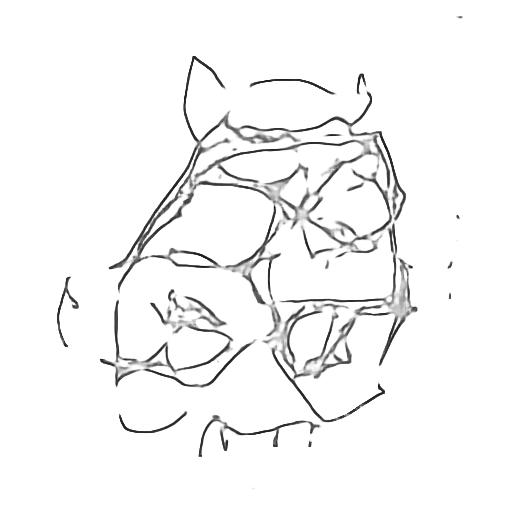} 
& \includegraphics[width=0.075\columnwidth]{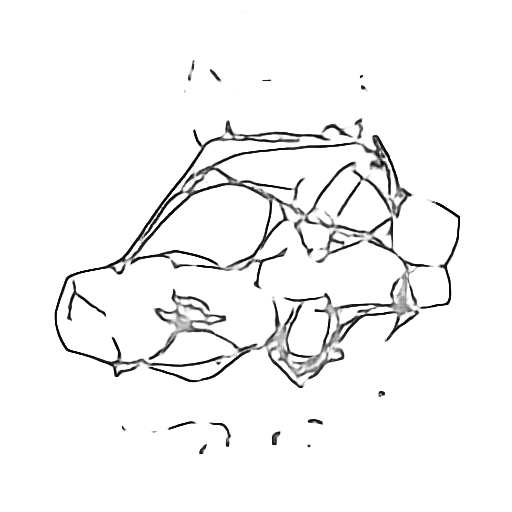} 
& \includegraphics[width=0.075\columnwidth]{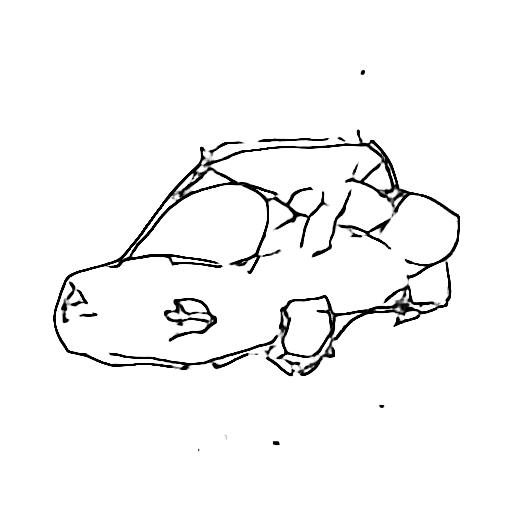} 
& \includegraphics[width=0.075\columnwidth]{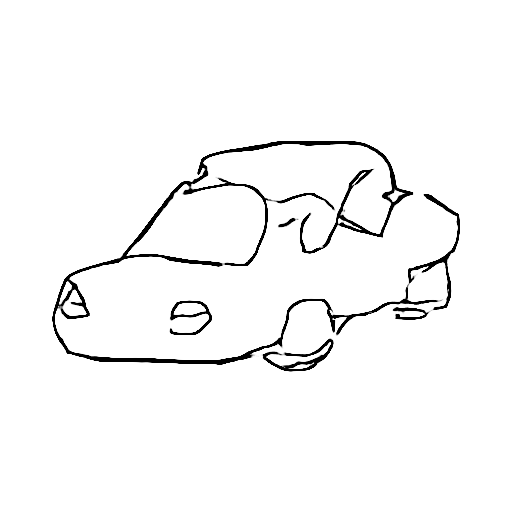} \\
\end{tabular}

\caption{Interpolations between two $512\times512$ images from the \emph{Quick, Draw!} dataset using Geomloss, our model and the Deep Wasserstein Embedding (DWE) method from ~\citep{courty2017learning} adapted to handle $512\times512$ images}
\label{figure:512x512_dwe_comparison}
\end{figure}

We compare our method with the Deep Wasserstein Embedding (DWE) model developed by ~\citet{courty2017learning} on \emph{Quick, Draw!} images. We propose two versions of DWE. The first version relies on the exact original architecture which can only process $28\times28$ images, retrained on a downsampled version of our shape contours dataset -- see Fig.~\ref{figure:28x28_dwe_comparison} for this comparison. In the second version, we adapt their network to process $512\times512$ inputs. The encoder and decoder of this second version have the same architecture as the contractive and expansive paths that we use in our model without our skip connections, but is used to compute the embedding rather than directly predicting barycenters -- see Fig.~\ref{figure:512x512_dwe_comparison}.

\begin{figure}[ht]
\centering
\includegraphics[width=.4\linewidth]{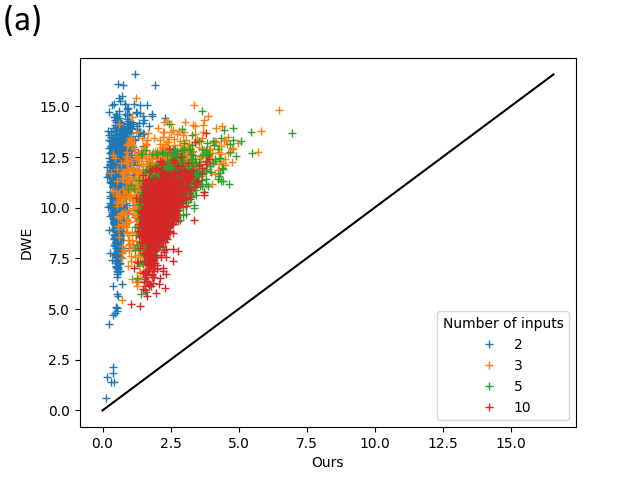}
\includegraphics[width=.4\linewidth]{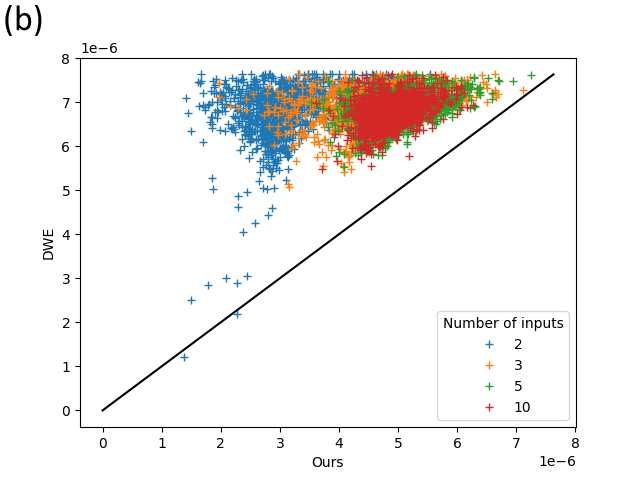}
\caption{Approximation error of our model compared to the ones of DWE (version adapted to handle $512\times512$ images), respectively measured in terms of (a) KL-Divergence and (b) L1 distance, on images coming from our synthetic test dataset. Each one of the $1000\times4$ points corresponds to a barycenter. The x-axis represents the error measured between the GeomLoss barycenter and the barycenter predicted by our model while the y-axis represents the one between the GeomLoss barycenter and the barycenter predicted by DWE. The color of a point associated to a barycenter represents its number of inputs}
\label{figure:errs_Ours_vs_DWE}
\end{figure}

In Fig.\ref{figure:errs_Ours_vs_DWE}, we show a numerical comparison  of approximation errors between our model and DWE adapted to $512\times512$ images of our shape contours dataset, in terms of KL-divergence and L1 distance. Our results clearly show that our method is able to approximate more accurately the Wasserstein barycenter on $512\times512$ input measures. 

\begin{figure}[ht]
\centering

\begin{tabular}{@{}c@{\hspace{2mm}}c@{}c@{}c@{}c@{}c@{}c}
\raisebox{2\normalbaselineskip}[0pt][0pt]{\rotatebox[origin=c]{90}{\hspace{2mm}\scriptsize GeomLoss}} & \includegraphics[width=0.125\columnwidth]{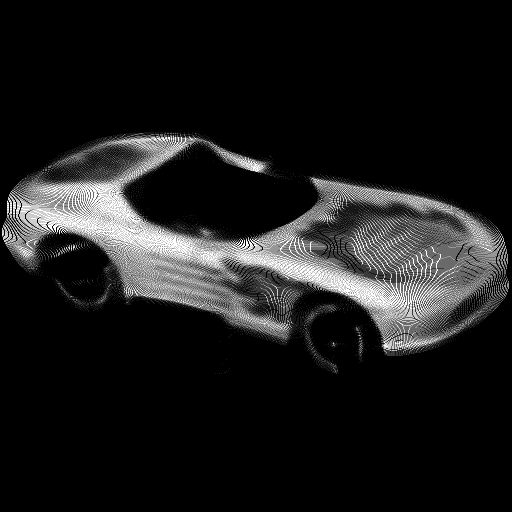}
& \includegraphics[width=0.125\columnwidth]{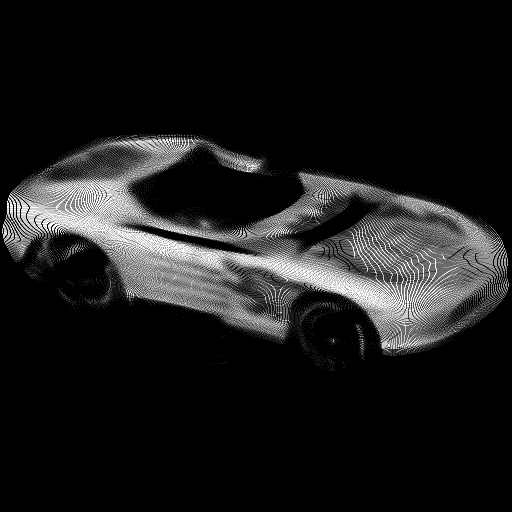}
& \includegraphics[width=0.125\columnwidth]{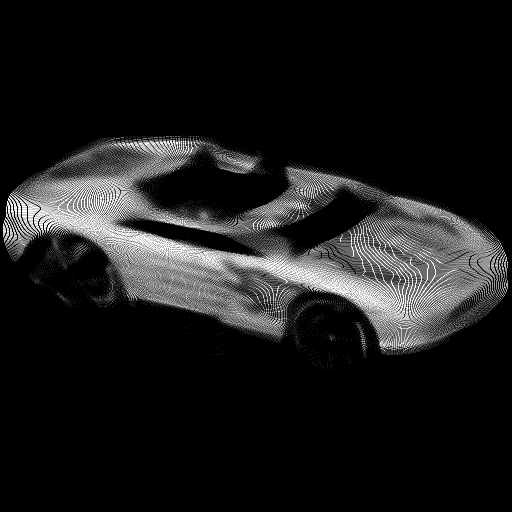}
& \includegraphics[width=0.125\columnwidth]{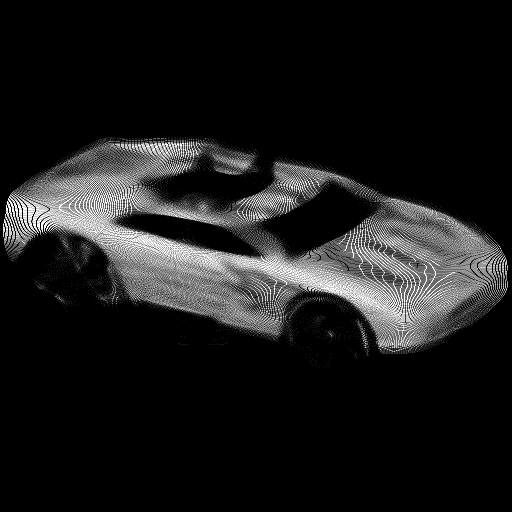}
& \includegraphics[width=0.125\columnwidth]{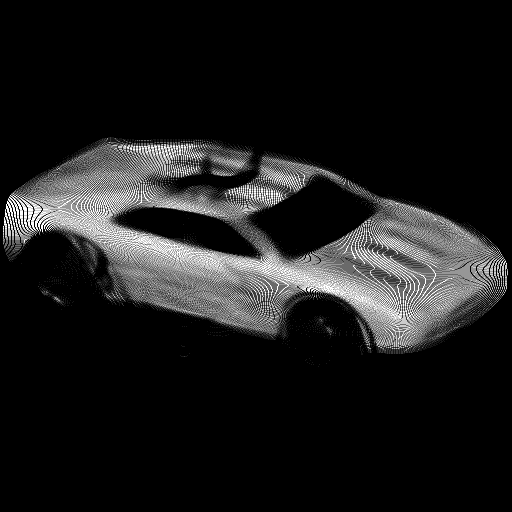}
& \includegraphics[width=0.125\columnwidth]{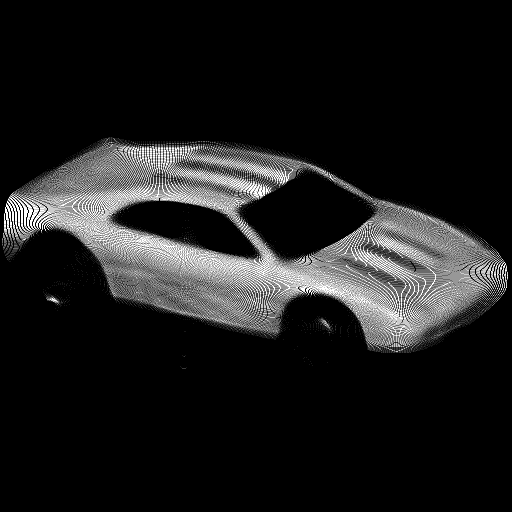} \\

\raisebox{1.8\normalbaselineskip}[0pt][0pt]{\rotatebox[origin=c]{90}{\hspace{3mm}\scriptsize Ours}} 
& \includegraphics[width=0.125\columnwidth]{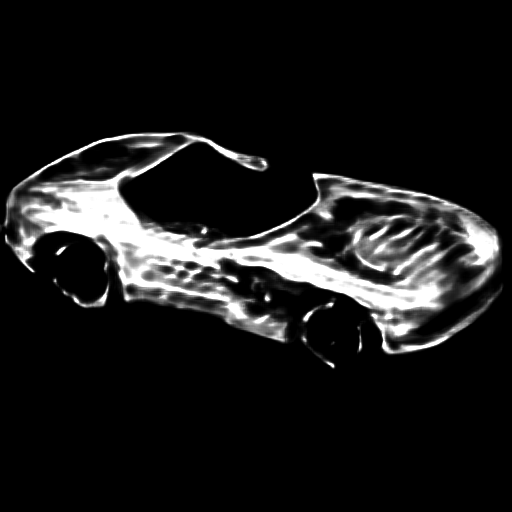}
& \includegraphics[width=0.125\columnwidth]{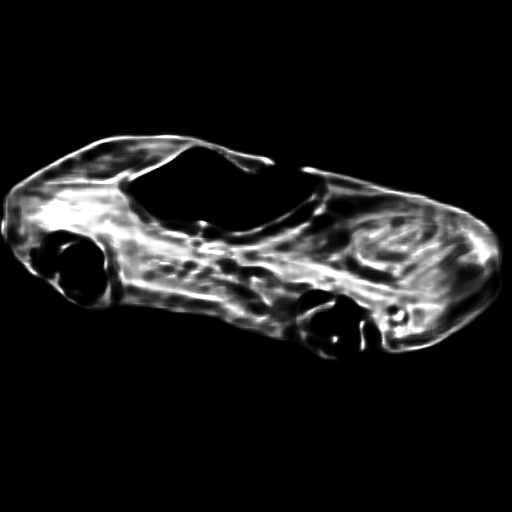}
& \includegraphics[width=0.125\columnwidth]{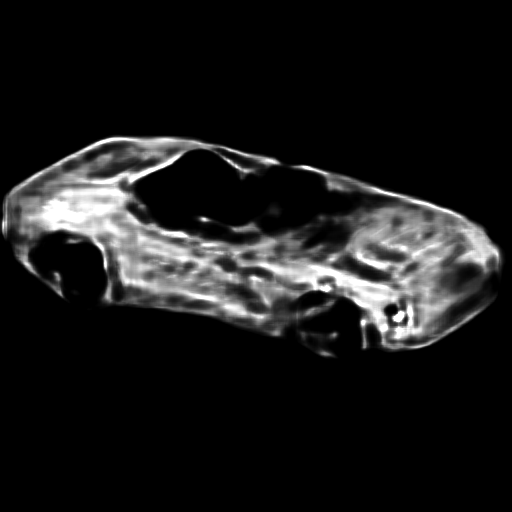}
& \includegraphics[width=0.125\columnwidth]{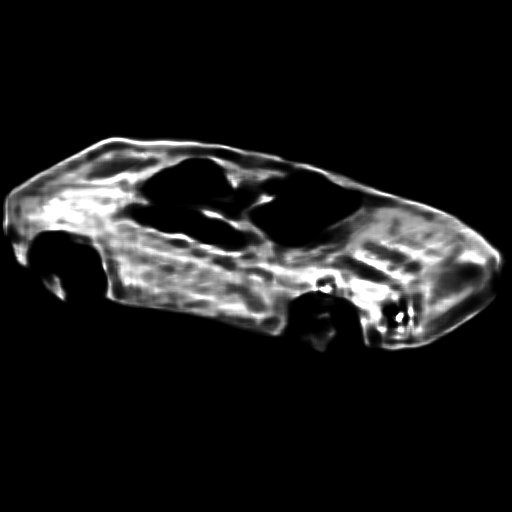}
& \includegraphics[width=0.125\columnwidth]{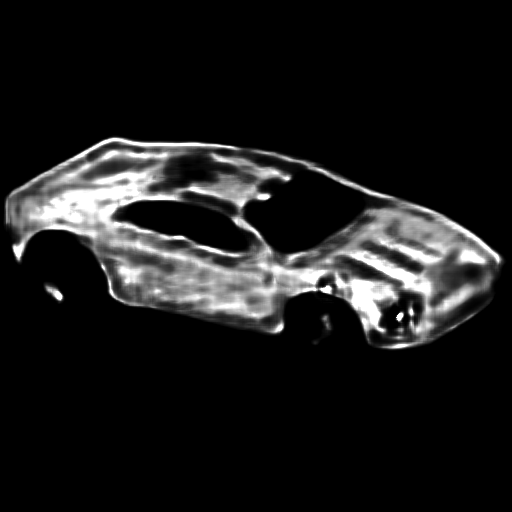}
& \includegraphics[width=0.125\columnwidth]{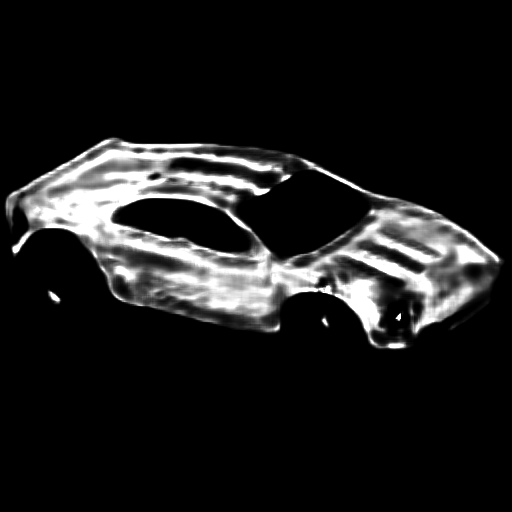} \\
\end{tabular}

\caption{Interpolations between two $512\times512$ images from the \emph{Coil20} dataset \citep{nane1996columbia} using GeomLoss and our model trained with synthetic shape contours. For visualization purposes, white values represent high mass concentration, while dark values represent low mass concentration}
\label{figure:coil20_cars}
\end{figure}

Finally, we study the limitations of the generalization of our network on the \emph{Coil20} dataset \citep{nane1996columbia}, which consists of images of objects on a black background. In figure \ref{figure:coil20_cars}, we show the interpolation of 2 cars ; additional results are available in appendix \ref{appendix:additionalresults}, figure \ref{figure:coil20_cup_car}. 

\subsection{N-way barycenters}
Even if our model has been trained using only barycenters computed from pairs of inputs, we can apply it to predict barycenters of more than two measures. This section illustrates N-way barycenters on 2-d sketch images and color distributions.

\begin{figure}[ht]
\centering

\begin{tabular}{@{}c@{}c@{}c@{}c@{}c@{}c@{}c@{}c@{}}
\raisebox{2\normalbaselineskip}[0pt][0pt]{\rotatebox[origin=c]{90}{\tiny Inputs}} & &  \includegraphics[width=0.1\columnwidth]{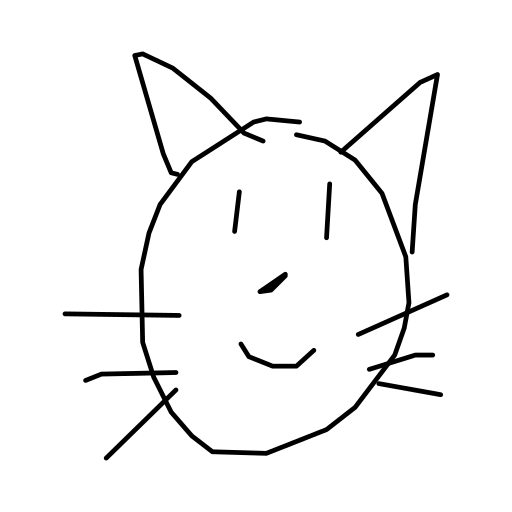} & \includegraphics[width=0.1\columnwidth]{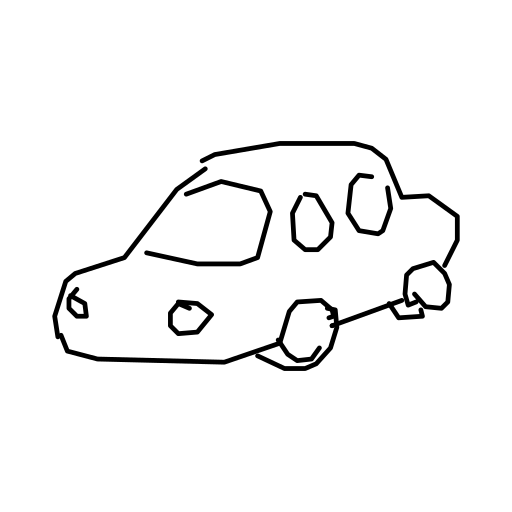} & \includegraphics[width=0.1\columnwidth]{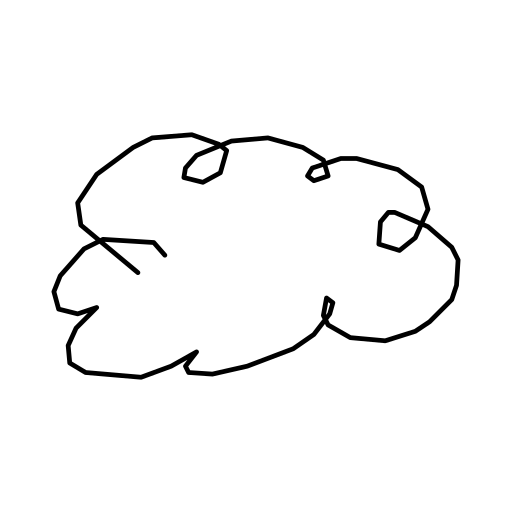} & \includegraphics[width=0.1\columnwidth]{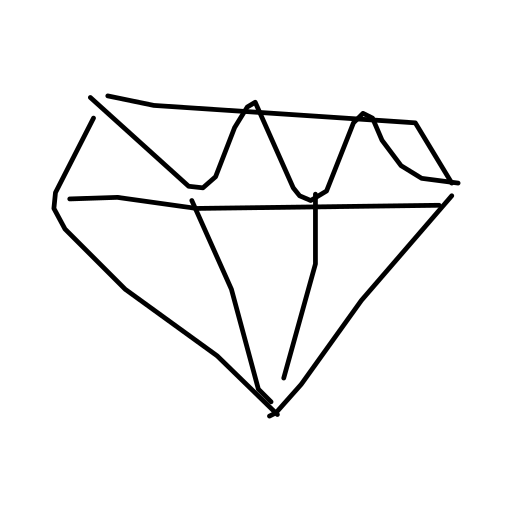} & \includegraphics[width=0.1\columnwidth]{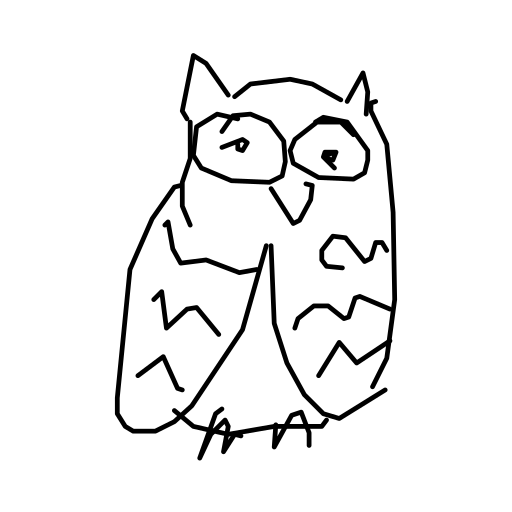} &  \\
\raisebox{2\normalbaselineskip}[0pt][0pt]{\rotatebox[origin=c]{90}{\tiny GeomLoss (3)}} & \includegraphics[width=0.1\columnwidth]{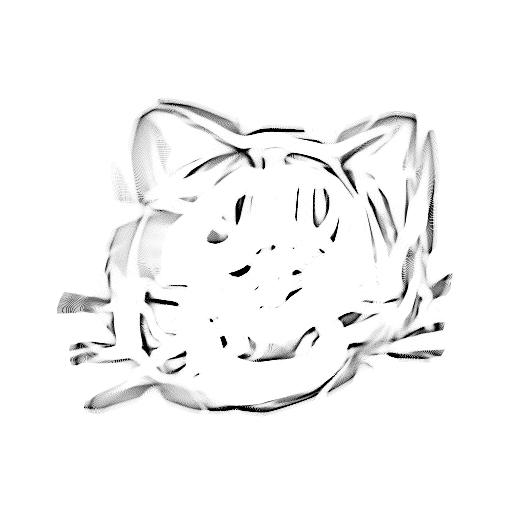} & \includegraphics[width=0.1\columnwidth]{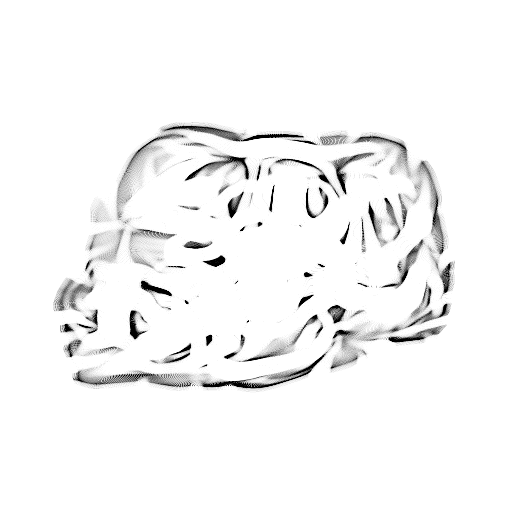} & \includegraphics[width=0.1\columnwidth]{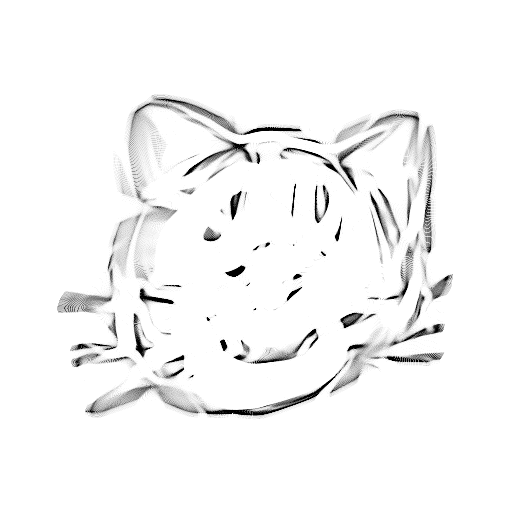} & \includegraphics[width=0.1\columnwidth]{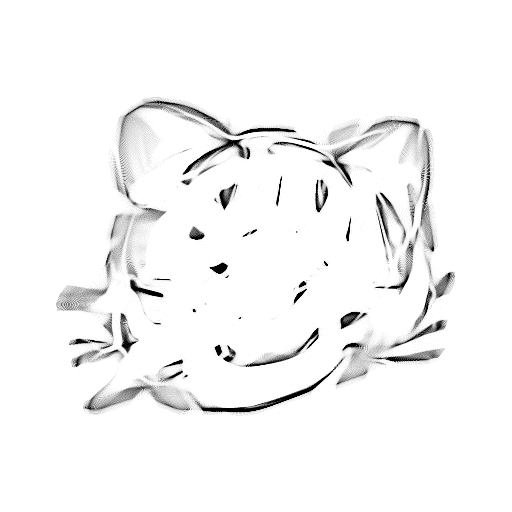} & \includegraphics[width=0.1\columnwidth]{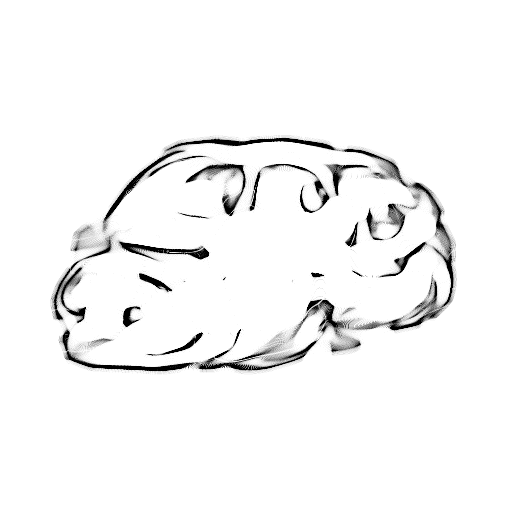} & \includegraphics[width=0.1\columnwidth]{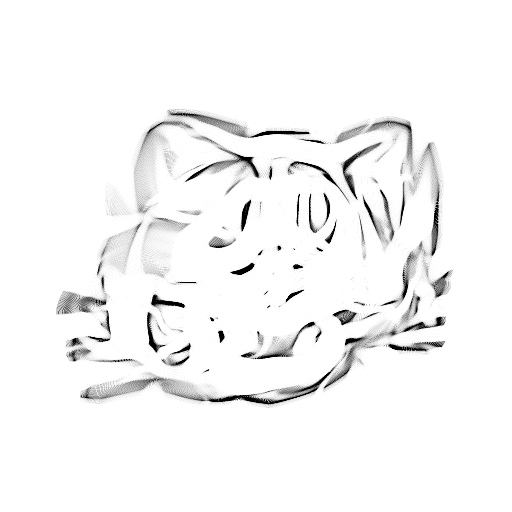} & \includegraphics[width=0.1\columnwidth]{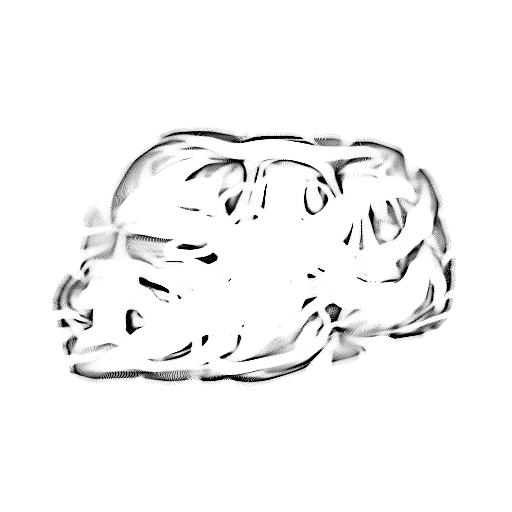} \\
\raisebox{2\normalbaselineskip}[0pt][0pt]{\rotatebox[origin=c]{90}{\tiny Ours (3)}} & \includegraphics[width=0.1\columnwidth]{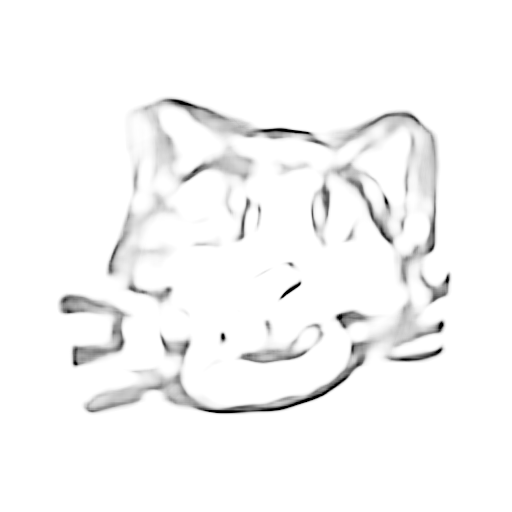} & \includegraphics[width=0.1\columnwidth]{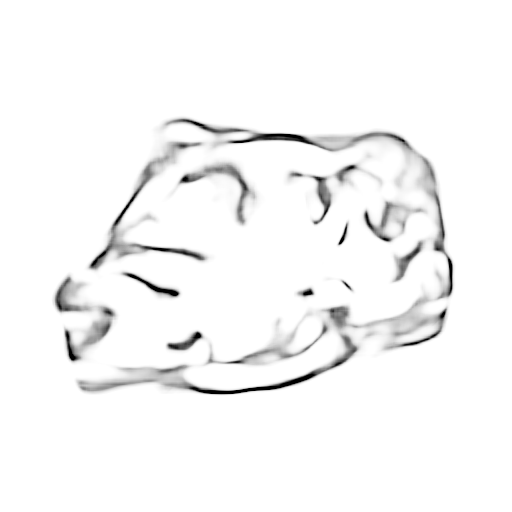} & \includegraphics[width=0.1\columnwidth]{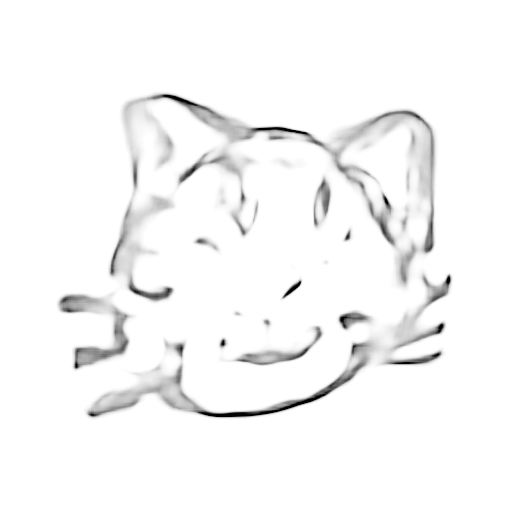} & \includegraphics[width=0.1\columnwidth]{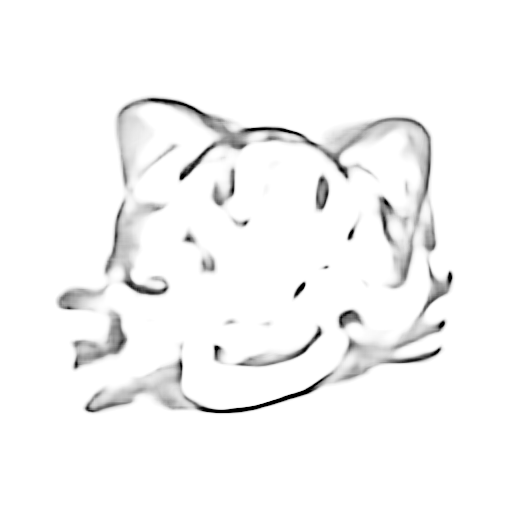} & \includegraphics[width=0.1\columnwidth]{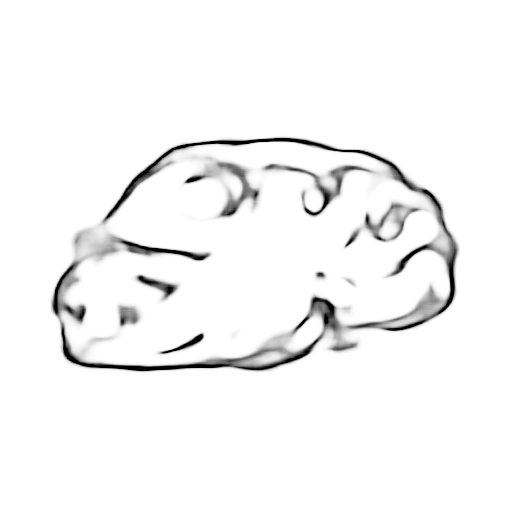} & \includegraphics[width=0.1\columnwidth]{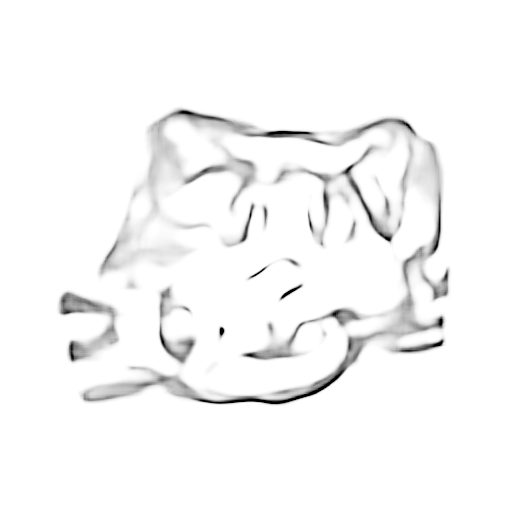} & \includegraphics[width=0.1\columnwidth]{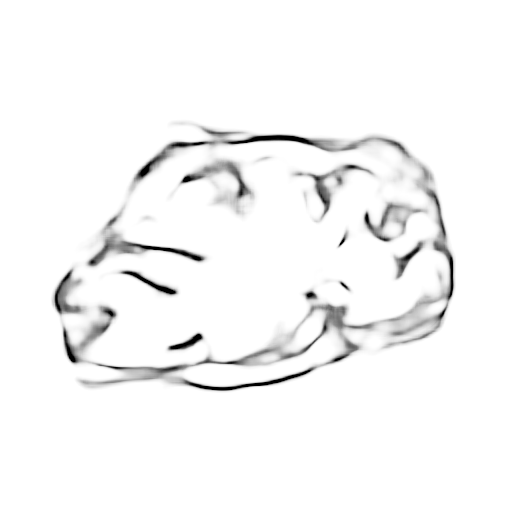} \\
\raisebox{2\normalbaselineskip}[0pt][0pt]{\rotatebox[origin=c]{90}{\tiny GeomLoss (5)}} & \includegraphics[width=0.1\columnwidth]{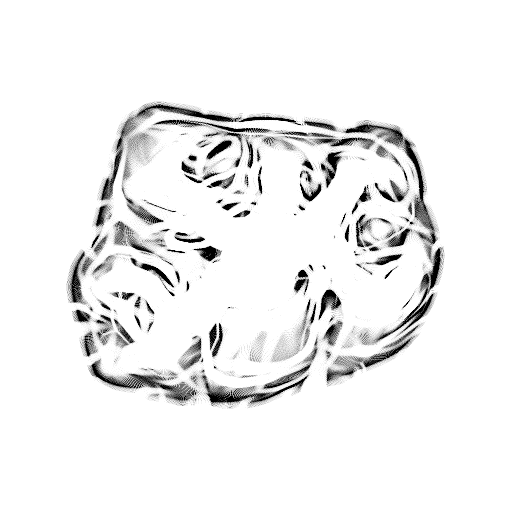} & \includegraphics[width=0.1\columnwidth]{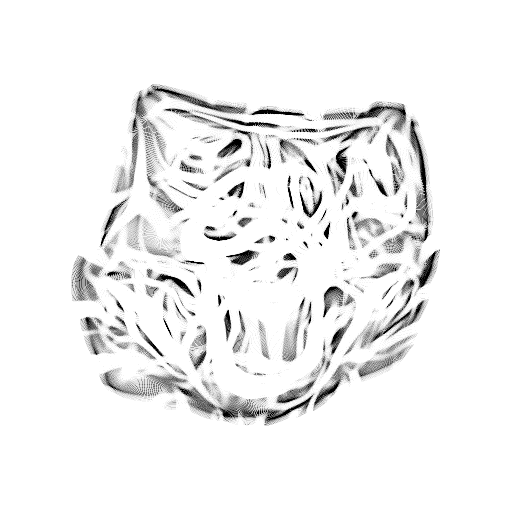} & \includegraphics[width=0.1\columnwidth]{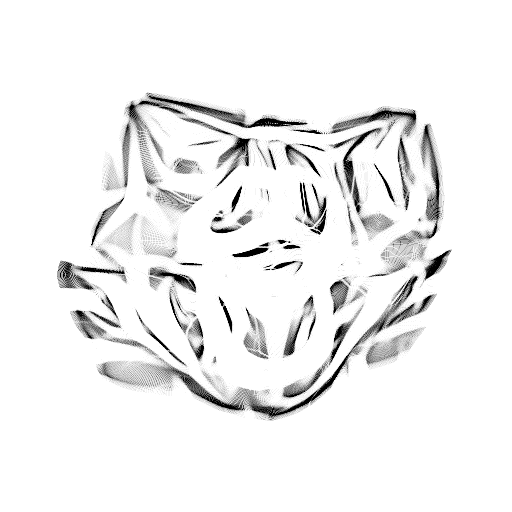} & \includegraphics[width=0.1\columnwidth]{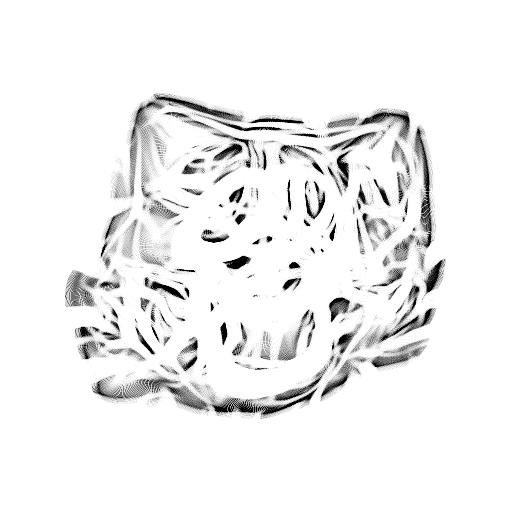} & \includegraphics[width=0.1\columnwidth]{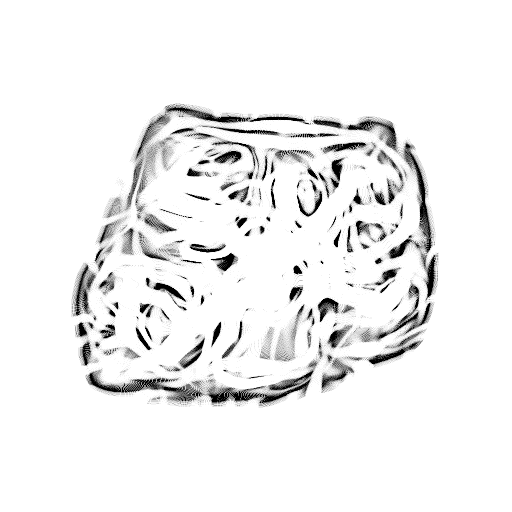} & \includegraphics[width=0.1\columnwidth]{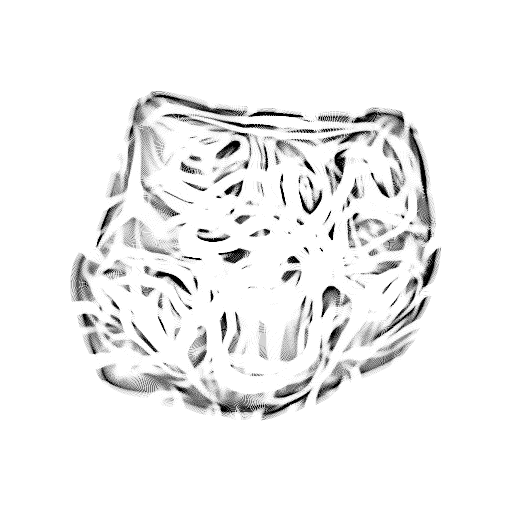} & \includegraphics[width=0.1\columnwidth]{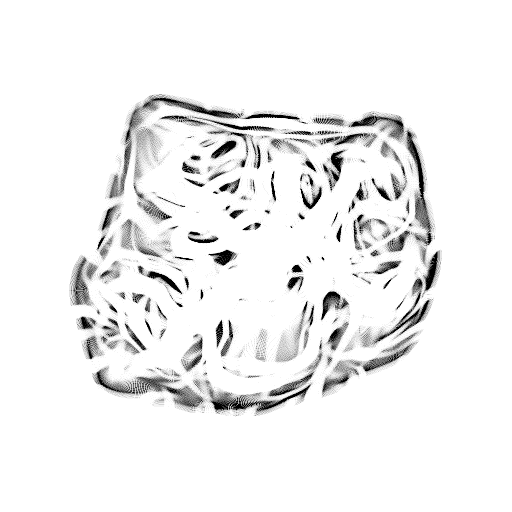} \\
\raisebox{2\normalbaselineskip}[0pt][0pt]{\rotatebox[origin=c]{90}{\tiny Ours (5)}} & \includegraphics[width=0.1\columnwidth]{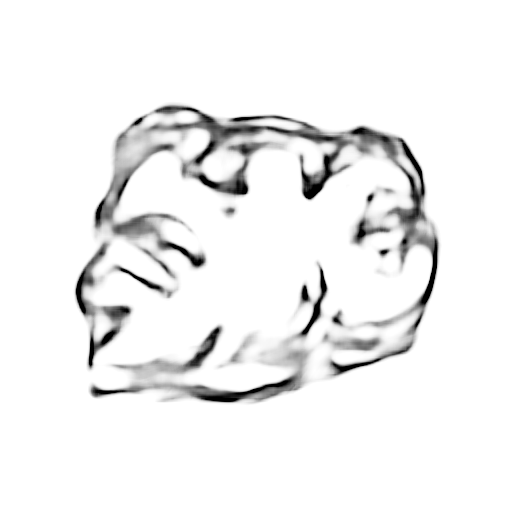} & \includegraphics[width=0.1\columnwidth]{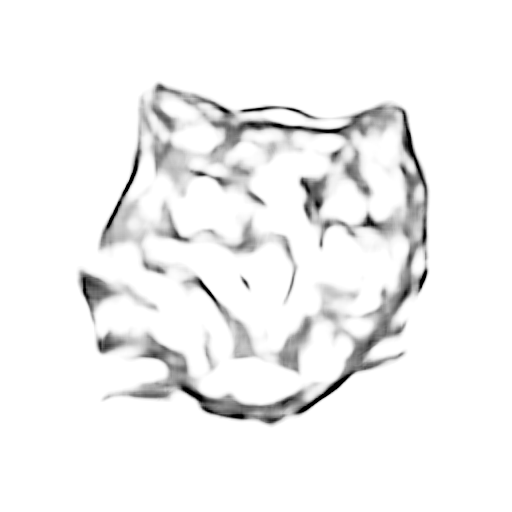} & \includegraphics[width=0.1\columnwidth]{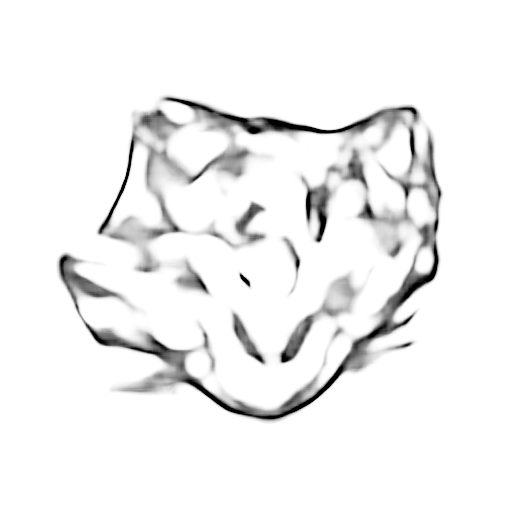} & \includegraphics[width=0.1\columnwidth]{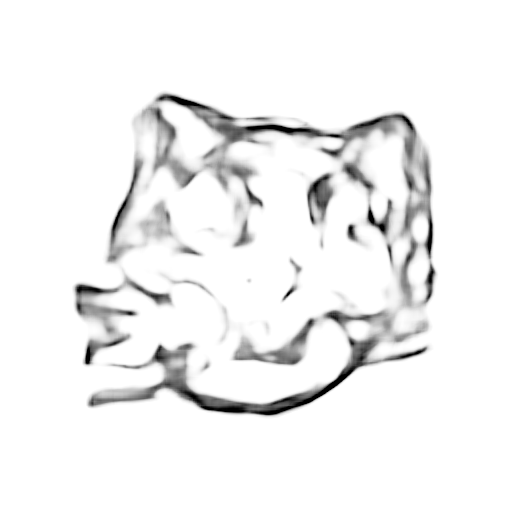} & \includegraphics[width=0.1\columnwidth]{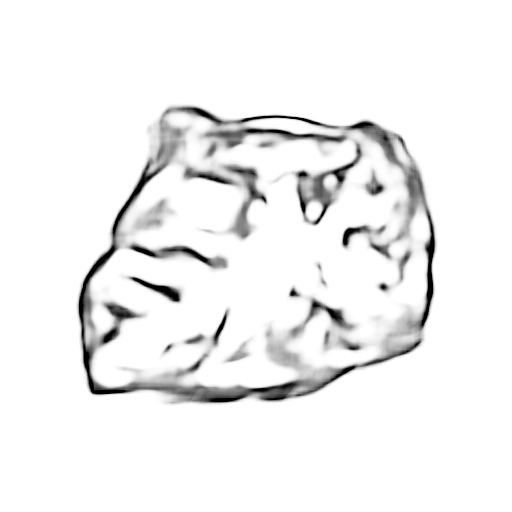} & \includegraphics[width=0.1\columnwidth]{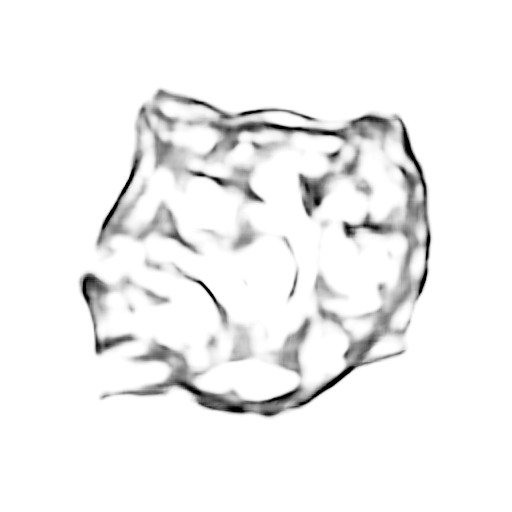} & \includegraphics[width=0.1\columnwidth]{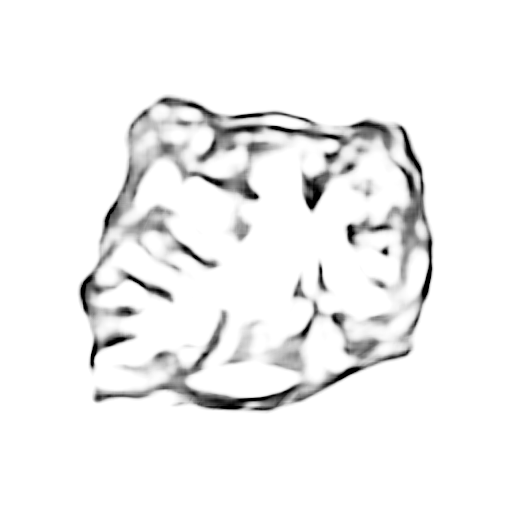} \\
\end{tabular}

\caption{Wasserstein barycenters of three inputs (top rows) and five inputs (bottom rows) from \emph{Quick, Draw!}, respectively computed with Geomloss and with our model trained with only pairs from our synthetic training dataset. Barycentric weights are randomly chosen}
\label{figure:3_5_interpolations}
\vskip -0.2in
\end{figure}

\paragraph{Sketch interpolation.}
We display interpolations between respectively three and five input measures in Fig.\ref{figure:3_5_interpolations}, which surprisingly tends to show that our model can generalize what it learned on pairs of inputs, at least partially. Additional results on \emph{Quick, Draw!} are also shown in appendix \ref{appendix:generalization_n_inputs}, Fig.~\ref{figure:diff_pentagon}. A 100-way barycenter comparison can be found in appendix Sec.~\ref{appendix:additionalresults}, Fig.~\ref{fig:100way}. 

Numerically when the number of inputs is greater than $2$, our model also achieve to find better approximations than the ones obtained with DWE, as shown in Fig.~\ref{figure:errs_Ours_vs_DWE}. 

\paragraph{Interpolating color distributions.}
We also propose color transfer between images as another application of our method in the n-way case, as performed in the literature~\citep{solomon2015convolutional,bonneel2015sliced}. More particularly here we focus on a color grading application: given $n$ images, we are interested in the weighted interpolation of their color histograms. Then we alter the color histogram of a target image so that it matches the interpolated histogram in order to transfer colors. Based on recommendations by \cite{reinhard2011colour}, we consider images in the CIE-Lab space and we perform the transfer by modifying the luminance and the chrominance channels independently. While the transfer of luminance only requires 1D optimal transport plan, chrominance has 2 dimensions. In order to transfer it, we follow the procedure detailed in \citep{solomon2015convolutional}: we first compute the $n$ 2D chrominance histograms $\{\mu_{i}\}_{i=1..n}$ and also $\nu$, that of the target image. 
We then interpolate the $\{\mu_{i}\}_{i=1..n}$ using our model in order to obtain their barycenter $\hat{\mu}$ for given weights. Color transfer requires an explicit knowledge of the transport plan $\pi$ between $\nu$ and $\hat{\mu}$. In our method, $\pi$ is computed using the OT solver GeomLoss to retrieve the dual potentials $f$ and $g$ which are combined yielding $\pi=\exp{\frac{1}{\epsilon}(f+g-C)\cdot\nu\otimes\hat{\mu}}$ where $C$ corresponds to the cost matrix and with $\epsilon=5.10^{-2}$. Note that for this part we do not use Sinkhorn divergences and we instead consider the regularized OT problem in order to retrieve $f$ and $g$. 
This transport plan $\pi$ is used to retrieve the chrominance $T$ associated with the target image:
$T(i)=\frac{1}{\nu}\sum_{j\in M}\pi_{ij}j$ where $i,j\in M$ the set of all the possible discretized chrominance values. 

\begin{figure}[h!]
\vspace{1em}
\centering
\begin{tabular}{@{}c@{}c@{}c@{}c@{}c@{}c@{}c@{}}

& \scriptsize{$\nu$} & \scriptsize{$\mu_{1}$} & \scriptsize{$\mu_{2}$} & \scriptsize{$\mu_{3}$} & \scriptsize{$\mu_{4}$} & \scriptsize{$\mu_{5}$} \\

& \includegraphics[height=0.065\textheight]{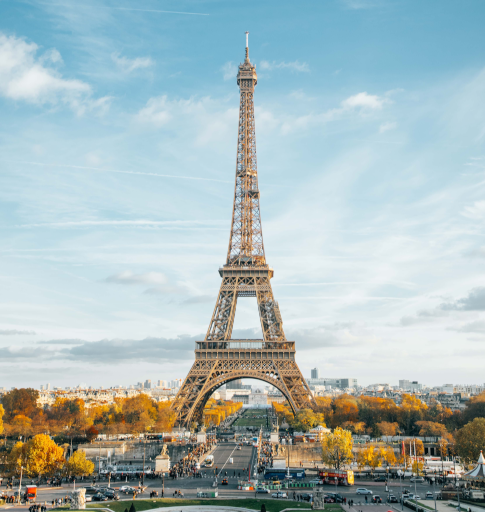} & \includegraphics[height=0.065\textheight]{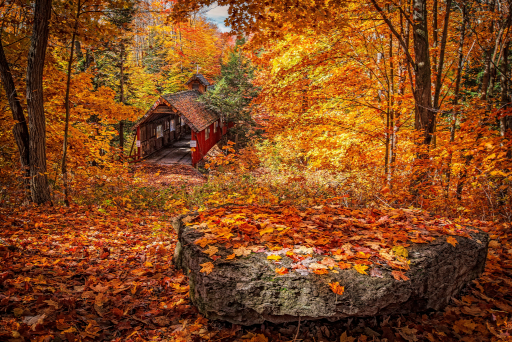} & \includegraphics[height=0.065\textheight]{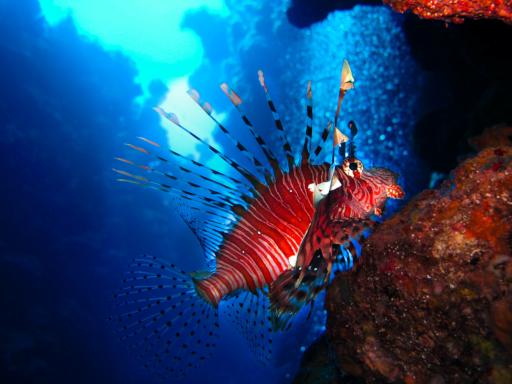} & \includegraphics[height=0.065\textheight]{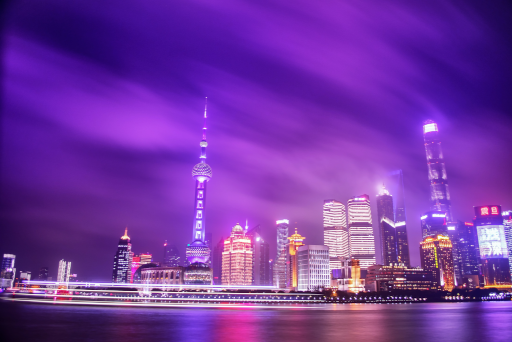} & \includegraphics[height=0.065\textheight]{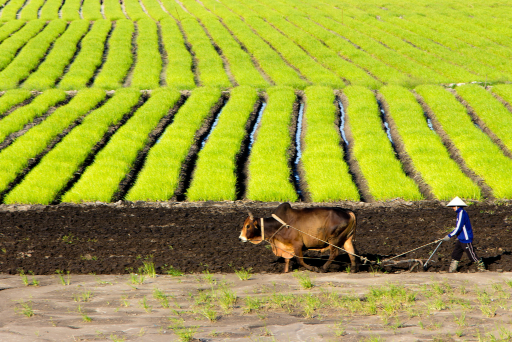} & \includegraphics[height=0.065\textheight]{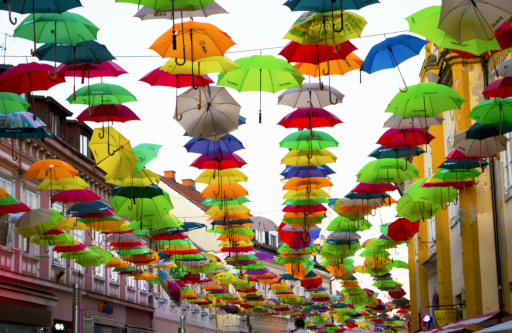} \\

\raisebox{7\normalbaselineskip}[0pt][0pt]{\rotatebox[origin=c]{90}{\tiny GeomLoss}} & \multicolumn{3}{c}{\includegraphics[width=0.31\columnwidth]{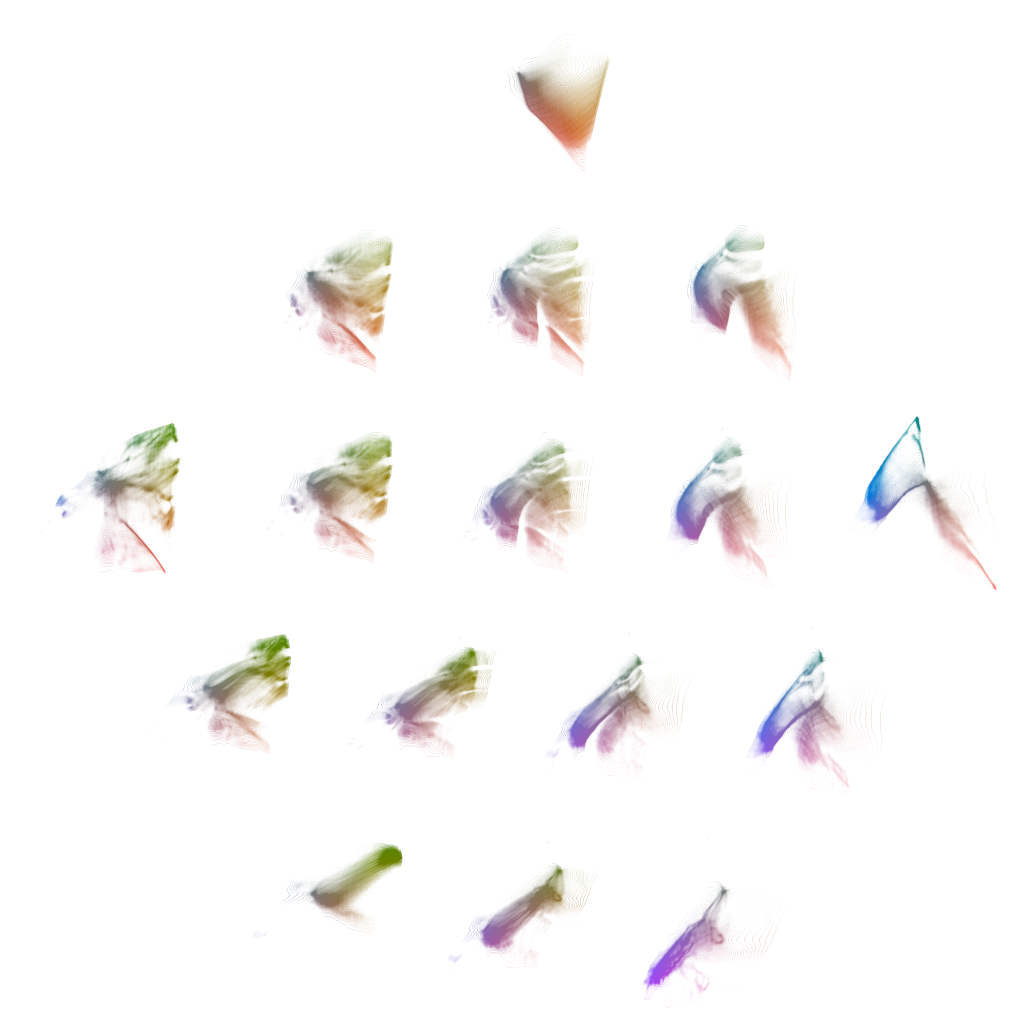}} & \multicolumn{3}{c}{\includegraphics[width=0.31\columnwidth]{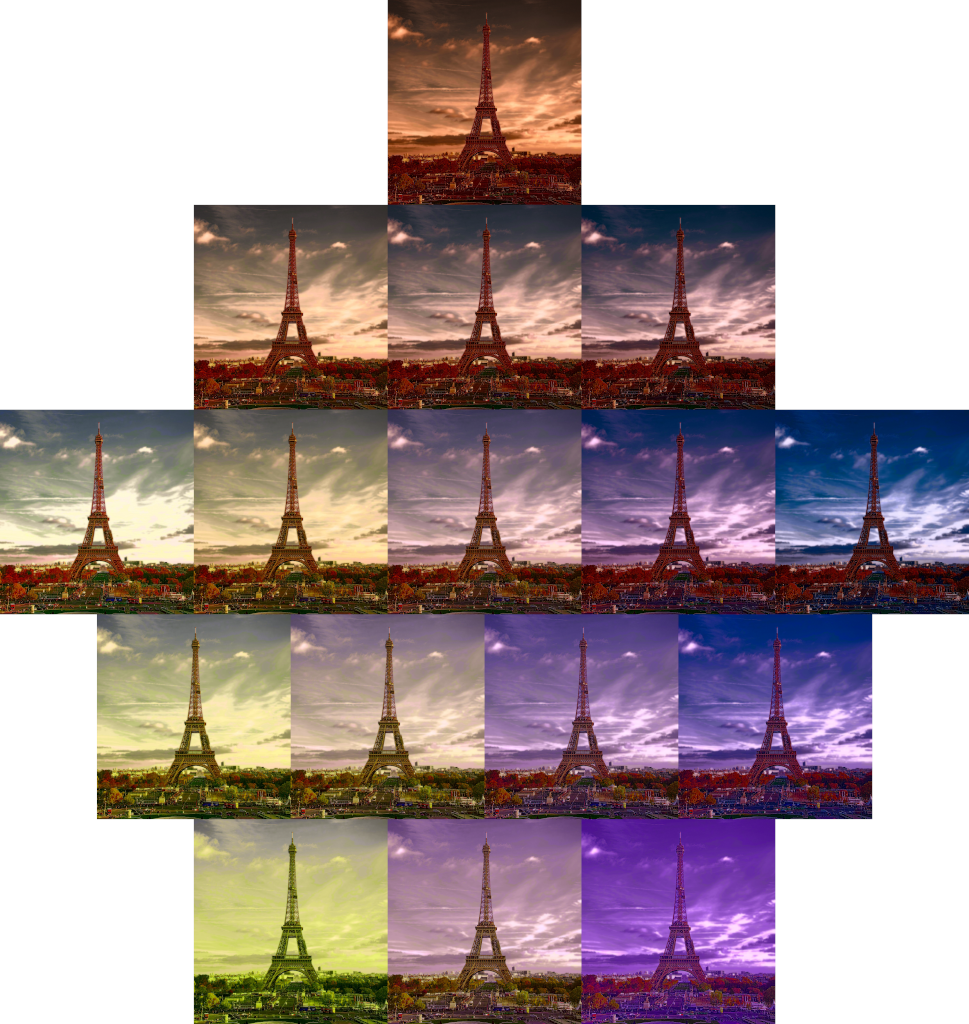}} \\

\raisebox{7\normalbaselineskip}[0pt][0pt]{\rotatebox[origin=c]{90}{\tiny \emph{ContoursDS}}} &
\multicolumn{3}{c}{\includegraphics[width=0.31\columnwidth]{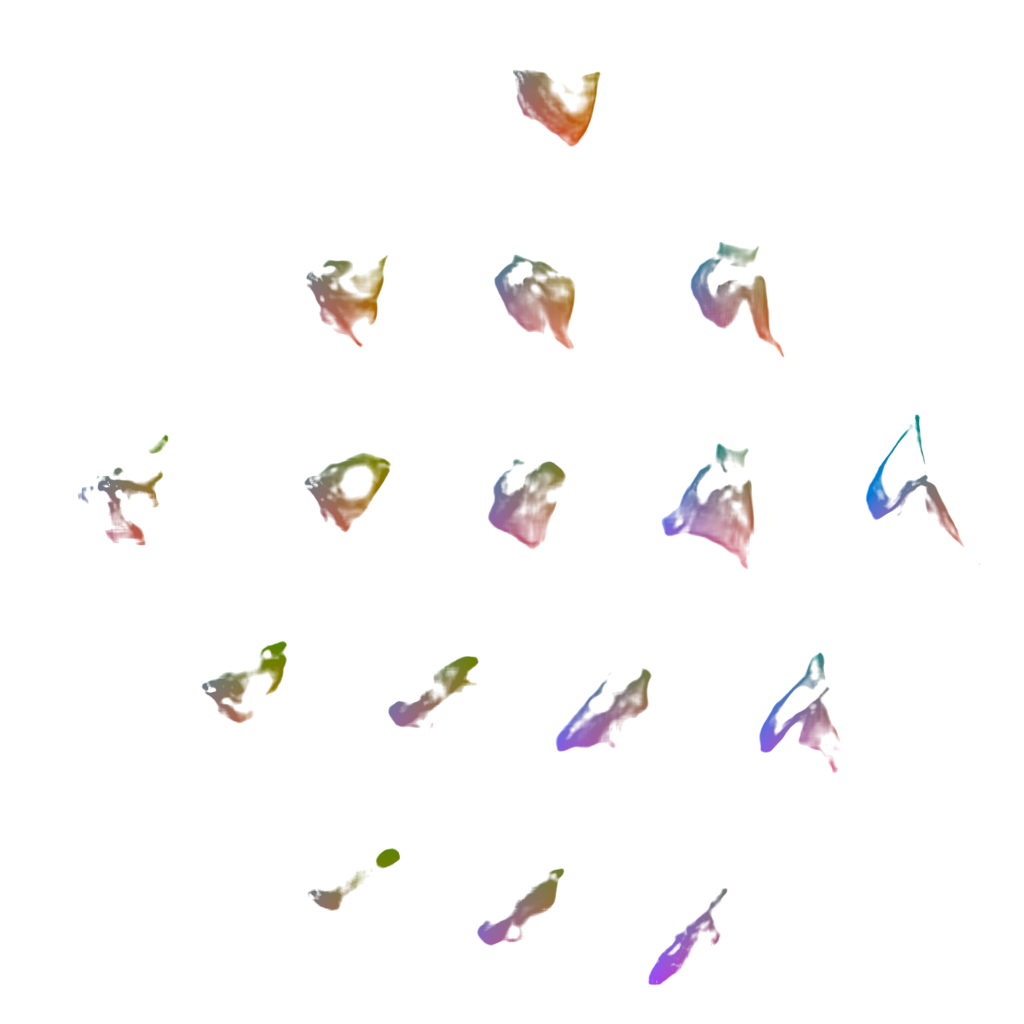}} & \multicolumn{3}{c}{\includegraphics[width=0.31\columnwidth]{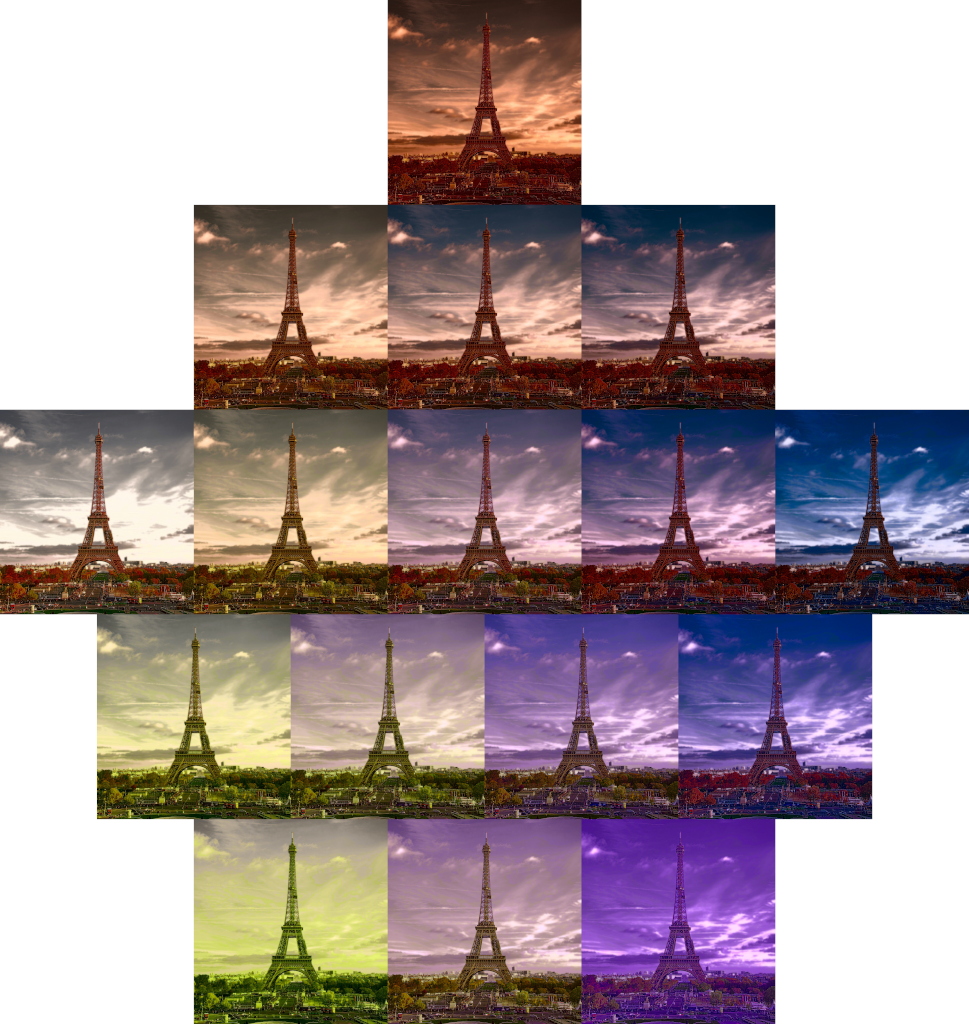}} \\

\raisebox{7\normalbaselineskip}[0pt][0pt]{\rotatebox[origin=c]{90}{\tiny \emph{HistoDS}}} &
\multicolumn{3}{c}{\includegraphics[width=0.31\columnwidth]{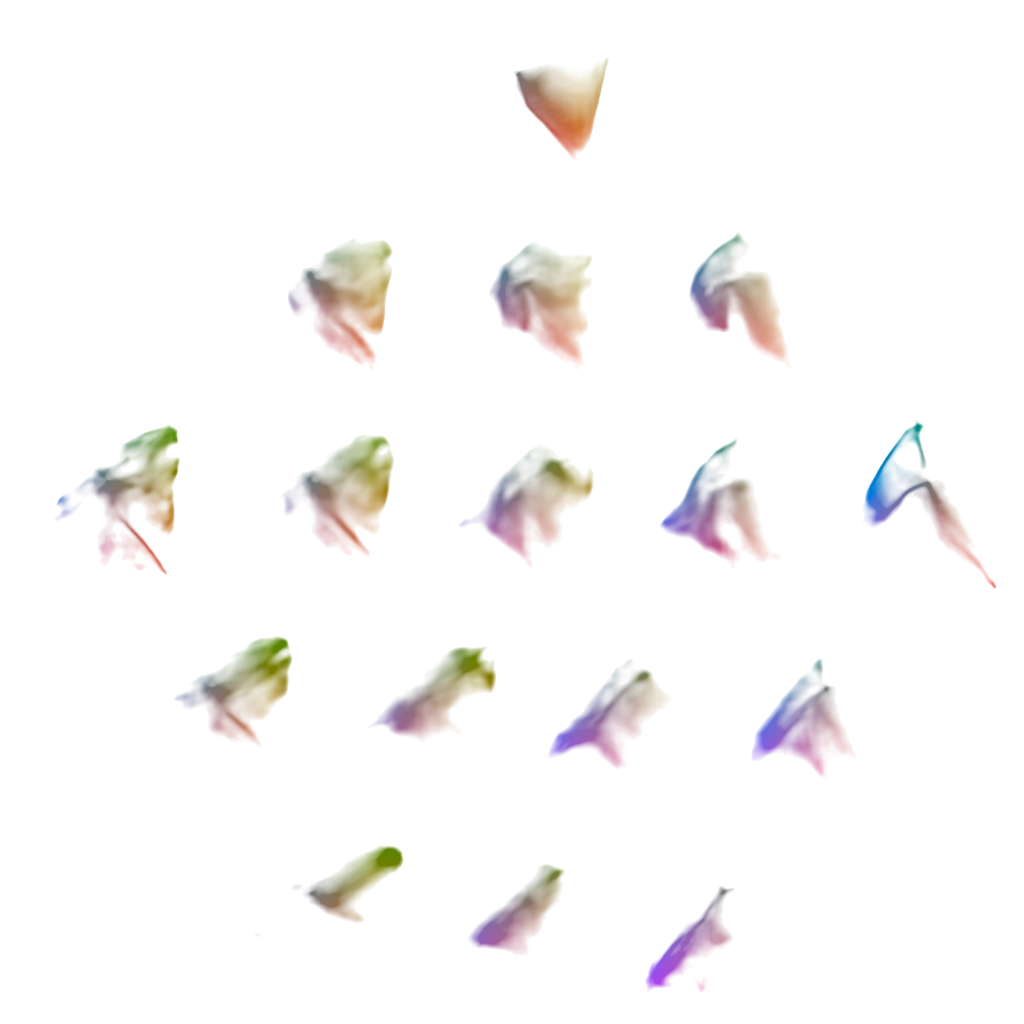}} & \multicolumn{3}{c}{\includegraphics[width=0.31\columnwidth]{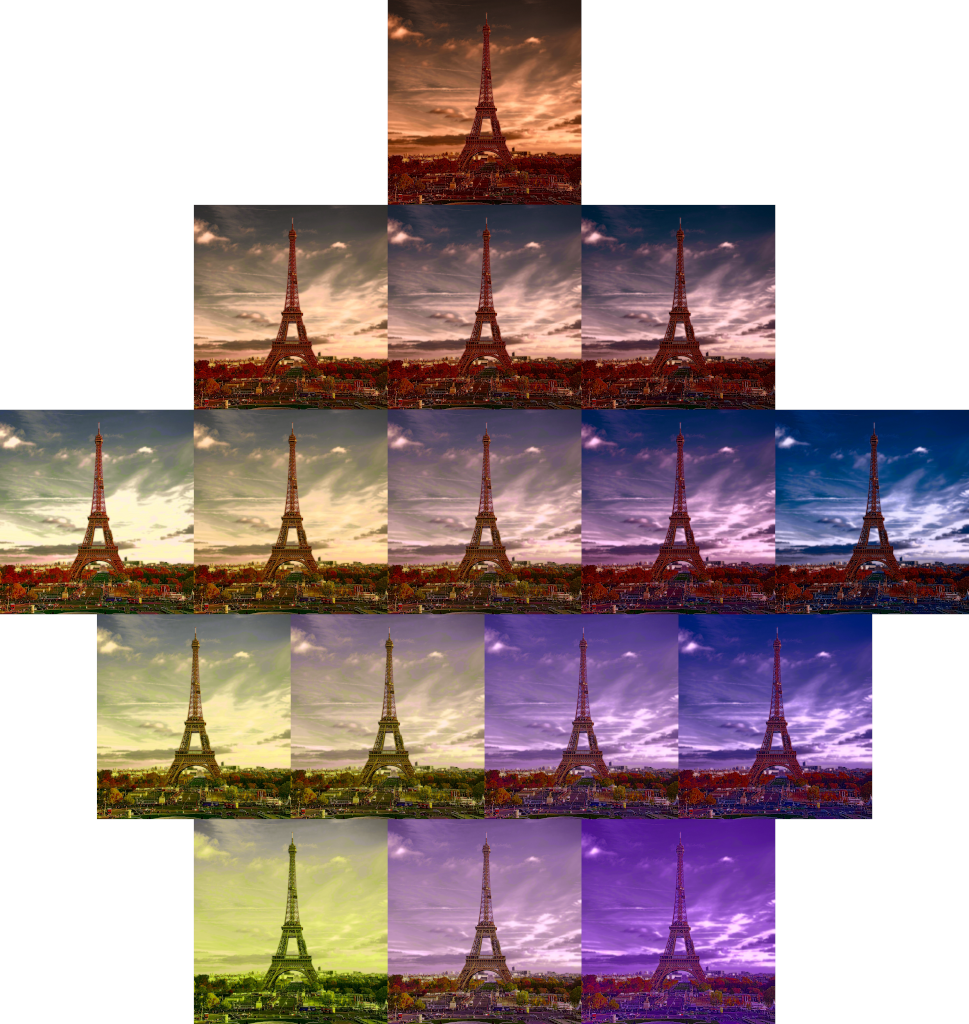}} \\
\end{tabular}
\caption{Color grading obtained by transferring the colors of $n=5$ images onto a target image. Results are shown in pentagons (Left: interpolated chrominance histograms; Right: corresponding transfer results). The images corresponding to the target chrominance histogram $\nu$ and to the histograms $\mu_{i}$ - which are interpolated to obtain a barycenter - are shown in top row. Each $\mu_{i}$ corresponds to a vertex of the pentagon in a clockwise order beginning with $i=1$ at the uppermost vertex. Each row presents the results for a different method, from top to bottom: GeomLoss, our model trained on synthetic shape contours (\emph{ContoursDS}) and our model trained on chrominance histograms from Flickr images (\emph{HistoDS})}
\label{fig:color_transfer_pentagon}
\end{figure}

After this color transfer step, similarly to \citep{bonneel2015sliced}, we apply a post-processing technique from \citep{rabin2011removing} using iterative guided filtering in order to reduce visual artefacts caused by the color transfer. We repeat this color transfer for each image shown in each of the pentagons of figure \ref{fig:color_transfer_pentagon} (last column). This figure presents a comparison of barycenter and color transfer results obtained with GeomLoss, our model trained on synthetic shape contours (\emph{ContoursDS}) and our model directly trained with chrominance histograms extracted from images from the Flickr dataset (\emph{HistoDS}). Even if predicted chrominance histograms are clearly better with \emph{HistoDS}, the predictions made with \emph{ContoursDS} are good enough to obtain a consistent and visually pleasing color transfer which is close to the one obtained using GeomLoss. Additional results are provided in appendix \ref{appendix:additionalresults}, figure \ref{fig:color_transfer_triangle}.

\subsection{Speed}
In order to assess computational times, we obtain average running time over $1000$ barycenter computations -- on average, our model predicts barycenters of two images in 0.0092 seconds. We compare the average speed of our model with GeomLoss in two different settings. The first one considers the full $512\times512$ images -- GeomLoss computes such barycenters in 1.41 seconds.
The second setting takes advantage of the sparsity of our images and only uses the 2D coordinates of the points with non-zero mass -- in this case, GeomLoss computes barycenters in 0.589 seconds. Our method provides nearly 64x speedup compared with this last approach. 
In comparison, an exact barycenter computation of two (sparse) measures using a network simplex~\citep{BPPH11} ranges from 4--80 seconds for typical shape contours images that contains few thousands of pixels carrying mass. The time required to compute barycenters using the method of ~\citep{claici2018stochastic} depends on the number of iterations, in our setting $100$ iterations with the inputs shown in figure \ref{figure:comparison_with_claici} require 37 hours while $50$ iterations are achieved in 14 hours.
A $512\times512$ Radon barycenter~\citep{bonneel2015sliced} requires 0.2 seconds for 720 projection directions, but remains far from the expected barycenter. 

\section{Discussion and conclusion}
\label{sec:discussion}

\begin{figure}[!h]
\vspace{1em}
\centering
\begin{tabular}{ccccc}
\rotatebox{90}{\tiny{GeomLoss}} 
&\includegraphics[width=0.085\columnwidth]{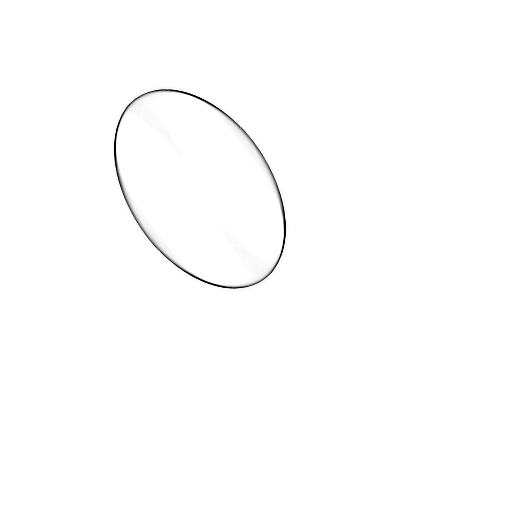}
&\includegraphics[width=0.085\columnwidth]{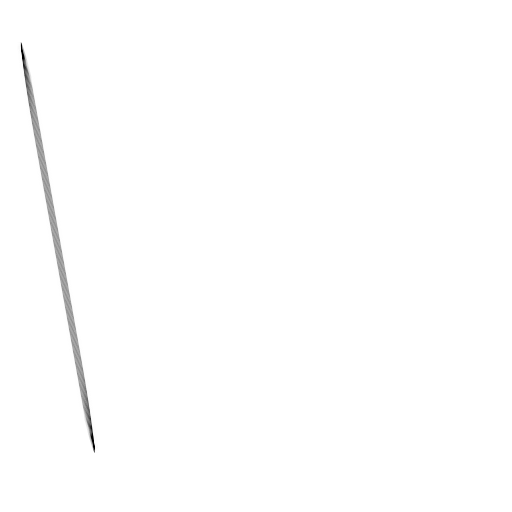} 
&\includegraphics[width=0.085\columnwidth]{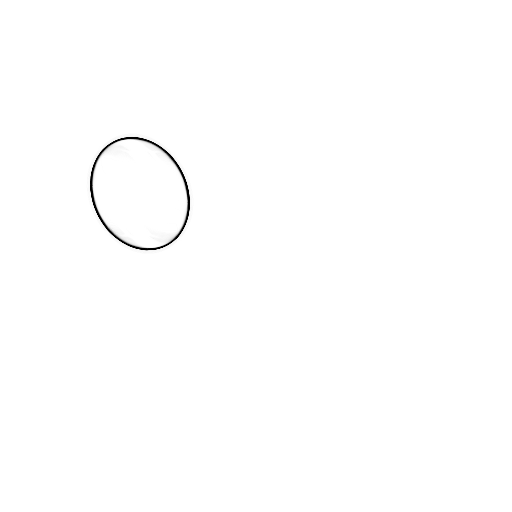}
&\includegraphics[width=0.085\columnwidth]{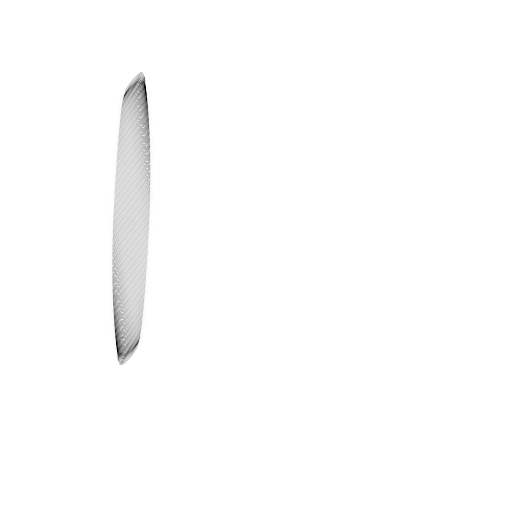} 
\\
\rotatebox{90}{\tiny{Ours}} 
&\includegraphics[width=0.085\columnwidth]{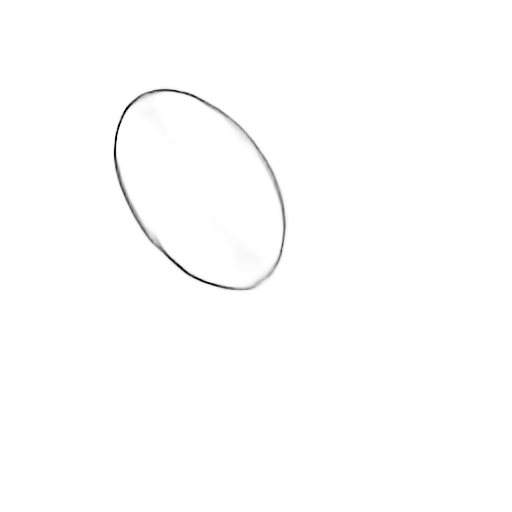}
&\includegraphics[width=0.085\columnwidth]{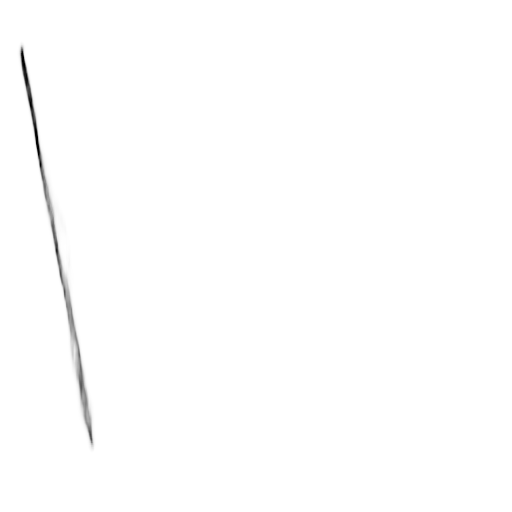}
&\includegraphics[width=0.085\columnwidth]{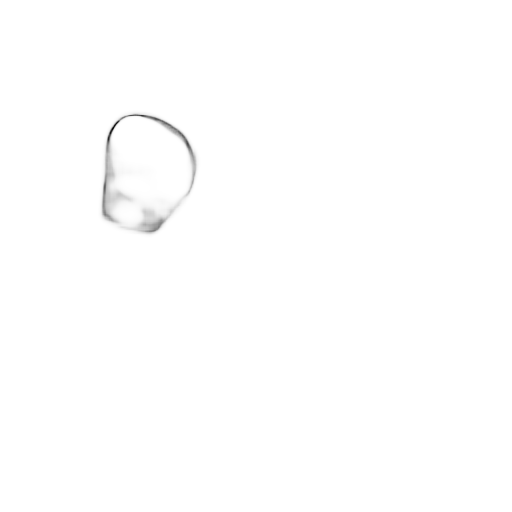}
&\includegraphics[width=0.085\columnwidth]{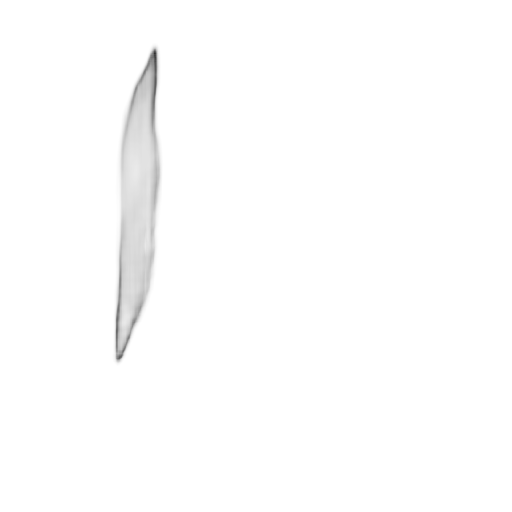}\\
& \tiny{2-circles} & \tiny{2-lines}  & \tiny{5-circles}  & \tiny{5-lines} \\
\end{tabular}
\caption{Wasserstein barycenters of sets of lines or ellipses should result in lines (resp. ellipses). Our prediction for two-way barycenters (here, with equal weights) of such shapes remains correct (left). However, the predicted barycenter is highly distorted for 5-way barycenters of simple shapes (right) although it remains plausible for more complex shapes (see Fig.~\ref{figure:3_5_interpolations})}
\label{fig:failure}
\end{figure}

While our method produces good approximation of Wasserstein barycenters of $n$ inputs, some shapes are surprisingly difficult to handle. The barycenter of simple translated and scaled shapes such as lines or ellipses should theoretically also be lines or ellipses, but are failure cases for our model (Fig.~\ref{fig:failure}), while more complex shapes are well handled (Fig.~\ref{figure:3_5_interpolations}).  In addition, we rely on a linearized barycenter to train our network \citep{Nader18,wang2013linear,Moosmuller20,Merigot20}, which incurs some error. This can be seen in appendix Sec.~\ref{sec:linearized},  Fig.~\ref{figure:geomloss_multi_iters_vs_ours}. While using more iterations of gradient descent yields more accurate results and removes this linearity, it also prevents  easy combination and makes the dataset generation intractable. 
Nevertheless, in many cases our DCNN is able to synthesize a barycenter from an arbitrary number of inputs. The main strength of our approach lies in its capacity to be trained from only $2$-inputs barycenters examples and to generalize to any number of inputs. We showed that the results exceeded the ones obtained by explicit Wasserstein Embedding computation while having a very low computation time. We hope our fast approach will accelerate the adoption of optimal transport in machine learning applications.

\paragraph{Acknowledgements}
This work was granted access to the HPC resources of IDRIS under the allocations 2020-AD011011538 and 2020-AD011012218 made by GENCI. We also thank the authors of all the images used in our color transfer figures. 

\paragraph{Funding}
Partial financial support was received from the ANR ROOT (RegressiOn with Optimal Transport): ANR-16-CE23-0009.

\paragraph{Conflicts of interest / Competing interests.}
The authors have no conflicts of interest to declare that are relevant to the content of this article.

\paragraph{Code availability}
Our implementation is publicly available at \url{https://github.com/jlacombe/learning-to-generate-wasserstein-barycenters}

\bibliographystyle{spbasic}
\bibliography{main}

\begin{thebibliography}{52}
\providecommand{\natexlab}[1]{#1}
\providecommand{\url}[1]{{#1}}
\providecommand{\urlprefix}{URL }
\expandafter\ifx\csname urlstyle\endcsname\relax
  \providecommand{\doi}[1]{DOI~\discretionary{}{}{}#1}\else
  \providecommand{\doi}{DOI~\discretionary{}{}{}\begingroup
  \urlstyle{rm}\Url}\fi
\providecommand{\eprint}[2][]{\url{#2}}

\bibitem[{Amos et~al.(2017)Amos, Xu, and Kolter}]{amos2017input}
Amos B, Xu L, Kolter JZ (2017) Input convex neural networks. In: International
  Conference on Machine Learning, pp 146--155

\bibitem[{Andoni et~al.(2008)Andoni, Indyk, and Krauthgamer}]{andoni2008earth}
Andoni A, Indyk P, Krauthgamer R (2008) Earth mover distance over
  high-dimensional spaces. In: SODA, vol~8, pp 343--352

\bibitem[{Andoni et~al.(2016)Andoni, Naor, and
  Neiman}]{andoni2016impossibility}
Andoni A, Naor A, Neiman O (2016) Impossibility of sketching of the 3d
  transportation metric with quadratic cost. In: 43rd International Colloquium
  on Automata, Languages, and Programming (ICALP 2016), Schloss
  Dagstuhl-Leibniz-Zentrum fuer Informatik

\bibitem[{Arjovsky et~al.(2017)Arjovsky, Chintala, and Bottou}]{arjovsky2017}
Arjovsky M, Chintala S, Bottou L (2017) Wasserstein gan. \eprint{1701.07875}

\bibitem[{Backhoff-Veraguas et~al.(2018)Backhoff-Veraguas, Fontbona, Rios, and
  Tobar}]{backhoff2018bayesian}
Backhoff-Veraguas J, Fontbona J, Rios G, Tobar F (2018) Bayesian learning with
  wasserstein barycenters. arXiv preprint arXiv:180510833

\bibitem[{Bigot et~al.(2017)Bigot, Gouet, Klein, L{\'o}pez
  et~al.}]{bigot2017geodesic}
Bigot J, Gouet R, Klein T, L{\'o}pez A, et~al. (2017) Geodesic pca in the
  wasserstein space by convex pca. In: Annales de l'Institut Henri
  Poincar{\'e}, Probabilit{\'e}s et Statistiques, Institut Henri Poincar{\'e},
  vol~53, pp 1--26

\bibitem[{Bonneel et~al.(2011)Bonneel, van~de Panne, Paris, and
  Heidrich}]{BPPH11}
Bonneel N, van~de Panne M, Paris S, Heidrich W (2011) {Displacement
  Interpolation Using Lagrangian Mass Transport}. ACM Transactions on Graphics
  (SIGGRAPH ASIA 2011) 30(6)

\bibitem[{Bonneel et~al.(2015)Bonneel, Rabin, Peyr{\'e}, and
  Pfister}]{bonneel2015sliced}
Bonneel N, Rabin J, Peyr{\'e} G, Pfister H (2015) Sliced and radon wasserstein
  barycenters of measures. Journal of Mathematical Imaging and Vision
  51(1):22--45

\bibitem[{Bonneel et~al.(2016)Bonneel, Peyr{\'e}, and Cuturi}]{BPC16}
Bonneel N, Peyr{\'e} G, Cuturi M (2016) {Wasserstein Barycentric Coordinates:
  Histogram Regression Using Optimal Transport}. ACM Transactions on Graphics
  (SIGGRAPH 2016) 35(4)

\bibitem[{Burger et~al.(2012)Burger, Schuler, and Harmeling}]{burger2012image}
Burger HC, Schuler CJ, Harmeling S (2012) Image denoising: Can plain neural
  networks compete with bm3d? In: 2012 IEEE conference on computer vision and
  pattern recognition, IEEE, pp 2392--2399

\bibitem[{Claici et~al.(2018)Claici, Chien, and Solomon}]{claici2018stochastic}
Claici S, Chien E, Solomon J (2018) Stochastic wasserstein barycenters. arXiv
  preprint arXiv:180205757

\bibitem[{Courty et~al.(2014)Courty, Flamary, and Tuia}]{courty2014domain}
Courty N, Flamary R, Tuia D (2014) Domain adaptation with regularized optimal
  transport. In: Joint European Conference on Machine Learning and Knowledge
  Discovery in Databases, Springer, pp 274--289

\bibitem[{Courty et~al.(2017)Courty, Flamary, and Ducoffe}]{courty2017learning}
Courty N, Flamary R, Ducoffe M (2017) Learning wasserstein embeddings. arXiv
  preprint arXiv:171007457

\bibitem[{Cuturi(2013)}]{cuturi2013sinkhorn}
Cuturi M (2013) Sinkhorn distances: Lightspeed computation of optimal
  transport. In: Advances in neural information processing systems, pp
  2292--2300

\bibitem[{Dognin et~al.(2019)Dognin, Melnyk, Mroueh, Ross, Santos, and
  Sercu}]{dognin2019wasserstein}
Dognin P, Melnyk I, Mroueh Y, Ross J, Santos CD, Sercu T (2019) Wasserstein
  barycenter model ensembling. arXiv preprint arXiv:190204999

\bibitem[{Domazakis et~al.(2020)Domazakis, Drivaliaris, Koukoulas, Papayiannis,
  Tsekrekos, and Yannacopoulos}]{domazakis2020clustering}
Domazakis G, Drivaliaris D, Koukoulas S, Papayiannis G, Tsekrekos A,
  Yannacopoulos A (2020) Clustering measure-valued data with wasserstein
  barycenters. arXiv preprint arXiv:191211801

\bibitem[{Fan et~al.(2020)Fan, Taghvaei, and Chen}]{fan2020scalable}
Fan J, Taghvaei A, Chen Y (2020) Scalable computations of wasserstein
  barycenter via input convex neural networks. arXiv preprint arXiv:200704462

\bibitem[{Feydy(2019)}]{geomloss}
Feydy J (2019) Geometric loss functions between sampled measures, images and
  volumes. \urlprefix\url{https://www.kernel-operations.io/geomloss/}

\bibitem[{Feydy et~al.(2018)Feydy, S{\'e}journ{\'e}, Vialard, Amari,
  Trouv{\'e}, and Peyr{\'e}}]{feydy2018interpolating}
Feydy J, S{\'e}journ{\'e} T, Vialard FX, Amari SI, Trouv{\'e} A, Peyr{\'e} G
  (2018) Interpolating between optimal transport and mmd using sinkhorn
  divergences. arXiv preprint arXiv:181008278

\bibitem[{Feydy et~al.(2019)Feydy, Roussillon, Trouv{\'e}, and
  Gori}]{feydy2019fast}
Feydy J, Roussillon P, Trouv{\'e} A, Gori P (2019) Fast and scalable optimal
  transport for brain tractograms. In: International Conference on Medical
  Image Computing and Computer-Assisted Intervention, Springer, pp 636--644

\bibitem[{Frogner et~al.(2019)Frogner, Mirzazadeh, and
  Solomon}]{frogner2019learning}
Frogner C, Mirzazadeh F, Solomon J (2019) Learning embeddings into entropic
  wasserstein spaces. arXiv preprint arXiv:190503329

\bibitem[{Genevay et~al.(2017)Genevay, Peyr{\'e}, and
  Cuturi}]{genevay2017learning}
Genevay A, Peyr{\'e} G, Cuturi M (2017) Learning generative models with
  sinkhorn divergences. arXiv preprint arXiv:170600292

\bibitem[{Google(2020)}]{quickdraw}
Google I (2020) The quick, draw! dataset.
  \urlprefix\url{https://github.com/googlecreativelab/quickdraw-dataset}

\bibitem[{Heitz et~al.(2019)Heitz, Bonneel, Coeurjolly, Cuturi, and
  Peyr{\'e}}]{HBCCP19}
Heitz M, Bonneel N, Coeurjolly D, Cuturi M, Peyr{\'e} G (2019) {Ground Metric
  Learning on Graphs}. Tech. Rep. arXiv:1911.03117

\bibitem[{Kantorovich(1942)}]{kantorovich1942transfer}
Kantorovich L (1942) On the transfer of masses (in russian). In: Doklady
  Akademii Nauk, vol~37, pp 227--229

\bibitem[{Lai et~al.(2017)Lai, Huang, Ahuja, and Yang}]{lai2017deep}
Lai WS, Huang JB, Ahuja N, Yang MH (2017) Deep laplacian pyramid networks for
  fast and accurate super-resolution. In: Proceedings of the IEEE conference on
  computer vision and pattern recognition, pp 624--632

\bibitem[{Ledig et~al.(2017)Ledig, Theis, Husz{\'a}r, Caballero, Cunningham,
  Acosta, Aitken, Tejani, Totz, Wang et~al.}]{ledig2017photo}
Ledig C, Theis L, Husz{\'a}r F, Caballero J, Cunningham A, Acosta A, Aitken A,
  Tejani A, Totz J, Wang Z, et~al. (2017) Photo-realistic single image
  super-resolution using a generative adversarial network. In: Proceedings of
  the IEEE conference on computer vision and pattern recognition, pp 4681--4690

\bibitem[{Lefkimmiatis(2017)}]{lefkimmiatis2017non}
Lefkimmiatis S (2017) Non-local color image denoising with convolutional neural
  networks. In: Proceedings of the IEEE Conference on Computer Vision and
  Pattern Recognition, pp 3587--3596

\bibitem[{Liu et~al.(2018)Liu, Reda, Shih, Wang, Tao, and
  Catanzaro}]{liu2018image}
Liu G, Reda FA, Shih KJ, Wang TC, Tao A, Catanzaro B (2018) Image inpainting
  for irregular holes using partial convolutions. In: Proceedings of the
  European Conference on Computer Vision (ECCV), pp 85--100

\bibitem[{Liutkus et~al.(2019)Liutkus, Simsekli, Majewski, Durmus, and
  St{\"o}ter}]{liutkus2019sliced}
Liutkus A, Simsekli U, Majewski S, Durmus A, St{\"o}ter FR (2019)
  Sliced-wasserstein flows: Nonparametric generative modeling via optimal
  transport and diffusions. In: International Conference on Machine Learning,
  PMLR, pp 4104--4113

\bibitem[{Loshchilov and Hutter(2016)}]{loshchilov2016sgdr}
Loshchilov I, Hutter F (2016) Sgdr: Stochastic gradient descent with warm
  restarts. arXiv preprint arXiv:160803983

\bibitem[{M\'erigot et~al.(2020)M\'erigot, Delalande, and Chazal}]{Merigot20}
M\'erigot Q, Delalande A, Chazal F (2020) Quantitative stability of optimal
  transport maps and linearization of the 2-wasserstein space. Proceedings of
  Machine Learning Research, vol 108, pp 3186--3196

\bibitem[{Metelli et~al.(2019)Metelli, Likmeta, and
  Restelli}]{metelli2019propagating}
Metelli AM, Likmeta A, Restelli M (2019) Propagating uncertainty in
  reinforcement learning via wasserstein barycenters. In: Advances in Neural
  Information Processing Systems, pp 4333--4345

\bibitem[{Mi et~al.(2018)Mi, Zhang, Gu, and Wang}]{mi2018variational}
Mi L, Zhang W, Gu X, Wang Y (2018) Variational {W}asserstein clustering. In:
  Proceedings of the European Conference on Computer Vision (ECCV), pp 322--337

\bibitem[{Moosmüller and Cloninger(2020)}]{Moosmuller20}
Moosmüller C, Cloninger A (2020) Linear optimal transport embedding: Provable
  fast wasserstein distance computation and classification for nonlinear
  problems. \eprint{2008.09165}

\bibitem[{Nader and Guennebaud(2018)}]{Nader18}
Nader G, Guennebaud G (2018) Instant transport maps on 2d grids. ACM Trans
  Graph 37(6)

\bibitem[{Nane et~al.(1996)Nane, Nayar, and Murase}]{nane1996columbia}
Nane S, Nayar S, Murase H (1996) Columbia object image library: Coil-20. Dept
  Comp Sci, Columbia University, New York, Tech Rep

\bibitem[{Peyr{\'e} et~al.(2019)Peyr{\'e}, Cuturi
  et~al.}]{peyre2019computational}
Peyr{\'e} G, Cuturi M, et~al. (2019) Computational optimal transport.
  Foundations and Trends{\textregistered} in Machine Learning 11(5-6):355--607

\bibitem[{Rabin et~al.(2011{\natexlab{a}})Rabin, Delon, and
  Gousseau}]{rabin2011removing}
Rabin J, Delon J, Gousseau Y (2011{\natexlab{a}}) Removing artefacts from color
  and contrast modifications. IEEE Transactions on Image Processing
  20(11):3073--3085

\bibitem[{Rabin et~al.(2011{\natexlab{b}})Rabin, Peyr{\'e}, Delon, and
  Bernot}]{rabin2011wasserstein}
Rabin J, Peyr{\'e} G, Delon J, Bernot M (2011{\natexlab{b}}) Wasserstein
  barycenter and its application to texture mixing. In: International
  Conference on Scale Space and Variational Methods in Computer Vision,
  Springer, pp 435--446

\bibitem[{Reinhard and Pouli(2011)}]{reinhard2011colour}
Reinhard E, Pouli T (2011) Colour spaces for colour transfer. In: International
  Workshop on Computational Color Imaging, Springer, pp 1--15

\bibitem[{Rolet et~al.(2016)Rolet, Cuturi, and Peyr{\'e}}]{rolet2016fast}
Rolet A, Cuturi M, Peyr{\'e} G (2016) Fast dictionary learning with a smoothed
  wasserstein loss. In: Artificial Intelligence and Statistics, pp 630--638

\bibitem[{Ronneberger et~al.(2015)Ronneberger, Fischer, and
  Brox}]{ronneberger2015u}
Ronneberger O, Fischer P, Brox T (2015) U-net: Convolutional networks for
  biomedical image segmentation. In: International Conference on Medical image
  computing and computer-assisted intervention, Springer, pp 234--241

\bibitem[{Schmitz et~al.(2018)Schmitz, Heitz, Bonneel, Mboula, Coeurjolly,
  Cuturi, Peyr{\'e}, and Starck}]{SHBMCCPS18}
Schmitz MA, Heitz M, Bonneel N, Mboula FMN, Coeurjolly D, Cuturi M, Peyr{\'e}
  G, Starck JL (2018) Wasserstein dictionary learning: Optimal transport-based
  unsupervised non-linear dictionary learning. SIAM Journal on Imaging Sciences
  11(1)

\bibitem[{Schmitzer(2019)}]{schmitzer2019stabilized}
Schmitzer B (2019) Stabilized sparse scaling algorithms for entropy regularized
  transport problems. SIAM Journal on Scientific Computing 41(3):A1443--A1481

\bibitem[{Solomon et~al.(2015)Solomon, De~Goes, Peyr{\'e}, Cuturi, Butscher,
  Nguyen, Du, and Guibas}]{solomon2015convolutional}
Solomon J, De~Goes F, Peyr{\'e} G, Cuturi M, Butscher A, Nguyen A, Du T, Guibas
  L (2015) Convolutional wasserstein distances: Efficient optimal
  transportation on geometric domains. ACM Transactions on Graphics (TOG)
  34(4):1--11

\bibitem[{Tai et~al.(2017)Tai, Yang, and Liu}]{tai2017image}
Tai Y, Yang J, Liu X (2017) Image super-resolution via deep recursive residual
  network. In: Proceedings of the IEEE conference on computer vision and
  pattern recognition, pp 3147--3155

\bibitem[{Ulyanov et~al.(2016)Ulyanov, Vedaldi, and
  Lempitsky}]{ulyanov2016instance}
Ulyanov D, Vedaldi A, Lempitsky V (2016) Instance normalization: The missing
  ingredient for fast stylization. arXiv preprint arXiv:160708022

\bibitem[{Ulyanov et~al.(2018)Ulyanov, Vedaldi, and
  Lempitsky}]{ulyanov2018deep}
Ulyanov D, Vedaldi A, Lempitsky V (2018) Deep image prior. In: Proceedings of
  the IEEE Conference on Computer Vision and Pattern Recognition, pp 9446--9454

\bibitem[{Wang et~al.(2013)Wang, Slep{\v{c}}ev, Basu, Ozolek, and
  Rohde}]{wang2013linear}
Wang W, Slep{\v{c}}ev D, Basu S, Ozolek JA, Rohde GK (2013) A linear optimal
  transportation framework for quantifying and visualizing variations in sets
  of images. International journal of computer vision 101(2):254--269

\bibitem[{Xie et~al.(2012)Xie, Xu, and Chen}]{xie2012image}
Xie J, Xu L, Chen E (2012) Image denoising and inpainting with deep neural
  networks. In: Advances in neural information processing systems, pp 341--349

\bibitem[{Yeh et~al.(2017)Yeh, Chen, Yian~Lim, Schwing, Hasegawa-Johnson, and
  Do}]{yeh2017semantic}
Yeh RA, Chen C, Yian~Lim T, Schwing AG, Hasegawa-Johnson M, Do MN (2017)
  Semantic image inpainting with deep generative models. In: Proceedings of the
  IEEE conference on computer vision and pattern recognition, pp 5485--5493

\end{thebibliography}

\appendix
\section{Learning Strategy}
\label{appendix:training_details}

\begin{figure}[ht]
\vskip 0.2in
\begin{center}
\centerline{\includegraphics[width=0.75\columnwidth]{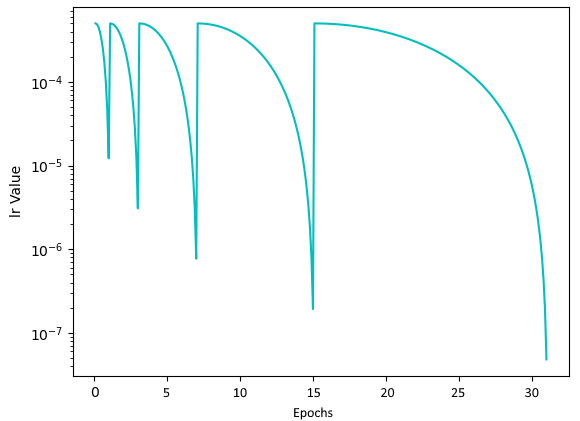}}
\caption{Learning rate schedule used to train our models, following the SGDR method described by \citet{loshchilov2016sgdr}. Our training runs for a total of $31$ epochs. Compared to a constant learning rate or to stepwise schedules, SGDR has empirically shown a better convergence in our context}
\label{figure:evo_lr}
\end{center}
\vskip -0.2in
\end{figure}

Instead of using a fixed learning rate or a decreasing learning rate, we choose a learning rate schedule with warm restart as proposed by~\cite{loshchilov2016sgdr}. The learning schedule is shown in Figure \ref{figure:evo_lr}: the learning rate decreased and is periodically restarted to its initial value, the period increasing as the number of epochs grows. This schedule was chosen after comparing with stepwise schedules or constant learning rates and yielded better convergence in practice.

\section{Additional results}
\label{appendix:additionalresults}
\begin{figure}[ht]
\centering

\begin{tabular}{@{}c@{\hspace{2mm}}c@{}c@{}c@{}c@{}c@{}c}

\raisebox{2\normalbaselineskip}[0pt][0pt]{\rotatebox[origin=c]{90}{\hspace{2mm}\scriptsize GeomLoss}}
& \includegraphics[width=0.125\columnwidth]{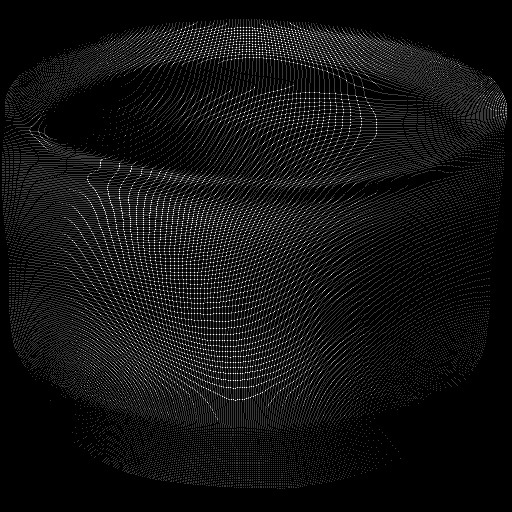}
& \includegraphics[width=0.125\columnwidth]{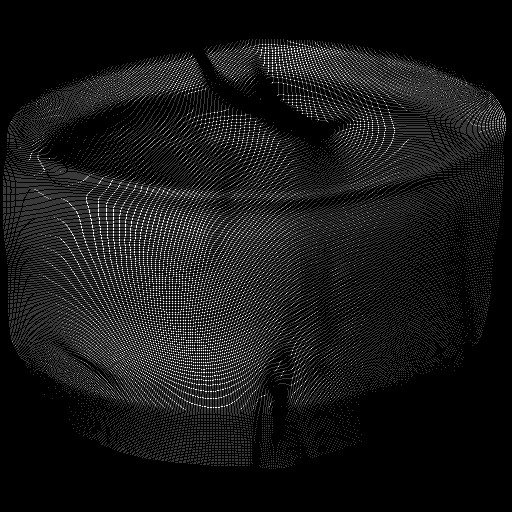}
& \includegraphics[width=0.125\columnwidth]{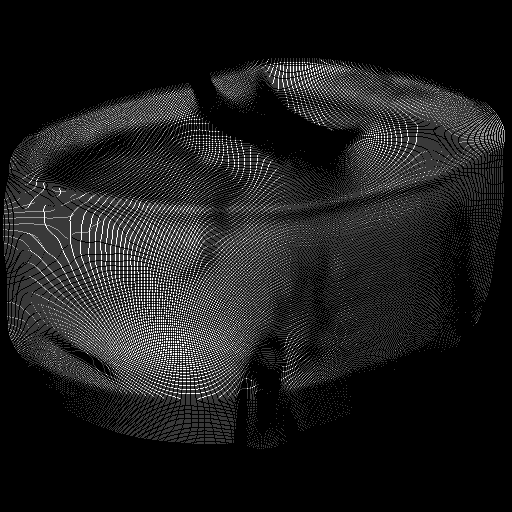}
& \includegraphics[width=0.125\columnwidth]{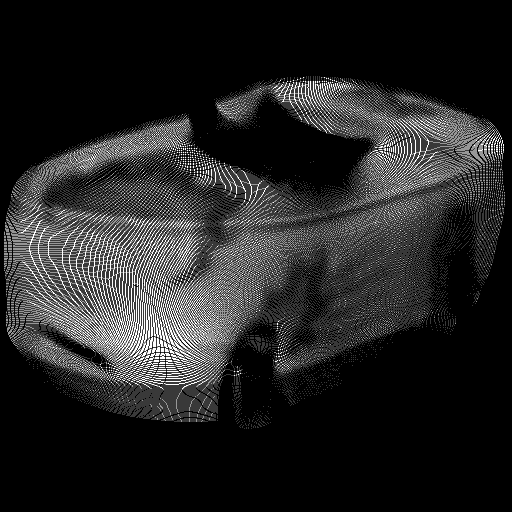}
& \includegraphics[width=0.125\columnwidth]{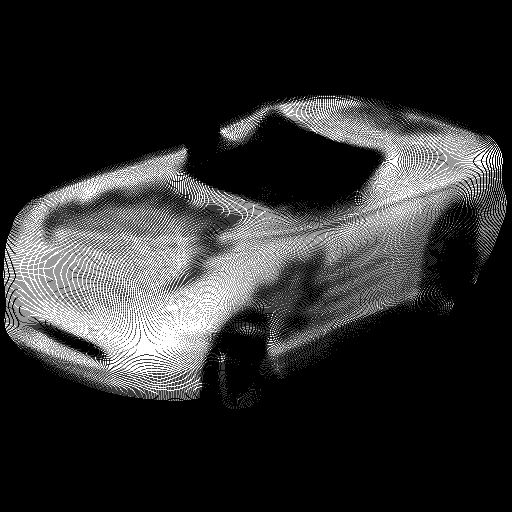}
& \includegraphics[width=0.125\columnwidth]{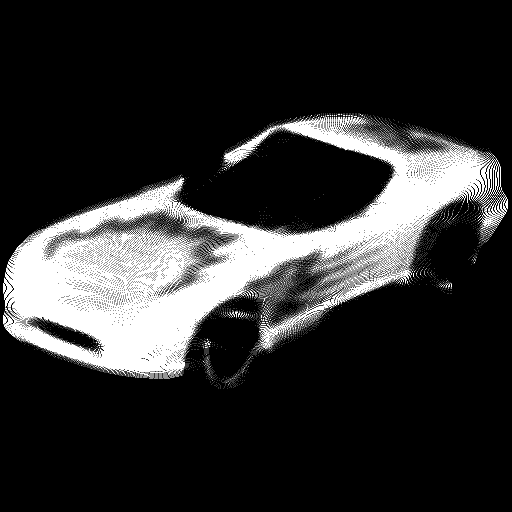} \\

\raisebox{1.8\normalbaselineskip}[0pt][0pt]{\rotatebox[origin=c]{90}{\hspace{3mm}\scriptsize Ours}} 
& \includegraphics[width=0.125\columnwidth]{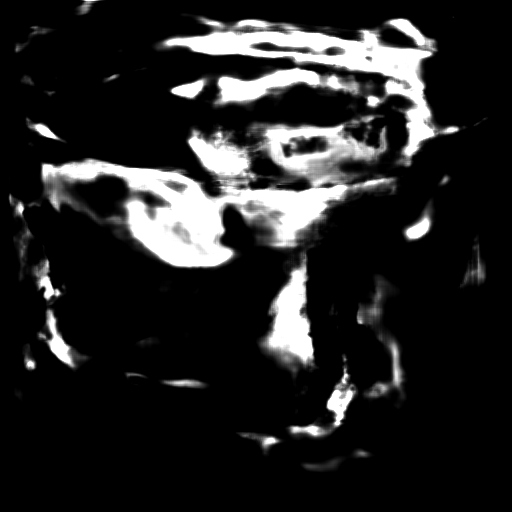}
& \includegraphics[width=0.125\columnwidth]{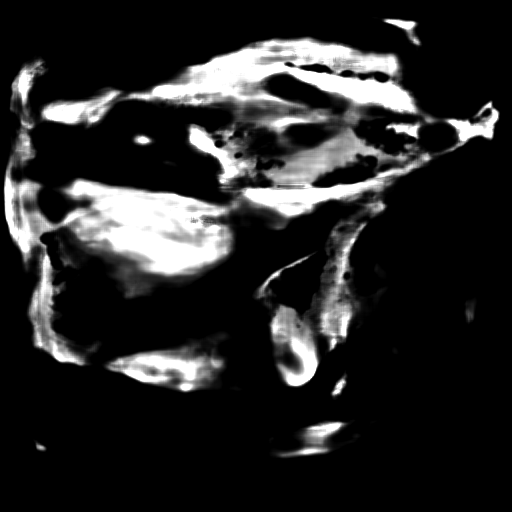}
& \includegraphics[width=0.125\columnwidth]{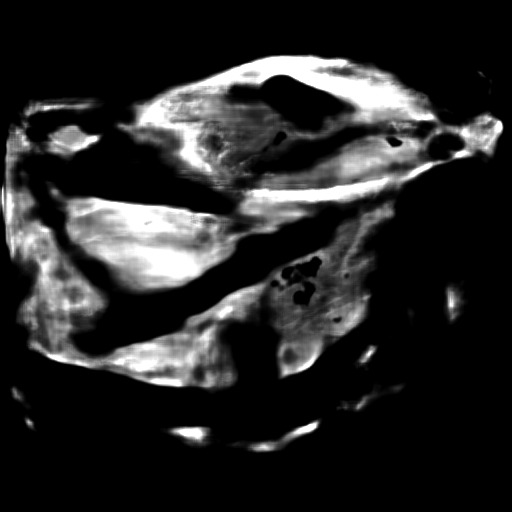}
& \includegraphics[width=0.125\columnwidth]{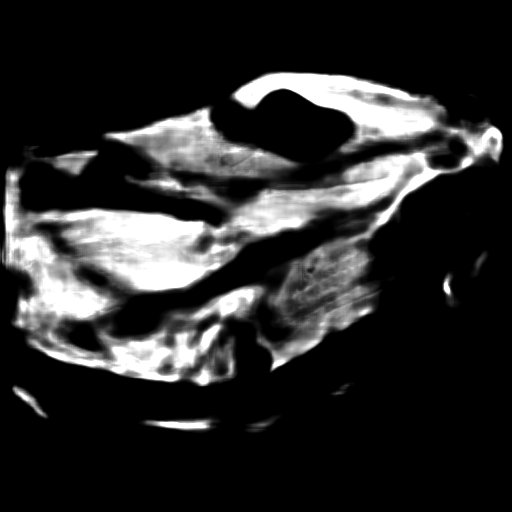}
& \includegraphics[width=0.125\columnwidth]{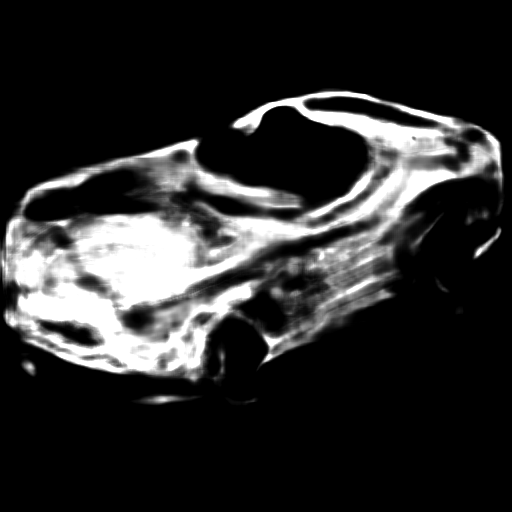}
& \includegraphics[width=0.125\columnwidth]{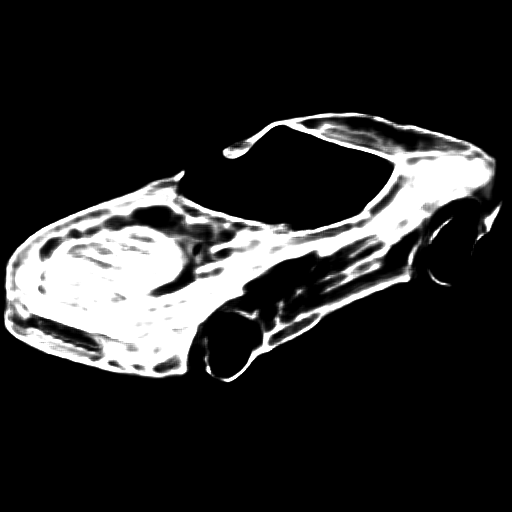} \\
\end{tabular}

\caption{Additional interpolations between two $512\times512$ images from the \emph{Coil20} dataset using GeomLoss and our model. For visualization purposes, white values represent high mass concentration, while dark values represent low mass concentration}
\label{figure:coil20_cup_car}
\end{figure}

\begin{figure}[ht]
\vskip 0.2in
\begin{center}
\centerline{\includegraphics[width=0.6\columnwidth]{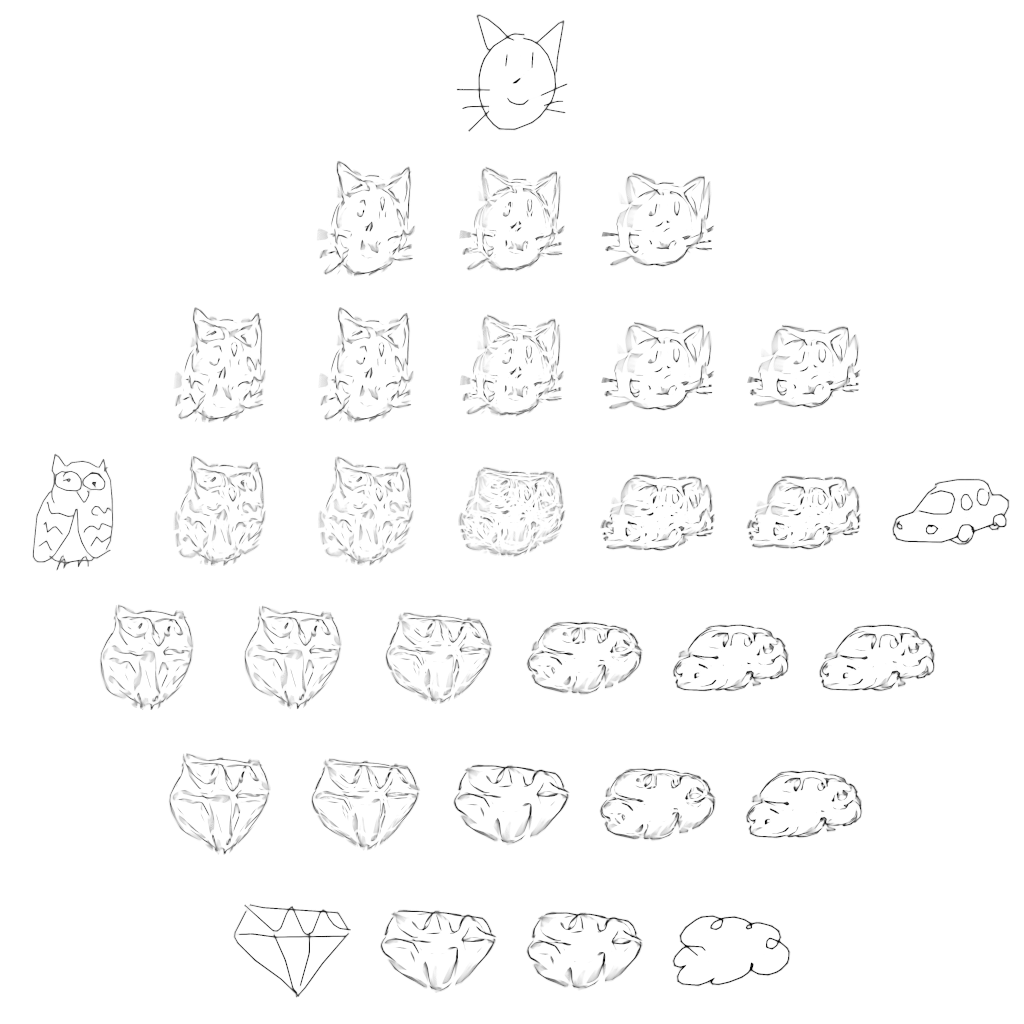}}
\centerline{\includegraphics[width=0.6\columnwidth]{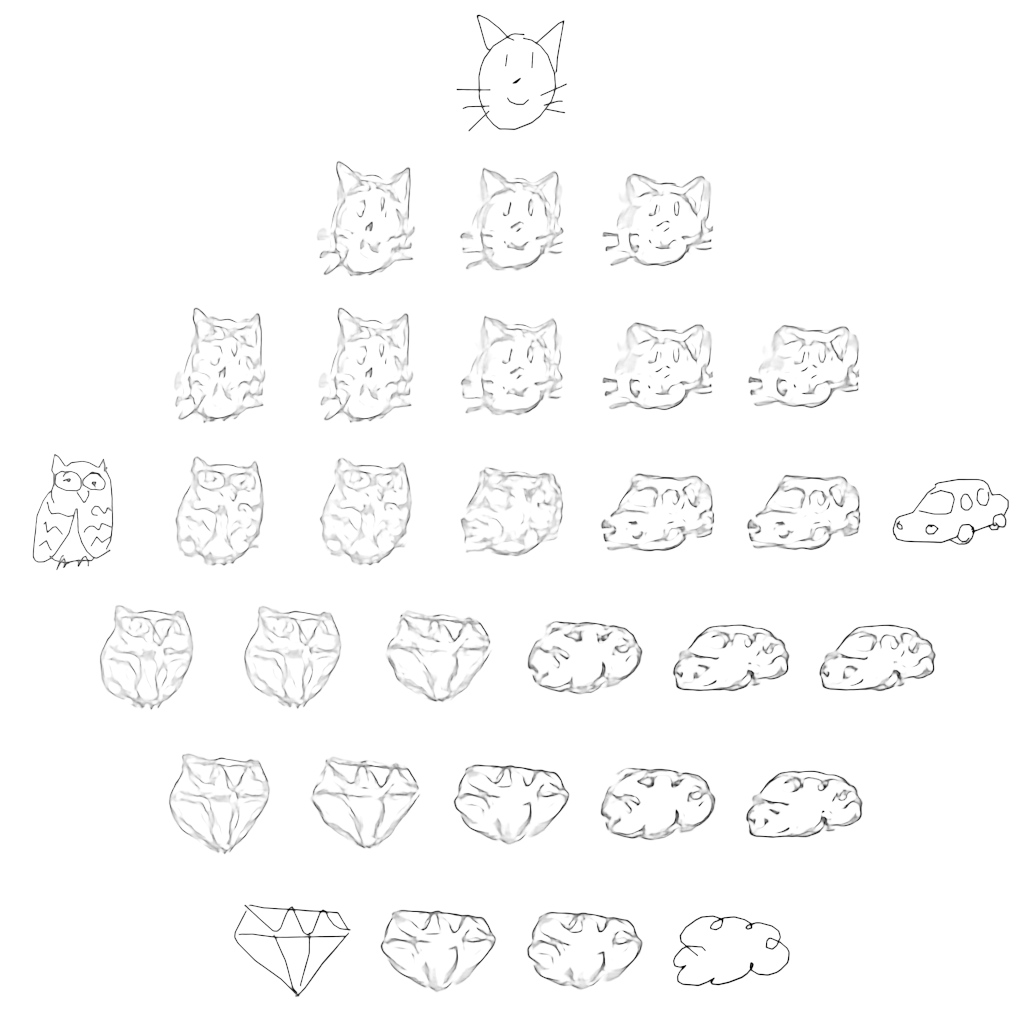}}
\caption{Interpolations between 5 inputs from \emph{Quick, Draw!}, shown as pentagons. Left pentagon corresponds to GeomLoss barycenters while the right one shows predictions of our model trained on our synthetic dataset}
\label{figure:diff_pentagon}
\end{center}
\vskip -0.2in
\end{figure}

\begin{figure}
\centering
\begin{tabular}{cc}
\includegraphics[width=5cm]{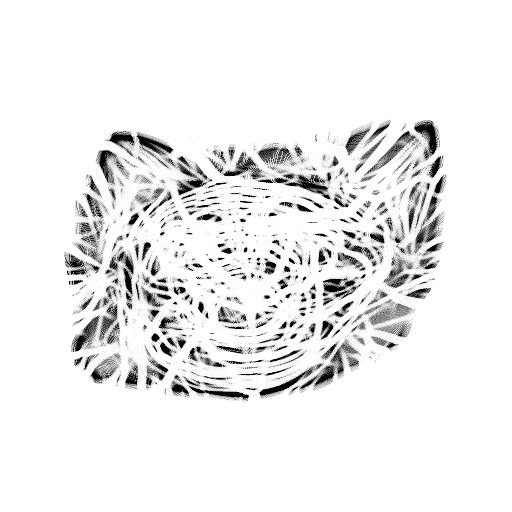} &
\includegraphics[width=5cm]{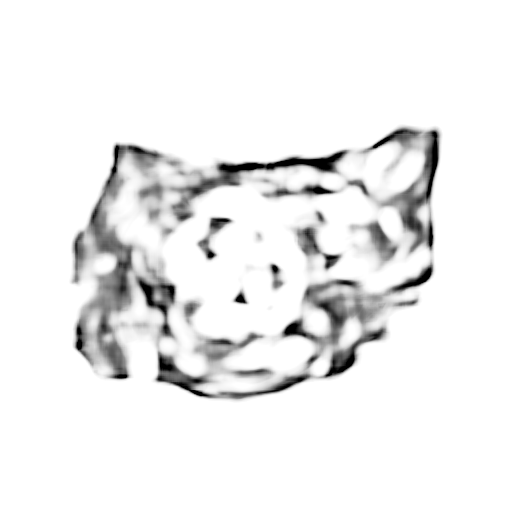} \\
GeomLoss & Our DCNN
\end{tabular}
\caption{Stress test. We predict a barycenter of 100 cats of the \emph{Quick, Draw!} dataset, with equal weights}
\label{fig:100way}
\end{figure}

\begin{figure}[h!]
\vspace{1em}
\centering
\begin{tabular}{@{}c@{}c@{}c@{}c@{}c@{}c@{}c@{}c@{}c@{}}

& \scriptsize{$\nu$} & \scriptsize{$\mu_{1}$} & \scriptsize{$\mu_{2}$} & \scriptsize{$\mu_{3}$} & & & & \\

& \includegraphics[height=0.065\textheight]{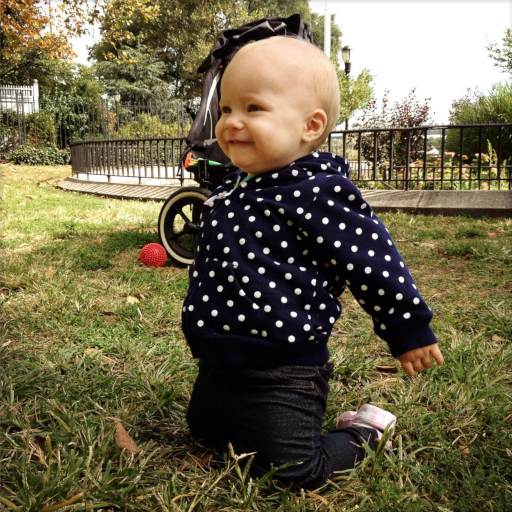} & \includegraphics[height=0.065\textheight]{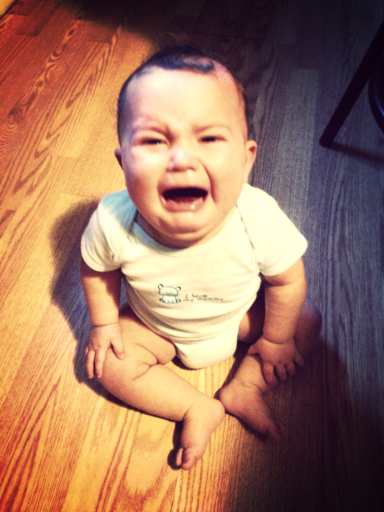} & \includegraphics[height=0.065\textheight]{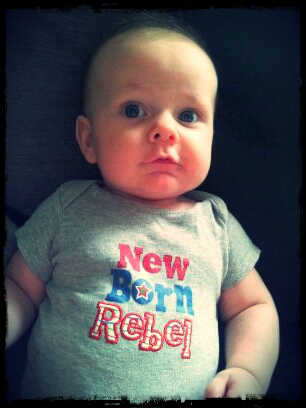} & \includegraphics[height=0.065\textheight]{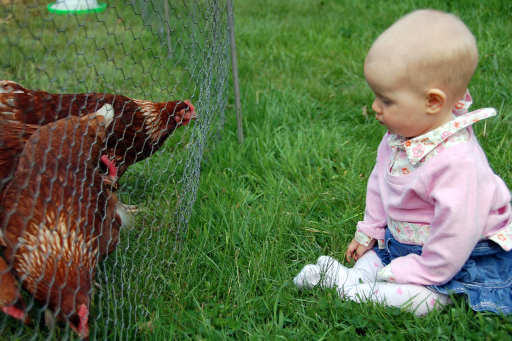} & & & & \\

\raisebox{7\normalbaselineskip}[0pt][0pt]{\rotatebox[origin=c]{90}{\tiny GeomLoss}} & \multicolumn{4}{c}{\includegraphics[width=0.31\columnwidth]{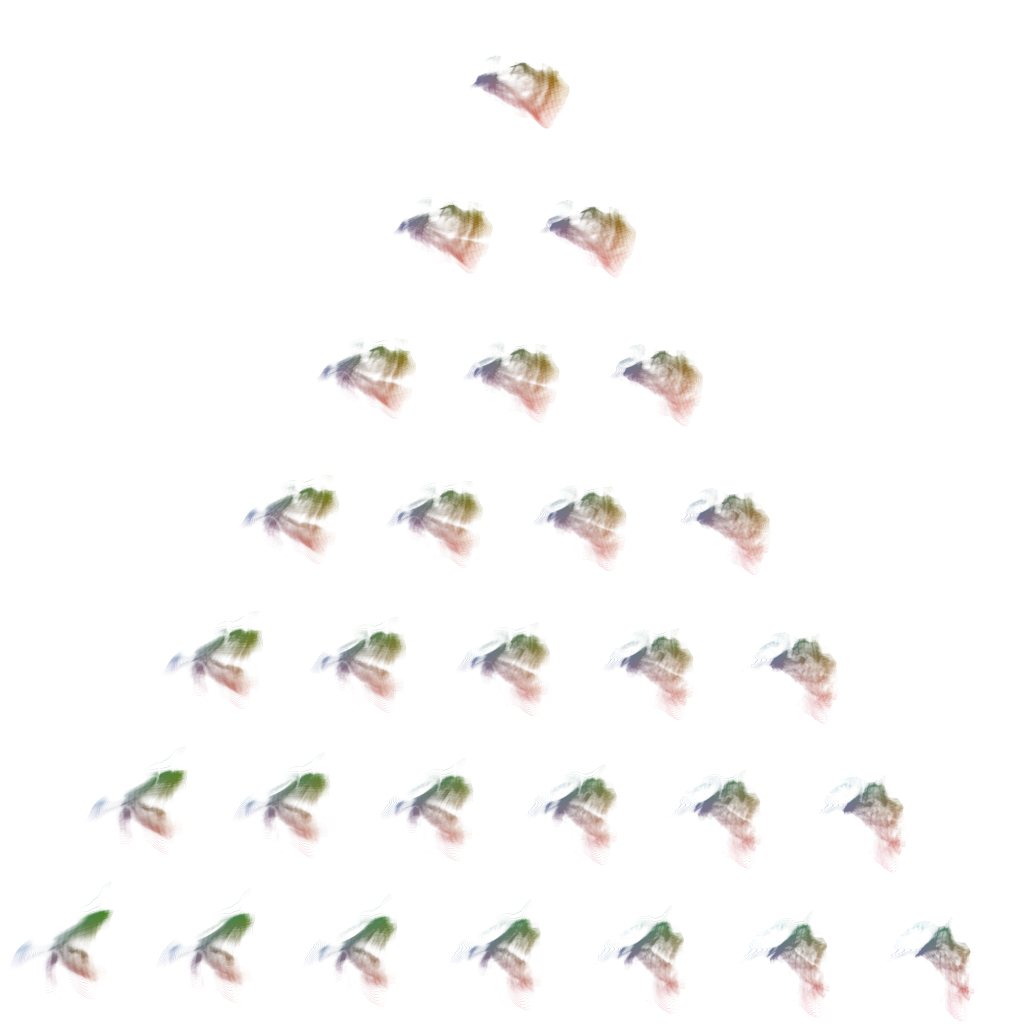}} & \multicolumn{4}{c}{\includegraphics[width=0.31\columnwidth]{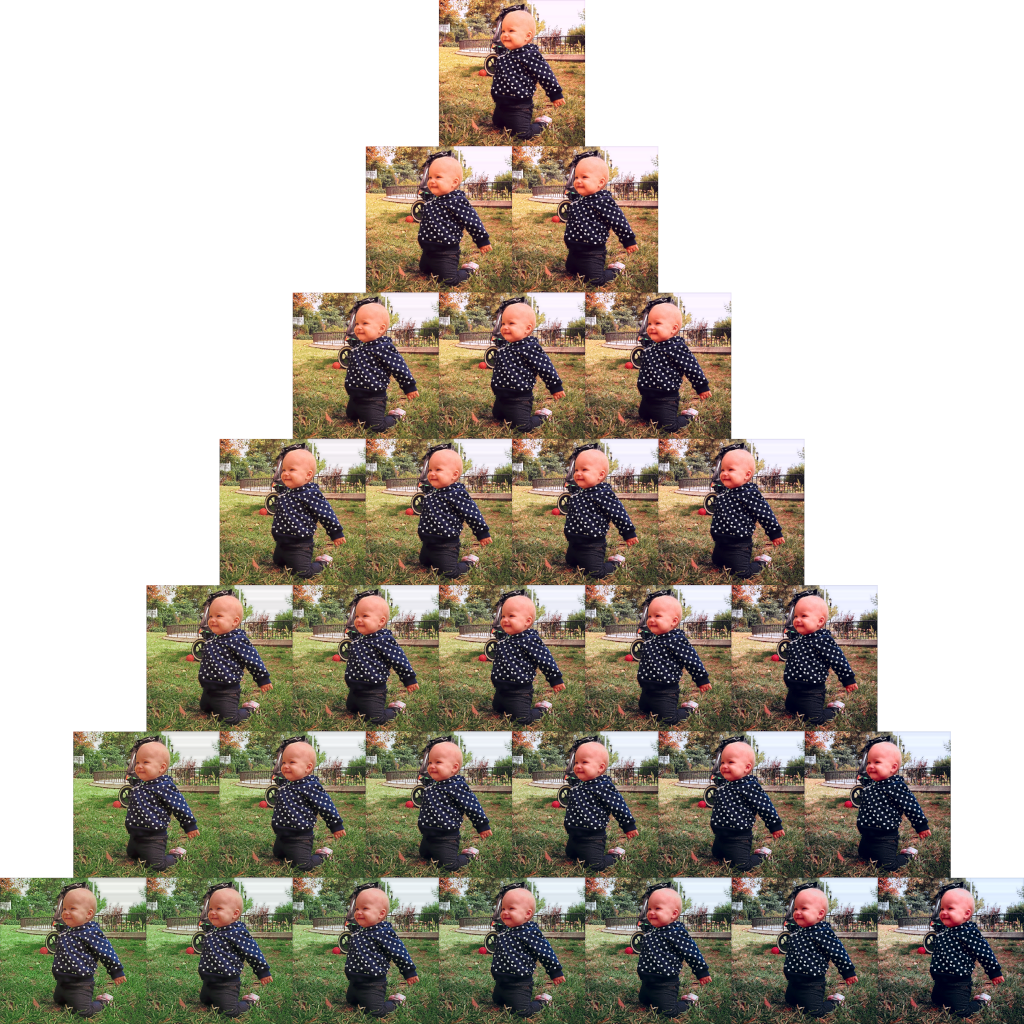}} \\

\raisebox{7\normalbaselineskip}[0pt][0pt]{\rotatebox[origin=c]{90}{\tiny \emph{ContoursDS}}} & \multicolumn{4}{c}{\includegraphics[width=0.31\columnwidth]{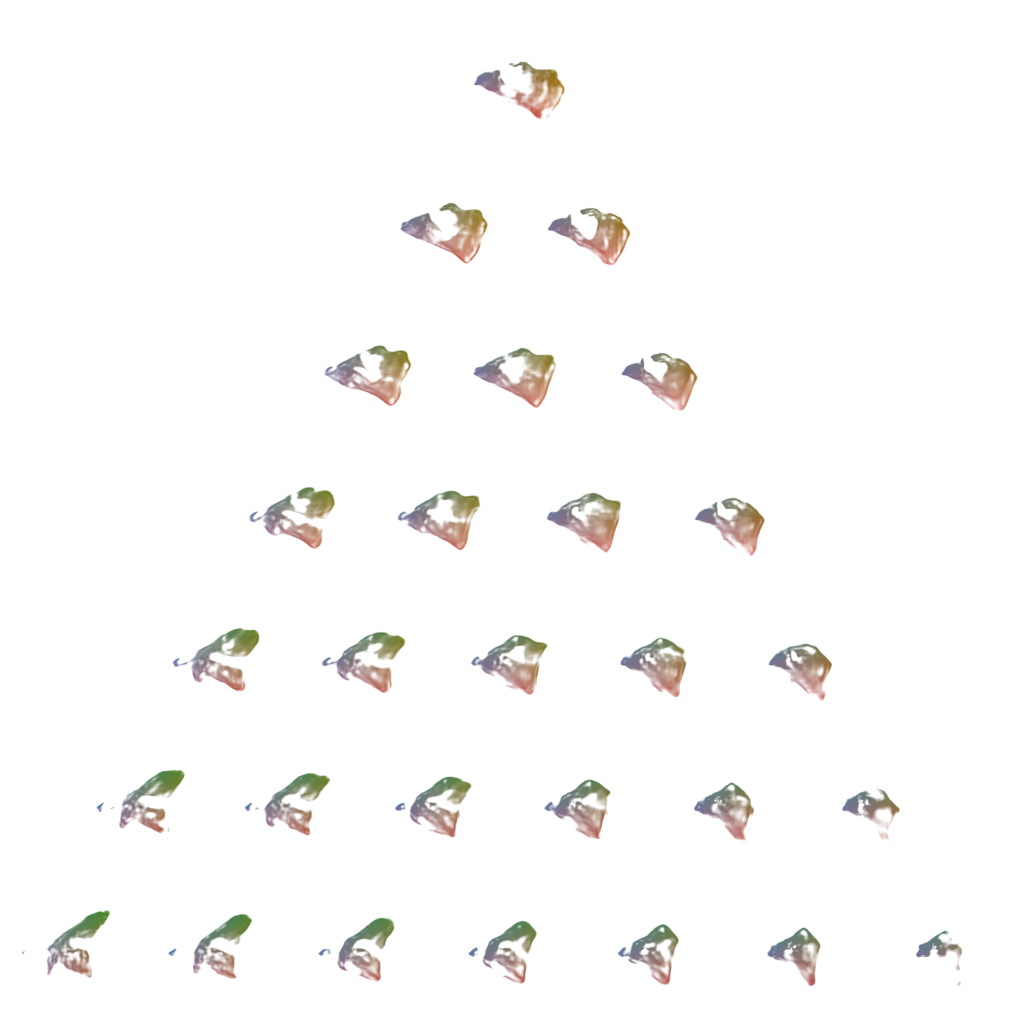}} & \multicolumn{4}{c}{\includegraphics[width=0.31\columnwidth]{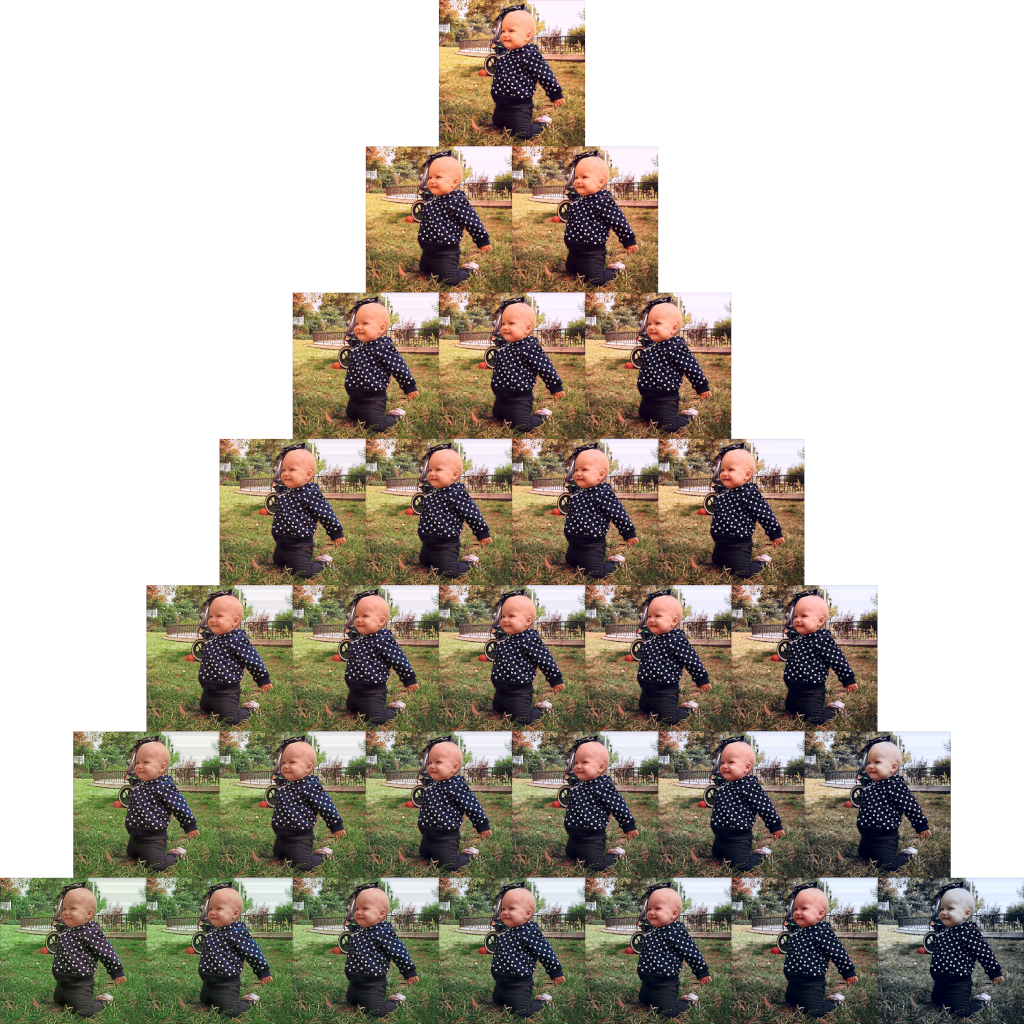}} \\

\raisebox{7\normalbaselineskip}[0pt][0pt]{\rotatebox[origin=c]{90}{\tiny \emph{HistoDS}}} & \multicolumn{4}{c}{\includegraphics[width=0.31\columnwidth]{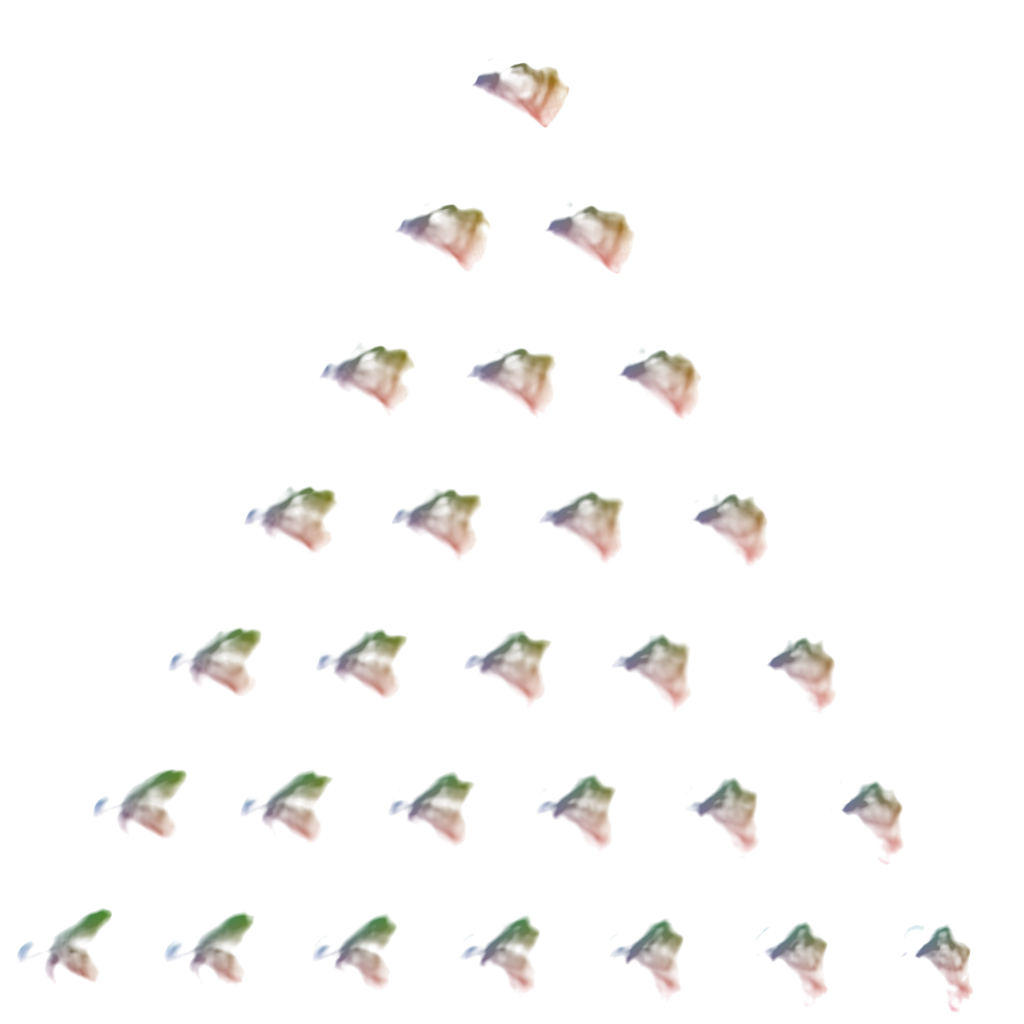}} & \multicolumn{4}{c}{\includegraphics[width=0.31\columnwidth]{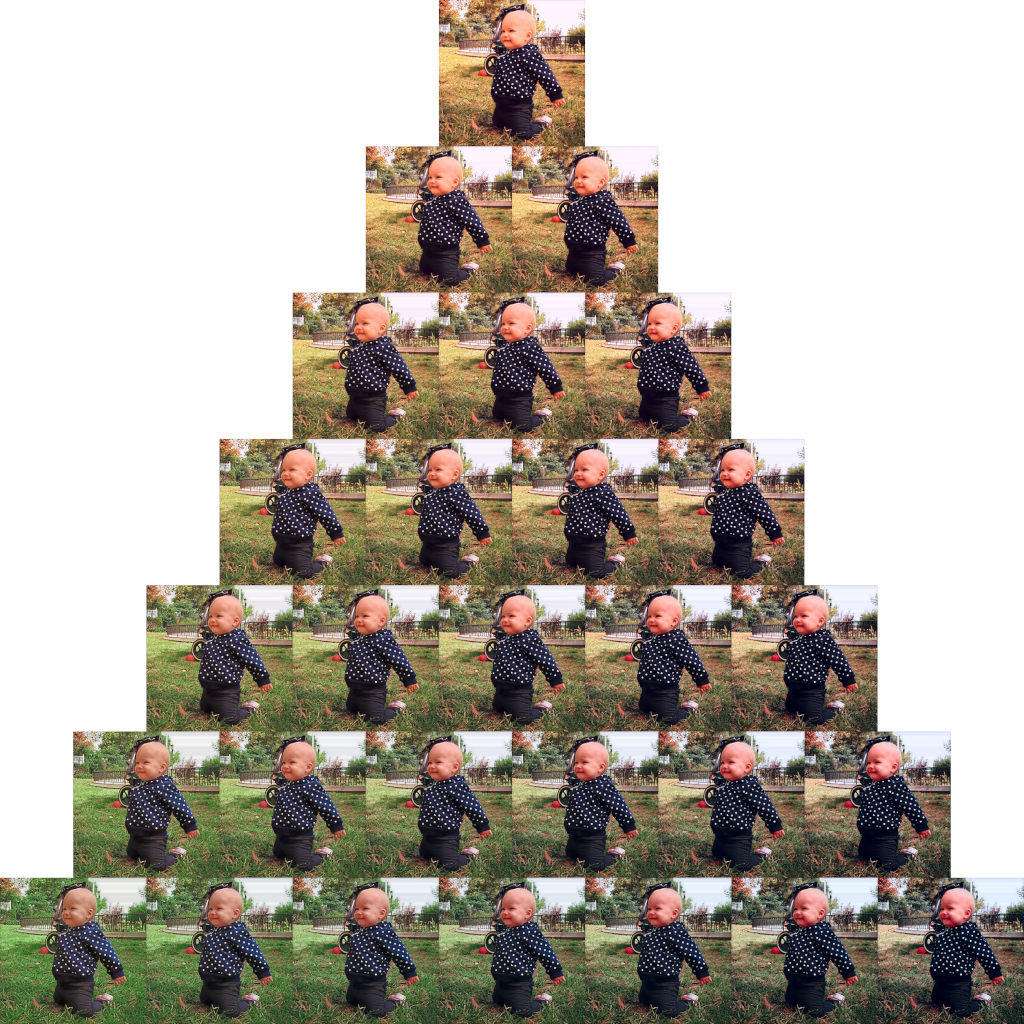}} \\
\end{tabular}
\caption{Color grading obtained by transferring the colors of $n=3$ images onto a target image, aiming at reproducing with our method the results from \citep{bonneel2015sliced}, figure 12. Results are shown in triangles (Left: interpolated chrominance histograms; Right: corresponding transfer results). The images corresponding to the target chrominance histogram $\nu$ and to the histograms $\mu_{i}$ - which are interpolated to obtain a barycenter - are shown in top row. Each $\mu_{i}$ corresponds to a vertex of the triangle in a clockwise order beginning with $i=1$ at the uppermost vertex. Each row presents the results for a different method, from top to bottom: GeomLoss, our model trained on synthetic shape contours (\emph{ContoursDS}) and our model trained on chrominance histograms from Flickr images (\emph{HistoDS})}
\label{fig:color_transfer_triangle}
\end{figure}

To better show the limitations of the generalization of our network when the number of inputs is $2$, we show additional interpolations between 2 objects from the \emph{Coil20} in figure \ref{figure:coil20_cup_car}. There are two reasons for these bad results: first, our model is trained in synthetic shape contours and do not look at all like these images. Furthermore, the cup image seem to be even more challenging than the car image for our network, and our best explanation for this failure is that the cup covers almost the whole image.
We provide additional experiments showing barycenters of 5 sketches on Figure \ref{figure:diff_pentagon}. The weights evolve linearly inside the pentagon. 
As a stress test, we also show a barycenter of 100 cats with equal weights in Fig.~\ref{fig:100way} and compare it with a barycenter computed with GeomLoss. While both results recover more or less the global shape of the cat, details are clearly lost and our result looks much smoother.
Finally, we provide an additional color transfer experiment in figure \ref{fig:color_transfer_triangle} reproducing an experiment from \cite{bonneel2015sliced} with our model trained with \emph{ContoursDS} and \emph{HistoDS}.

\label{appendix:generalization_n_inputs}

\section{Linearized barycenters}
\label{sec:linearized}
\begin{figure}[ht]
\vskip 0.2in
\centering
\begin{tabular}{ccccc}
{\tiny input 1} & {\tiny input 2} & {\tiny Geomloss (1) } & {\tiny Geomloss (10)} & {\tiny Ours} \\ 
\includegraphics[width=0.18\columnwidth]{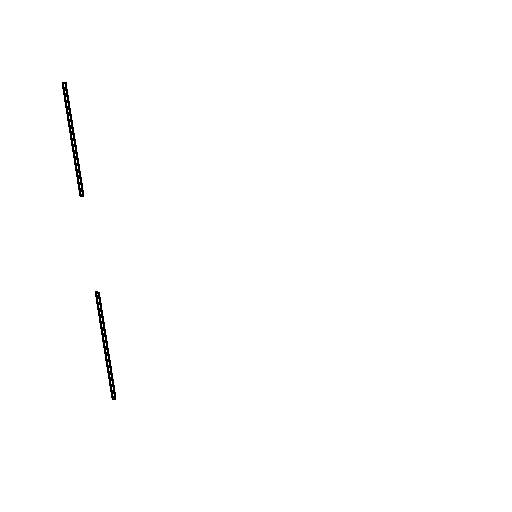} & \includegraphics[width=0.18\columnwidth]{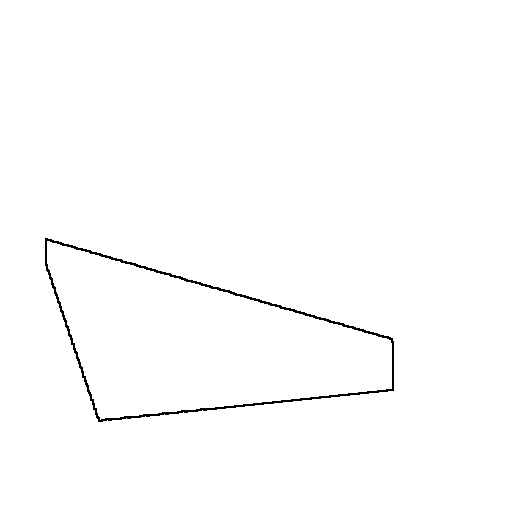} & \includegraphics[width=0.18\columnwidth]{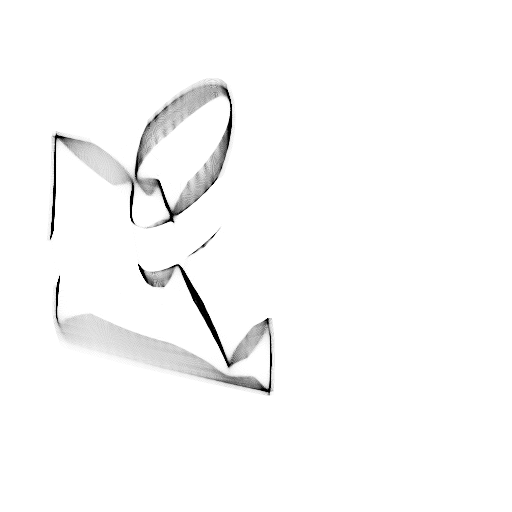} & \includegraphics[width=0.18\columnwidth]{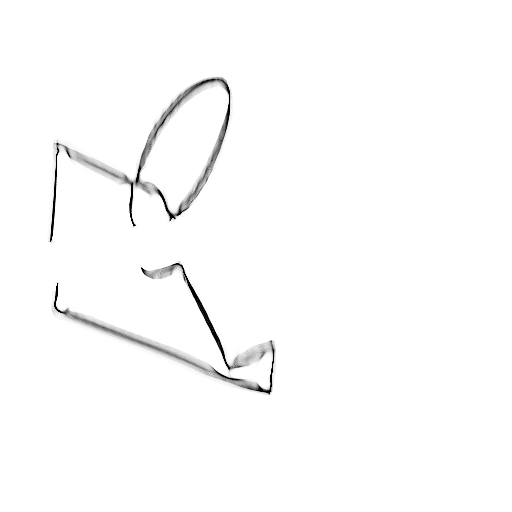} & \includegraphics[width=0.18\columnwidth]{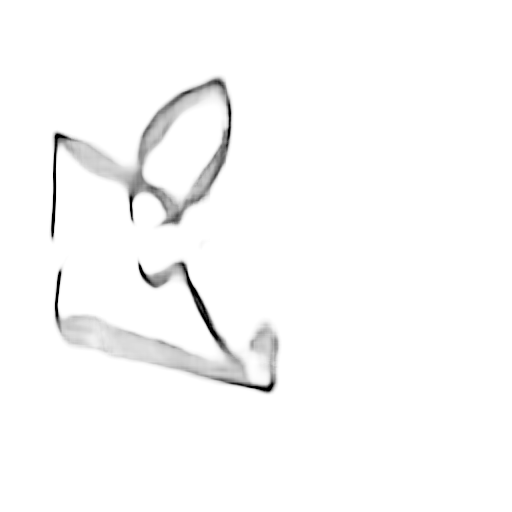} \\
 {\tiny $\lambda_{1}=0.4382$} & {\tiny $\lambda_{2}=0.5618$} & & & \\
\end{tabular}
\caption{Wasserstein barycenter computed from a pair of inputs respectively using Geomloss with only one descent step, Geomloss with $10$ descent steps and using our model trained on our synthetic training dataset}
\label{figure:geomloss_multi_iters_vs_ours}
\vskip -0.2in
\end{figure}

Fig.~\ref{figure:geomloss_multi_iters_vs_ours} shows the error introduced by using a linearized version of Wasserstein barycenters \citep{Nader18,wang2013linear,Moosmuller20,Merigot20}. Our predicted barycenters reflect this error.

\end{document}